\documentclass[nohyperref]{article}

\usepackage{microtype}
\usepackage{graphicx}
\usepackage{subfig}
\usepackage{booktabs}
\usepackage{hyperref}

\usepackage[accepted]{icml2022}
\usepackage{amsmath}
\usepackage{amssymb}
\usepackage{mathtools}
\usepackage{amsthm}

\theoremstyle{plain}
\newtheorem{theorem}{Theorem}[section]

\newtheorem{lemma}[theorem]{Lemma}
\newtheorem{corollary}[theorem]{Corollary}
\theoremstyle{definition}

\theoremstyle{remark}

\newcommand{\Prb}[1]{\mathbb{P}\left(#1\right)}
\newcommand{\E}{\mathbb{E}}
\newcommand{\ci}{\perp\!\!\!\perp}
\newcommand{\argmin}{\operatornamewithlimits{arg\;min}}

\usepackage{xspace}
\newcommand\ie{i.\,e.\xspace}
\newcommand\eg{e.\,g.\xspace}
\newcommand{\MILP}{\big(\textrm{MILP}\big)\xspace}
\newcommand{\RMILP}{\big(\textrm{RMILP}\big)\xspace}
\newcommand{\RMILPK}[1]{\textrm{RMILP}(#1)\xspace}
\newcommand{\MILPK}[1]{\textrm{MILP}(#1)\xspace}

\newcommand{\dd}[1][]{\mathrm{d}#1}
\usepackage[textsize=tiny]{todonotes}

\usepackage{bm}
\usepackage{amsfonts}
\usepackage{nicefrac}
\usepackage{placeins}
\usepackage[capitalize,noabbrev,sort&compress]{cleveref}
\crefname{algocf}{algorithm}{algorithms}
\Crefname{algocf}{Algorithm}{Algorithms}

\icmltitlerunning{Interpretable Off-Policy Learning via Hyperbox Search}

\begin{document}

\twocolumn[
\icmltitle{Interpretable Off-Policy Learning via Hyperbox Search}

\begin{icmlauthorlist}
\icmlauthor{Daniel Tschernutter}{ETH}
\icmlauthor{Tobias Hatt}{ETH}
\icmlauthor{Stefan Feuerriegel}{ETH,LMU}
\end{icmlauthorlist}

\icmlaffiliation{ETH}{ETH Zurich, Switzerland}
\icmlaffiliation{LMU}{LMU, Germany}

\icmlcorrespondingauthor{Daniel Tschernutter}{dtschernutter@ethz.ch}
\icmlcorrespondingauthor{Tobias Hatt}{thatt@ethz.ch}

\icmlkeywords{Interpretable Machine Learning, Off-policy Learning}

\vskip 0.3in
]

\printAffiliationsAndNotice{}

\begin{abstract}
Personalized treatment decisions have become an integral part of modern medicine. Thereby, the aim is to make treatment decisions based on individual patient characteristics. Numerous methods have been developed for learning such policies from observational data that achieve the best outcome across a certain policy class. Yet these methods are rarely interpretable. However, interpretability is often a prerequisite for policy learning in clinical practice. In this paper, we propose an algorithm for \emph{interpretable} off-policy learning via hyperbox search. In particular, our policies can be represented in disjunctive normal form (\ie, OR-of-ANDs) and are thus intelligible. We prove a universal approximation theorem that shows that our policy class is flexible enough to approximate any measurable function arbitrarily well. For optimization, we develop a tailored column generation procedure within a branch-and-bound framework. Using a simulation study, we demonstrate that our algorithm outperforms state-of-the-art methods from \emph{interpretable} off-policy learning in terms of regret. Using real-word clinical data, we perform a user study with actual clinical experts, who rate our policies as highly interpretable.
\end{abstract}

\section{Introduction}
\label{sec:introduction}

Personalized treatment decisions have received wide attention in clinical practice \cite{Chan.2011, Collins.2015, Hamburg.2010}. Thereby, the aim is to select treatments that are effective for individual patients. One way to formalize personalized decision-making are so-called policies, which map patient-level characteristics to treatment decisions. Numerous methods have been developed for learning such policies using data collected from existing clinical trials or observational studies, which are subsumed under \emph{off-policy learning} \cite{Athey.2021, Beygelzimer.2009, Dudik.2014, Kallus.2018, Kallus.2018b, Kitagawa.2018}.

A major challenge for off-policy learning in clinical practice is to learn policies that satisfy two requirements: (i)~the policy class must be \emph{sufficiently rich} and (ii)~the policies must be \emph{interpretable} \cite{Kattan.2016, Kosorok.2019, Blumlein.2022}. A policy class is said to be \emph{sufficiently rich} if it contains a large class of policies. This is desirable as it is directly related to policies that yield a low regret (\ie, the difference between clinical outcome across the population of the learned policy and the a~priori best policy should be small). A policy is said to be \emph{interpretable} if its decision can be explained in understandable terms to humans \cite{DoshiVelez.2017}. Particularly, in a clinical setting, it is crucial for clinical practitioners to understand \emph{which} treatment is chosen \emph{when} \cite{Ahmad.2018}. 

Logical models can achieve interpretability by providing a few descriptive rules that can help clinical practitioners to understand the decision process and get an intuition about the underlying data \cite{Wang.2017b,Rudin.2019}. Thereby, each rule is a conjunction of conditions \cite{Wang.2017b}. As an example, a set of rules for deciding whether or not a patient should receive a specific treatment with a drug could be:

{\footnotesize
\begin{tabular}{p{1.6cm}p{5.9cm}}
\toprule
\textbf{IF} a patient & (has an age $\le60$ \textbf{AND} a weight $\le80$ kg \textbf{AND} has mild symptoms)\\
\textbf{OR} & (has an age $\le40$ \textbf{AND} a weight $\le80$ kg \textbf{AND} has severe symptoms)\\
\textbf{OR} & (has an age $\le30$ \textbf{AND} a weight $\le100$ kg \textbf{AND} has severe symptoms)\\
\textbf{THEN} & treat with drug\\
\bottomrule
\end{tabular}}%

The above model suggests treatments and provides patient characteristics that lead to the decision. As such, it makes it easy for domain experts to understand and critically assess the underlying logic. For instance, in the above example, one observes that younger patients seem to better respond to the drug treatment even if they have severe symptoms and/or are too heavy. Thus, the age of a patient seems to be crucial for when it is recommended to apply the drug. Furthermore, patients over 60 years should not receive the drug treatment in general. Based on the decision logic, clinical practitioners can compare the suggested treatment rule against their domain knowledge, \eg, they may know that the molecular of the drug is non-functioning for older patients or has severe side-effects for them.

Besides being interpretable, a crucial part of logical models is the form of the conditions \cite{Rudin.2019}, \eg, age $\le60$ in the above example. The reason is that, even in simple settings, the best policy is often a nonlinear and complex function of patient characteristics \cite{Laber.2014,Robins.2004, Schulte.2014}. Hence, flexible policy classes should be used in order to mitigate the risk of model misspecification. For logical models, it is generally known that such flexibility results in computationally hard problems \cite{Rudin.2019}. Thus, the challenge is to formulate expressive conditions in a way that allows for efficient optimization.

\begin{figure}
	\label{fig:example_policy}
	\caption{Illustration of our interpretable policies.}
	\begin{center}
		\includegraphics*[scale=0.25]{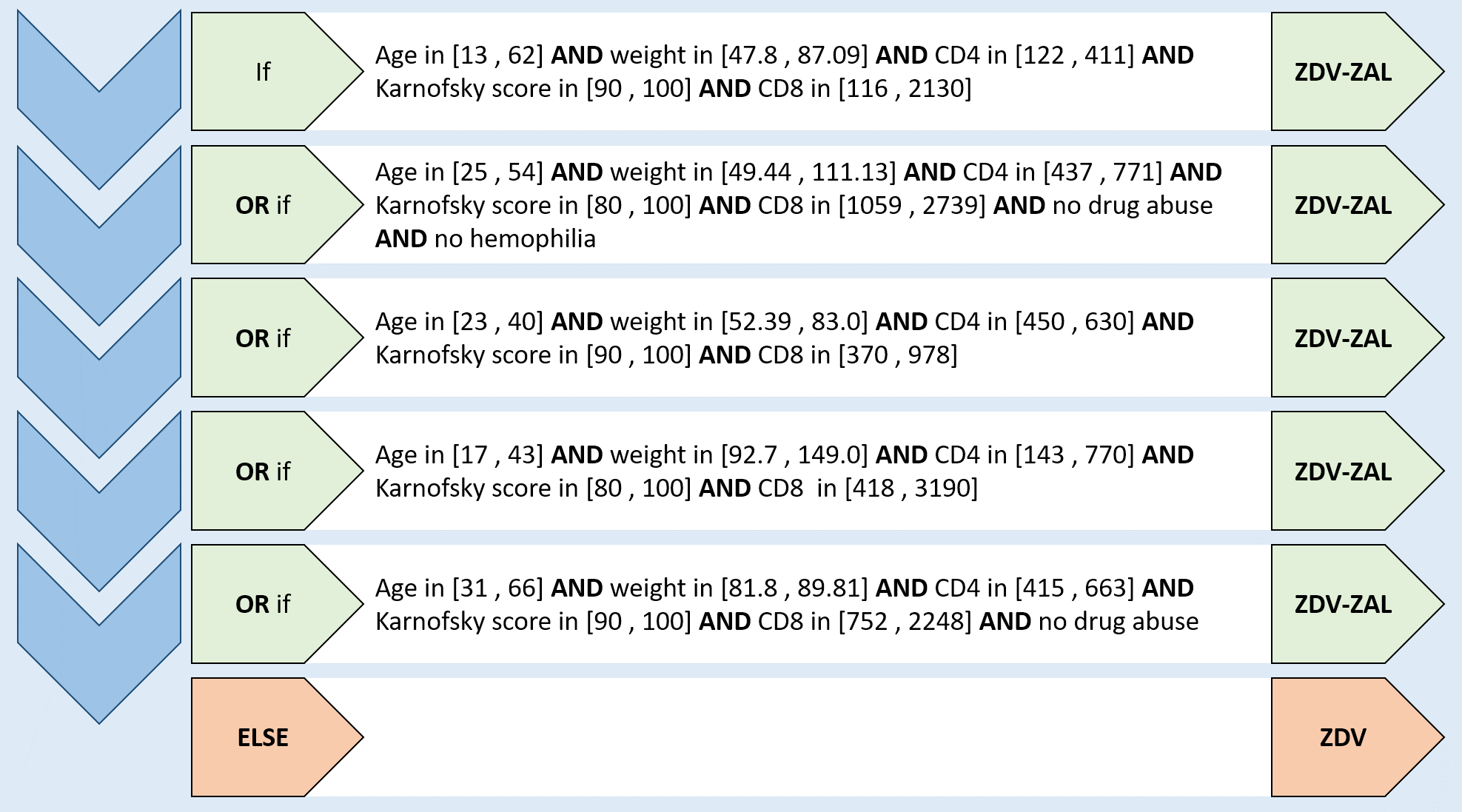}
	\end{center}
	{\scriptsize{\emph{Notes}. The policy was estimated using real-world clinical data from the ACTG study 175; see \Cref{sec:clinical_data} for details. A clinical practitioner follows the flow chart from top to bottom (as indicated by the blue arrows). If the patient fulfills an if-clause, a practitioner should assign the treatment in the corresponding arrow to the right. We provide details on the two treatments (ZDV-ZAL and ZDV) in \Cref{sec:clinical_data}.}}
	\vspace{-1em}
\end{figure}

In this paper, we address the problem of learning \emph{interpretable} policies from observational data, where the policies should originate from a sufficiently rich policy class. To this end, we propose an algorithm for \emph{interpretable off-policy learning} called IOPL. IOPL relies on an intelligible yet flexible class of policies. Specifically, our policy class is based on a representation of the decision region as a union of hyperboxes. As a result, our policies can be written in disjunctive normal form (\ie, OR-of-ANDs) and, thus, belong to the class of logical models. In particular, this representation gives rules which are conjunctions of interval conditions, \eg, age $\in[20,60]$. These are easy enough to be intelligible but expressive enough to approximate any measurable function arbitrarily well. An example of such a policy is shown in Figure~\labelcref{fig:example_policy}.

Optimizing over this policy class, \ie, learning optimal unions of hyperboxes, relies on a mixed integer linear program formulation. The latter involves an exponentially growing number of variables, which renders the optimization problem non-trivial. As a remedy, we develop a tailored optimization algorithm in form of a column generation procedure within a branch-and-bound framework (also known as branch-and-price). In this way, IOPL is able to efficiently search over the exponential number of hyperboxes and yields optimal solutions with theoretical guarantees. In addition, we prove a universal approximation theorem for our policy class: we show that our policies are flexible enough to approximate any measurable policy arbitrarily well. As a result, our policy class is sufficiently expressive to learn policies that are complex functions of patient characteristics. Using synthetic data, we demonstrate that IOPL outperforms existing baselines for interpretable off-policy learning in terms of regret. Furthermore, the regret of IOPL is on par with ``black box'' approaches (that are more flexible yet non-intelligible). We further confirm that IOPL yields policies that are perceived as interpretable in clinical practice. For this, we use real clinical data and perform a user study in which clinical experts assess the interpretability of our policies.

\section{Problem Setup}

We consider $n$ independent and identically distributed samples $\{(X_i,T_i,Y_i)\}_{i=1}^n$, where $X\in\mathcal{X}\subseteq\mathbb{R}^d$ denotes the patient covariates, $T\in\{-1,1\}$ denotes the treatment assignment, and $Y\in\mathbb{R}$ denotes the observed outcome. We use the convention that lower outcomes are more desirable. Using the Rubin-Neyman potential outcomes framework \cite{Rubin.2005}, we let $Y(1), Y(-1)$ be the potential outcomes for each of the treatments. To ensure identifiability of the outcomes from observational data, we make the following three standard assumptions \cite{Imbens.2015}: (i)~consistency (\ie, \mbox{$Y=Y(T)$}); (ii)~positivity (\ie, \mbox{$0<\Prb{T=t \mid X=x}$} for $t\in\{-1,1\}$ and all $x$); and (iii)~strong ignorability (\ie, \mbox{$Y(-1), Y(1) \ci T \mid X$}).

We let a \emph{policy} $\pi$ be a map from the patient covariates to a treatment decision, \ie, $\pi:\mathcal{X}\to\{-1,1\}$. Then, the \emph{policy value} of $\pi$ is given by
\vspace{-0.4em}
\begin{align}
\label{eq:value_function}
\begin{split}
V(\pi)=\mathbb{E}[Y(\pi(X))]=\mathbb{E}[&Y(1) I(\pi(X)=1)\\
+&Y(-1) I(\pi(X)=-1)],
\end{split}
\end{align}
which gives the population-wide outcome if treatments are assigned according to the policy $\pi$. The objective of \emph{policy learning} is to find a policy $\pi^\mathrm{opt}$ in a policy class $\Pi$ that minimizes the policy value $V(\pi)$, \ie,
\begin{align}\label{eq:policy_value_optimization}
\pi^\mathrm{opt} \in \argmin_{\pi\in\Pi}{V(\pi)}.
\end{align}
Most existing methods first estimate the policy value $V(\pi)$ in \labelcref{eq:value_function} and then optimize over a pre-specified policy class $\Pi$. The policy class $\Pi$ is often chosen such that $\pi^{\mathrm{opt}}$ achieves a low regret, \ie, such that Regret$(\pi^{\mathrm{opt}})=\min_{\pi\in\{-1,1\}^\mathcal{X}} V(\pi) - V(\pi^\text{opt})$ is small.\footnote{Thereby, we used $\{-1,1\}^\mathcal{X}$ to denote all functions from $\mathcal{X}$ to $\{-1,1\}$, \ie, all $\pi:\mathcal{X}\to\{-1,1\}$.} Hence, the policy class $\Pi$ must be rich enough to capture complex dependencies between patient covariates and outcome. This is usually achieved by choosing flexible policy classes such as kernel methods or neural networks \cite{Bennett.2020, Laber.2014,Qian.2011,Zhao.2019,Zhou.2017}. Even though flexible policy classes mitigate the risk of model misspecification, they render the estimated policy unintelligible. That is, interpretability is precluded. 

\section{Related Work}\label{sec::related_work}
In this section, we briefly review the literature on off-policy learning. As IOPL relies on the solution of a large-scale combinatorial optimization problem, we further discuss a technique called branch-and-price to solve such problems. The latter represents the foundation of IOPL as discussed in more detail later on.

\noindent\textbf{Off-policy learning.} There is a large literature on learning policies using data from observational data. Methods have been developed in research from both statistics \cite{Blatt.2004,Murphy.2003,Pan.2021,Qian.2011,Robins.2004,Zhao.2019,Zhou.2017} and machine learning \cite{Athey.2017,Athey.2021,Beygelzimer.2009,Dudik.2014, Kallus.2018,Kallus.2018b, Kitagawa.2018,Zhou.2018}. Therein, the authors first estimate the policy value in \labelcref{eq:value_function} and then optimize the policy value over a pre-specified policy class $\Pi$. Estimators for the policy value can be broadly divided into three categories: (i)~\emph{Direct methods}~(DM) estimate the outcome functions $\mu_t(x) = \E[Y(t)\mid X=x]$ and plug them into \Cref{eq:value_function} \cite{Bennett.2020, Qian.2011}. Direct methods are known to be weak against model misspecification with regards to $\mu_t(x)$. (ii)~\emph{Inverse propensity score weighted methods}~(IPS) seek weights that render the outcome data as if it were generated by a new policy \cite{Beygelzimer.2009,Bottou.2013,Horvitz.1952,Kallus.2018,Kallus.2018b,Li.2011}. As such, IPS methods correct for the distributional mismatch between the outcomes observed under the behavioral policy and the outcomes that would be observed under a new policy. (iii)~\emph{Doubly robust methods}~(DR) combine direct and IPS methods, typically using the augmented inverse propensity score estimator \cite{Thomas.2016,Dudik.2014,Athey.2021}. As a result, either the direct method or the IPS method need to be consistent for the doubly robust estimator to be consistent \cite{Dudik.2014,Thomas.2016}. Once an estimate of the policy value is obtained, the authors optimize over a pre-specified $\Pi$. In order to allow for complex policies, flexible policy classes such as kernel methods or neural networks are used \cite{Bennett.2020, Laber.2014, Qian.2011,Zhao.2019,Zhou.2017}. Yet, the interpretability of such policies is precluded, which limits the use in clinical practice \cite{Rudin.2019}.

Approaches for interpretable off-policy learning include works based on decision lists, \ie, lists of ``if-then'' clauses \cite{Lakkaraju.2017,Zhang.2018,Zhang.2015}, eligibility scores \cite{Kitagawa.2018} or decision trees \cite{Laber.2015}. However, besides being interpretable, none of these provide any theoretical guarantees that the used policy class is sufficiently rich to capture complex policies as we do. 

\noindent\textbf{Branch-and-price.} Branch-and-price is a general method to solve large scale integer linear programs. For a detailed introduction, we refer to \cite{Barnhart.1998}. Branch-and-price is based on the idea to use column generation within an enumeration framework, \ie, within a branch-and-bound approach. Column generation itself has already been used in machine learning research \cite{Bi.2004,Demiriz.2002,Li.2013}. Furthermore, column generation has also been used to derive interpretable classifiers \cite{Dash.2018} but not in the context of off-policy learning, resulting in three major differences to our work: (i)~\citeauthor{Dash.2018} consider general binary classification problems using the hamming loss, while we minimize the policy value over a certain policy class. (ii)~The authors merely develop a column generation framework for their relaxed problem formulation. The authors do not provide a branch-and-price approach as we do. Thus, they also offer no general convergence guarantees. (iii)~The authors make the assumption that only binary-valued features are given, whereas we refrain from any structural assumptions.

\section{Interpretable Off-Policy Learning}

In this section, we introduce our algorithm for interpretable off-policy learning. We proceed as follows. In \Cref{sec:policy_class}, we introduce a policy class $\Pi_{H}^M$ based on hyperboxes that is sufficiently rich while, at the same time, offers an intelligible structure and thus allows for interpretability. In \Cref{sec:MILP_formulation}, we reformulate the empirical version of \labelcref{eq:policy_value_optimization} as a mixed integer linear program. The latter is intractable due to an exponentially growing number of variables and is only for theoretical purposes to explain our conceptual framework. In \Cref{sec:theoretical_analysis}, we yield theoretical results regarding the approximation capability of our policy class $\Pi_{H}^M$, while in \Cref{sec:optimization_algorithm}, we introduce IOPL, which uses a tailored branch-and-price procedure to solve the aforementioned mixed integer linear program.

\subsection{Interpretable Policy Class $\Pi_{H}^M$ via Hyperboxes}
\label{sec:policy_class}

In the following, we formalize our policy class $\Pi_{H}^M$ consisting of policies that are built upon unions of hyperboxes. To do so, let $\mathcal{K}\subseteq2^{\mathbb{R}^d}$ denote a given finite set of hyperboxes in $\mathbb{R}^d$, \ie, $\mathcal{K}=\{S^1,\dots,S^N\}$ with ${S^j=[l^j_{1},u^j_{1}]\times[l^j_{2},u^j_{2}]\times\ldots\times[l^j_{d},u^j_{d}]}$ for $j\in\{1,\dots,N\}$. We consider policies of the form
\begin{align}
\footnotesize
\label{eq:definition_policy}
\pi_{\mathcal{D}}(X)=\begin{cases}
1, & \text{ if } \exists \ j\in\mathcal{D}\subseteq\{1,\dots,N\}: X\in S^j,\\
-1, & \text{ else,}
\end{cases}
\end{align}
where $\mathcal{D}$ denotes which of the $N$ hyperboxes are used for the treatment decision. Furthermore, we bound the number of used hyperboxes by $M$. That is, we define the policy class $\Pi_{H}^M=\{\pi_{\mathcal{D}}: \mathcal{D}\subseteq\{1,\dots,N\} \text{ and } \lvert\mathcal{D}\rvert\le M\}$. The policies in $\Pi_{H}^M$ can be represented as OR-of-ANDs and are thus in disjunctive normal form, which is deemed ``interpretable'' in the literature \cite{Rudin.2019}. An example is given in Figure~\labelcref{fig:example_policy}. We offer an in-depth discussion on the interpretability of our approach in Appendix~\ref{app:explainability}.

\subsection{Mixed Integer Linear Program Formulation}
\label{sec:MILP_formulation}

Optimizing the policy value over $\Pi_{H}^M$ as in \labelcref{eq:policy_value_optimization} is non-trivial. In the off-policy learning literature, \labelcref{eq:policy_value_optimization} is often reformulated as a binary classification problem. Our work also builds upon such a formulation, as it allows to reformulate the empirical version as a mixed integer linear program, which, in turn, allows for a column generation approach when relaxed (we show this in Sec.~\labelcref{sec:optimization_algorithm}). This will be the key ingredient of IOPL. 

We proceed as follows. Problem \labelcref{eq:policy_value_optimization} is equivalent to
\vspace{-0.5em}
\begin{align}
\label{eq:policy_value_optimization_psi}
\pi^\mathrm{opt} \in \argmin_{\pi\in\Pi} \ \mathcal{J}(\pi)
\end{align}
with $\mathcal{J}(\pi)=\E[\psi \ I(T\neq \pi(X))]$, where $I$ is the indicator function and $\psi$ corresponds to one of the three standard methods for policy learning: direct method $\psi^{\mathrm{DM}}$, inverse propensity score weighted method $\psi^{\mathrm{IPS}}$, and doubly robust method $\psi^{\mathrm{DR}}$. These are defined as
\vspace{-0.5em}
\begin{align}
\psi^{\mathrm{DM}}&=T(\mu_{-1}(X)-\mu_1(X)),\\ \psi^{\mathrm{IPS}}&=\frac{-Y}{e_T(X)},\\ \psi^{\mathrm{DR}}&=\psi^{\mathrm{DM}}+\psi^{\mathrm{IPS}}+\frac{\mu_T(X)}{e_T(X)}.
\end{align}
The nuisance functions $\mu_t(x)=\E[Y(t) \ | \ X=x]$ and $e_t(x)=\Prb{T=t \ | \ X=x}$ can be estimated from data. For details on this reformulation, see \Cref{app:classification_reformulation}. Then, we approximate \labelcref{eq:policy_value_optimization_psi}  with its empirical version resulting in the empirical binary classification problem
\vspace{-0.5em}
\begin{align}
\label{eq:empirical_policy_value}
\min\limits_{\pi\in\Pi} \mathcal{J}_n(\pi),
\tag{EBCP}
\end{align}
where $\mathcal{J}_n(\pi)=\frac{1}{n}\sum_{i=1}^n \psi_i I(T_i\neq \pi(X_i))$. By setting $\Pi=\Pi_{H}^M$ in \labelcref{eq:empirical_policy_value}, it is sufficient to consider only hyperboxes spanned by the subsets of $\{X_i\}_{i=1}^n$, \ie, hyperboxes with $l_{t}=\min_{i\in \mathcal{I}} X_{i,t}$ and $u_{t}=\max_{i\in \mathcal{I}} X_{i,t}$ for each $t\in\{1,\dots,d\}$, where $\mathcal{I}\subset\{1,\dots,n\}$ and $X_{i,t}$ denotes the $t$-th coordinate of $X_i$.\footnote{Note that an optimal union of hyperboxes with respect to \labelcref{eq:empirical_policy_value} can always be built from vertices of the hyperboxes spanned by the subsets of the covariates $\{X_i\}_{i=1}^n$. We can thus set $\mathcal{K}=\{S=[l_{1},u_{1}]\times[l_{2},u_{2}]\times\ldots\times[l_{d},u_{d}]: l_{t}=\min_{i\in \mathcal{I}} X_{i,t} \text{ and } u_{t}=\max_{i\in \mathcal{I}} X_{i,t} \text{ with } \mathcal{I}\subset\{1,\dots,n\}\}$} That is, the set $\mathcal{K}$ is indeed finite. However, this results in at most $2^n$ different boxes, \ie, $N\approx 2^n$, which calls for an efficient optimization approach. For this, \labelcref{eq:empirical_policy_value} can be reformulated as a mixed integer linear program as follows. From now on, we assume w.l.o.g. that $\psi_i\neq0$ for all $i\in\{1,\dots,n\}$. We split the set $\{1,\dots,n\}$ into indices containing positive values for $\psi$ and indices containing negative ones, \ie, ${\mathcal{P}=\{i\in\{1,\dots,n\}: \psi_i>0\}}$ and ${\mathcal{N}=\{i\in\{1,\dots,n\}: \psi_i<0\}}$. Furthermore, let ${I_t=\{i\in\{1,\dots,n\}: \ T_i=t\}}$ for $t\in\{-1,1\}$ denote the index sets of treated and untreated individuals and let $\mathcal{K}_i$ denote the set of all indices $j\in\{1,\dots,N\}$ such that $X_i\in S^j$. We introduce decision variables $s_j\in\{0,1\}$, which denote whether hyperbox $S^j$ is used in the policy or not, \ie, $\mathcal{D}=\{j: \ s_j=1\}$. Furthermore, let $\xi_i$ denote whether individual $i$ was treated according to $\pi_{\mathcal{D}}$ or not, \ie, $\xi_i=I(T_i\neq \pi_{\mathcal{D}}(X_i))$. The mixed integer linear program formulation of \labelcref{eq:empirical_policy_value}, which we denote by \MILP, is then as follows:
\vspace{-0.5em}
{\footnotesize
\begin{align}
&\min\limits_{s, \xi} \frac{1}{n}\sum\limits_{i=1}^n \psi_i \xi_i&&\label{eq:ILP_objective}\\
\textrm{s.t.} \ & \xi_i+\sum\nolimits_{j\in\mathcal{K}_i}s_j\ge 1 &&\hspace{-1em}\text{ for } i\in I_1\cap\mathcal{P},\label{eq:ILP_constraint_1}\\
&\xi_i\ge s_j &&\hspace{-1em}\text{ for } i\in I_{-1}\cap\mathcal{P} \text{ and } j\in\mathcal{K}_i,\label{eq:ILP_constraint_2}\\
&\xi_i\le 1-s_j &&\hspace{-1em}\text{ for } i\in I_1\cap\mathcal{N} \text{ and } j\in\mathcal{K}_i, \label{eq:ILP_constraint_3}\\
&\xi_i\le\sum\nolimits_{j\in\mathcal{K}_i}s_j &&\hspace{-1em}\text{ for } i\in I_{-1}\cap\mathcal{N},\label{eq:ILP_constraint_4}\\
&\sum\nolimits_{j=1}^N s_j\le M,&&\label{eq:ILP_constraint_5}\\
&s_j\in\{0,1\}, \xi_i\in[0,1].\label{eq:ILP_constraint_6}
\end{align}}%
Note that \MILP is equivalent to \labelcref{eq:empirical_policy_value} for $\Pi=\Pi_{H}^M$. See \Cref{app:equivalence_MILP} for a detailed discussion on the equivalence. Constraints \labelcref{eq:ILP_constraint_1,eq:ILP_constraint_3} enforce the equality ${\xi_i=I(T_i\neq \pi_{\mathcal{D}}(X_i))}$ for $i\in I_1$, while constraints \labelcref{eq:ILP_constraint_2,eq:ILP_constraint_4} enforce the same equality for $i\in I_{-1}$. Constraint \labelcref{eq:ILP_constraint_5} introduces the upper bound on the size of $\mathcal{D}$, \ie, the number of hyperboxes used for the treatment decision. The parameter $M$ can be used to control the complexity of the policy and, thus, also the degree of interpretability. Note that the integrality of $\xi_i$ follows by the integrality constraint of $s_j$. Hence, we dropped the extra constraint.

\subsection{Theoretical Analysis of $\Pi_H^M$}
\label{sec:theoretical_analysis}

In this section, we derive theoretical approximation guarantees for our policy class $\Pi_{H}^M$.

\subsubsection{Universal Approximation Theorem}
\label{sec:universal_approximation_theorem}

The approximation quality is strongly related to the regret, \ie, $\min_{\pi\in\{-1,1\}^\mathcal{X}} V(\pi) - V(\pi^\text{opt})$. In particular, we prove in \Cref{thm:universal_approximation_theorem} that, if an optimal policy $\pi^\ast\in\argmin_{\pi\in\{-1,1\}^\mathcal{X}} V(\pi)$ is a measurable function, it can be approximated arbitrarily well within our policy class $\Pi_{H}^M$.

\begin{theorem}
	\label{thm:universal_approximation_theorem}
	Let $1\le p<\infty$ and $\pi^\ast:\mathcal{X}\to\{-1,1\}$ be any Lebesgue measurable function\footnote{For simplicity, we assume $\mathcal{X}$ to have finite Lebesgue measure.}. Then, for every $\delta\in(0,1)$ and $\epsilon>0$, there exists a sample size $n_{\delta,\epsilon}\in\mathbb{N}$ and $M\in\mathbb{N}$ sufficiently large, as well as, a policy $\pi_{\mathcal{D}^\ast}\in\Pi_{H}^M$ as defined in \labelcref{eq:definition_policy}, such that
	\begin{align}
	\lVert \pi^\ast-\pi_{\mathcal{D}^\ast}\rVert_p<\epsilon
	\end{align}
	holds with probability at least $1-\delta$.
\end{theorem}

See Appendix~\ref{app:universal_approximation_theorem} for a proof. \Cref{thm:universal_approximation_theorem} proves that our policy class is sufficiently rich to approximate arbitrarily complex dependencies between patient covariates and treatment decisions.

\subsubsection{Asymptotic Estimation Properties} Constraint \labelcref{eq:ILP_constraint_5} in the definition of \MILP restricts the policy class to policies with $\lvert\mathcal{D}\rvert\le M$. From a computational perspective two questions remain: (i)~What is the asymptotic behavior of the estimated policy regarding the finite sample size, \ie, does the error $\lvert \mathcal{J}(\pi^\mathrm{opt})-\mathcal{J}(\hat\pi^\mathrm{opt}_n) \rvert$ for $\hat\pi^\mathrm{opt}_n\in\argmin_{\pi\in\Pi_H^M}\mathcal{J}_n(\pi)$ converge to zero for $n$ large enough? (ii)~What is the role of the parameter $M$ in this analysis, as the complexity of $\Pi_{H}^M$ seems to grow with $M$ and the sample size $n$? Both questions are addressed in the following.

\begin{theorem}
	\label{thm:rademacher_complexity}
	The empirical Rademacher complexity\footnote{The empirical Rademacher complexity is defined as $R_{n}(\Pi)=\frac{1}{n}\mathbb{E}_\sigma\left[\sup_{\pi\in\Pi}\sum_{i=1}^n \sigma_i\pi(X_i)\right]$, where the expectation is taken over i.i.d. samples of the Rademacher distribution, \ie, $\{\sigma_1,\dots,\sigma_n\}$ with $\mathbb{P}(\sigma_i=1)=\mathbb{P}(\sigma_i=-1)=\frac{1}{2}$.} $R_{n}(\Pi_{H}^M)$ of $\Pi_{H}^M$ can be bounded independently of the sample size $n$, \ie, it holds
	\begin{align}
	\label{eq:rademacher_complexity}
	R_{n}(\Pi_{H}^M)\le C\sqrt{2M},
	\end{align}
	for a constant $C>0$.
\end{theorem}

\noindent See Appendix~\ref{app:asymptotic_properties} for a proof. \Cref{thm:rademacher_complexity} directly implies the following Corollary.

\begin{corollary}
	\label{cor:asymptotic_behavior}
	For a fixed parameter $M$ it holds
	\begin{align}
	\lvert \mathcal{J}(\pi^\mathrm{opt})-\mathcal{J}(\hat\pi^\mathrm{opt}_n) \rvert = \mathcal{O}_p(\nicefrac{1}{\sqrt{n}}),
	\end{align}
	where $\hat\pi^\mathrm{opt}_n\in\argmin_{\pi\in\Pi_H^M}\mathcal{J}_n(\pi)$.
\end{corollary}

\noindent See Appendix~\ref{app:asymptotic_properties} for a proof. Corollary~\ref{cor:asymptotic_behavior} shows that optimizing \labelcref{eq:empirical_policy_value} yields near-optimal solutions for the policy value optimization problem \labelcref{eq:policy_value_optimization} with $\Pi=\Pi_H^M$. Furthermore, \Cref{thm:universal_approximation_theorem} shows that, for any given $\delta$ and $\epsilon$, there exists a sample size $n_{\delta,\epsilon}\in\mathbb{N}$ and $M^\ast=\lvert\mathcal{D}^\ast\rvert$, such that with probability $1-\delta$ there exists a policy $\pi_{\hat{\mathcal{D}}}\in\Pi_H^M$ for every $M\ge M^\ast$ with $\lVert \pi^\ast-\pi_{\hat{\mathcal{D}}}\rVert_p<\epsilon$.

\subsection{Choice of the Number of Hyperboxes}

Evidently, the derivation of $M^\ast$ is non-constructive. As a remedy, we discuss two approaches to calibrate the number of hyperboxes, \ie, rules, in practice: (i)~expert-informed selection and (ii)~an expert-informed penalty method.

\textbf{Expert-informed selection.} We expect clinical practitioners to bound the overall number of rules in our model. To do so, we explicitly model the number of hyperboxes as a constraint, see \labelcref{eq:ILP_constraint_5}. We think that this is natural in our setting in which interpretable policies are preferred. In this way, the accuracy-complexity tradeoff can easily be controlled in practice by an expert-informed hyperparameter grid for $M$.

\textbf{Expert-informed penalty method.} If it is preferred to determine the number of hyperboxes, \ie, rules, in a fully data-driven manner, we propose a penalty method for \MILP in Appendix~\ref{app:algorithmic_calibration}. In this case, practitioners have to set a value $\omega>0$ they are willing to sacrifice in the policy value in order to have one fewer rule in the final model \cite{Rudin.2019}. In our case study with real-world data, for instance, $Y$ decodes the increase in the CD4 cell count after 20 weeks. That is, setting for example $\omega=25$, one is willing to sacrifice a decrease of the population wide CD4 cell count by 25 cells/$\text{mm}^3$ for every rule that is eliminated in the final model.

\subsection{Derivation of IOPL}
\label{sec:optimization_algorithm}

The above formulation of \MILP is impractical for real-world datasets as the number of binary variables grows exponentially in $n$. As a remedy, we now propose an efficient optimization algorithm for solving \MILP, which we call IOPL. IOPL consists of (i)~a branch-and-bound framework that theoretically allows to solve \MILP to optimality (see \Cref{sec:branch_and_bound}) and (ii)~a column generation procedure that is able to solve the involved relaxations of \MILP (see \Cref{sec:branch_and_bound_subproblems}). Our algorithm thus falls in the wider area of branch-and-price algorithms \cite{Barnhart.1998}.

Our optimization algorithm proceeds as follows. Using a branch-and-price framework, we successively solve \emph{relaxations} of \MILP, where constraint \labelcref{eq:ILP_constraint_6} is replaced by $s_j,\xi_i\in[0,1]$ for all $i$ and $j$. We denote this relaxation by \RMILP. If a solution has a non-integer value in an $s_j$, a cutting plane is used to branch into two subproblems where each subproblem fixes $s_j$ either to zero or to one, \ie, adds constraints of the form $s_j\le 0$ or $s_j\ge 1$, for a certain $j$ determined in a so-called branching rule. To solve the relaxed problem \RMILP to optimality, most of the hyperboxes of $\mathcal{K}$ are left out as there are too many to handle them efficiently. For this, we restrict the set of hyperboxes $\mathcal{K}$ to a subset $\mathcal{W}\subseteq\mathcal{K}$ and solve the \emph{restricted} problem denoted by \RMILPK{$\mathcal{W}$}. To check if the solution of the restricted problem is also optimal for the unrestricted problem, we solve the so-called pricing problem. In the pricing problem, we try to identify new hyperboxes $S\in \mathcal{K}\setminus \mathcal{W}$ that one can add to the restricted linear program (by adding it to the set $\mathcal{W}$). This is done by identifying missing hyperboxes that yield a negative reduced cost. If all hyperboxes that have not been added have a non-negative reduced cost (\ie, the optimal solution of the pricing problem is greater or equal to zero), the solution of the restricted problem \RMILPK{$\mathcal{W}$} and the solution of the unrestricted problem $\RMILP=\RMILPK{\mathcal{K}}$ coincide. Note that each solution of \RMILP yields a lower bound of \MILP, which is used to prune suboptimal branches.

\subsubsection{Branch-and-Bound Framework for IOPL}
\label{sec:branch_and_bound}

In this section, we formalize IOPL; see Algorithm \ref{alg:branch_and_bound}. We use the set $\mathcal{C}\subseteq\{1,\dots,\lvert\mathcal{W}\rvert\}\times\{1,0\}$ within \RMILPK{$\mathcal{K}$,$\mathcal{C}$} to denote the added cutting planes, \ie, the indices $j\in\{1,\dots,\lvert\mathcal{W}\rvert\}$ and the type of cut (\ie, $s_j\le 0$ or $s_j\ge 1$).

\begin{algorithm}
	\caption{IOPL}
	\label{alg:branch_and_bound}
	\tiny
	\begin{algorithmic}
	\STATE {\bfseries Input:} Initial working set $\mathcal{W}_0$
	\STATE {\bfseries Output:} Optimal subset $\mathcal{K}^\ast\subseteq\mathcal{K}$, optimal solution $s^\ast$
	\STATE Initialize the list of active subproblems $\mathcal{L}\gets\{\RMILPK{\mathcal{W}_0,\emptyset}\}$
	\STATE Initialize iteration counter $l\gets 0$
	\REPEAT
	\STATE Select and remove first subproblem $\RMILPK{\mathcal{W},\mathcal{C}}$ from $\mathcal{L}$
	\STATE Perform column generation to get the current relaxed solution and the new working set $s^\prime, v^\prime,\mathcal{W}^\prime\gets\mathrm{ColumnGeneration}(\RMILPK{\mathcal{W},\mathcal{C}})$
	\IF{$l=0$}
		\STATE Solve restricted integer problem to get current optimal integer solution and objective value $s^\ast,v^\ast\gets \mathrm{Solve}(\MILPK{\mathcal{W^\prime}})$
		\STATE Update optimal subset $\mathcal{W}^\ast\gets\mathcal{W}^\prime$
	\ENDIF
	\IF{$v^\prime\le v^\ast$}
		\IF{$s^\prime$ integral}
			\STATE Update new optimal integer solution $(\mathcal{W}^\ast,s^\ast)\gets (\mathcal{W}^\prime,s^\prime)$
		\ELSE
			\STATE Set $j^\prime$ according to branching rule
			\STATE Branch by updating $\mathcal{L}\gets\mathcal{L}\cup\{ 		\RMILPK{\mathcal{W}^\prime,\mathcal{C}\cup\{(j^\prime,1)\}}, 		\RMILPK{\mathcal{W}^\prime,\mathcal{C}\cup\{(j^\prime,0)\}}\}$\label{line:branching}
			\STATE Solve restricted problem $(s^\prime,v^\prime)\gets\mathrm{Solve}(\MILPK{\mathcal{W}^\prime})$
			\IF{$v^\prime\le v^\ast$}
				\STATE Update new optimal integer solution $(\mathcal{W}^\ast,s^\ast)\gets (\mathcal{W}^\prime,s^\prime)$
			\ENDIF
		\ENDIF
	\ENDIF
	\STATE Increase counter $l\gets l+1$
	\UNTIL{$\mathcal{L}$ empty}
	\STATE {\bfseries return} $\mathcal{W}^\ast$, $s^\ast$
	\end{algorithmic}
\end{algorithm}
\Cref{alg:branch_and_bound} proceeds as follows. We initialize the list of active subproblems. Then, the algorithm loops through this list until there are no active subproblems left. In each step, we select the first active subproblem from the list and solve it via column generation (see next section for details). If the optimal objective function value is larger than the current best value, we can prune the current branch and continue. If, on the other hand, the objective function value is lower than the current best value, the algorithm checks if the corresponding solution is integral or not. In case it is already integral, we update the current solution. If the solution is fractional, we branch according to a specified branching rule. The latter returns an index $j^\prime$ and adds the corresponding two subproblems to the set of active subproblems. In addition to these branch-and-bound steps, we also solve the restricted \MILP, denoted by \MILPK{$\mathcal{W}$}, after each subproblem solution that is fractional and in the first iteration. This results in a faster convergence to reasonable candidate solutions by allowing for faster pruning.

\noindent\textbf{Customized branching rule.} A crucial step in every branch-and-bound algorithm is the branching rule. We first experimented with standard approaches such as most infeasible branching \cite{Achterberg.2005}, \ie $j^\prime\gets\arg\min_j\lvert s^\prime_j-0.5\rvert$. However, we achieve a faster convergence through the following customized branching rule: We set $j^\prime$ to the index which corresponds to the box $S^{j^\prime}$ with the largest volume among all boxes with fractional $s_{j^\prime}$. In this way, we fix large boxes early in the branching process, which leads to a faster convergence as these boxes usually have the largest effect in the objective function.

\subsubsection{Subproblem Solutions via Column Generation}
\label{sec:branch_and_bound_subproblems}

We now show how to solve the involved subproblems in \Cref{alg:branch_and_bound}. For this purpose, we propose an efficient column generation procedure for the subproblems. To do so, we first derive the reduced cost for a potential new hyperbox $S$. For this, let $\mu_i^1\ge 0$ for $i\in I_1\cap\mathcal{P}$, $\mu_{i,j}^2\ge 0$ for $i\in I_{-1}\cap\mathcal{P}$ and $j\in\mathcal{K}_i$, $\mu_{i,j}^3\ge 0$ for $i\in I_1\cap\mathcal{N}$ and $j\in\mathcal{K}_i$, and $\mu_i^4\ge 0$ for $i\in I_{-1}\cap\mathcal{N}$ denote the dual variables associated with constraints \labelcref{eq:ILP_constraint_1,eq:ILP_constraint_2,eq:ILP_constraint_3,eq:ILP_constraint_4} of \RMILPK{$\mathcal{W}$}. Furthermore, let $\lambda\ge0$ be the dual variable associated with constraint \labelcref{eq:ILP_constraint_5}. The reduced cost for a new potential hyperbox $S$ is given by
\begin{align}
\scriptsize
\label{eq:reduced_cost}
\begin{split}
&-\sum\limits_{i\in I_1\cap\mathcal{P}} \mu_i^1\delta_i+\sum\limits_{i\in I_{-1}\cap\mathcal{P}} \Bigg(\sum\limits_{j\in\mathcal{K}_i}\mu_{i,j}^2\Bigg)\delta_i\\
&+\sum\limits_{i\in I_1\cap\mathcal{N}} \Bigg(\sum\limits_{j\in\mathcal{K}_i}\mu_{i,j}^3\Bigg)\delta_i
-\sum\limits_{i\in I_{-1}\cap\mathcal{N}} \mu_i^4\delta_i+\lambda,
\end{split}
\end{align}
where $\delta_i\in\{0,1\}$ denotes whether $X_i\in S$ holds. The first four terms are the sums of the dual variables associated with the respective constraints in which the potential hyperbox appears. Then, the pricing problem consists of minimizing \labelcref{eq:reduced_cost} over all potential hyperboxes.

Based on the idea of column generation, we now successively solve the pricing problem. If the minimal value is greater or equal to zero, there is no hyperbox that can be added with a negative reduced cost and, hence, the column generation procedure terminates. If the minimal value is smaller than zero, we (i)~add the corresponding hyperbox to the working set $\mathcal{W}$, (ii)~solve \RMILPK{$\mathcal{W}$} for the new working set $\mathcal{W}$ to update the dual variables, and (iii)~solve again the pricing problem. In this manner, the working set $\mathcal{W}$ grows until the solution of \RMILPK{$\mathcal{W}$} and \RMILP coincide. Note also that the reduced cost is independent of additionally added cutting planes and, hence, can be used to generate new columns independently of the set $\mathcal{C}$.

One observes that \labelcref{eq:reduced_cost} is a linear combination of $\delta_i$ with positive and negative weights. Determining a hyperbox that maximizes such an objective is called hyperbox search in the literature \cite{Louveaux.2014}. We assume w.l.o.g. that all coefficients (\ie, dual variables or sums of dual variables) are strictly larger than zero (otherwise they can be ignored in the corresponding summation) and define the pricing problem as in \cite{Louveaux.2014} by
\vspace{-1em}

{
\begin{align}
\scriptstyle\max\limits_{\delta,\gamma,p,q} &\scriptstyle\sum\limits_{i\in I_1\cap\mathcal{P}} \mu_i^1\delta_i-\sum\limits_{i\in I_{-1}\cap\mathcal{P}} \Bigg(\sum\limits_{j\in\mathcal{K}_i}\mu_{i,j}^2\Bigg)\delta_i\\
\scriptstyle-&\scriptstyle\sum\limits_{i\in I_1\cap\mathcal{N}} \Bigg(\sum\limits_{j\in\mathcal{K}_i}\mu_{i,j}^3\Bigg)\delta_i
+\sum\limits_{i\in I_{-1}\cap\mathcal{N}} \mu_i^4\delta_i-\lambda\label{eq:Pricing_objective}\\
\scriptstyle\textrm{s.t.} \quad  &\scriptstyle\delta_i\le \gamma_t^{r^i_t} \ \text{for all} \ i\in (I_1\cap\mathcal{P})\cup (I_{-1}\cap\mathcal{N}) \ \text{ and } \ t\in\{1,\dots,d\},\label{eq:Pricing_constraint_1}\\
&\scriptstyle\delta_i+\sum\nolimits_{t=1}^d (1-\gamma_t^{r^i_t})\ge 1 \ \text{for all} \ i\in (I_{-1}\cap\mathcal{P})\cup (I_1\cap\mathcal{N}),\label{eq:Pricing_constraint_2}\\
&\scriptstyle\gamma_t^k=p_t^k-q_t^k \ \text{for all} \ t\in\{1,\dots,d\} \ \text{and} \ k\in\{1\dots,n_t\},\label{eq:Pricing_constraint_3}\\
&\scriptstyle p_t^k\le p_t^{k+1} \ \text{for all} \ t\in\{1,\dots,d\} \ \text{and} \ k\in\{1\dots,n_t-1\},\label{eq:Pricing_constraint_4}\\
&\scriptstyle q_t^k\le q_t^{k+1} \ \text{for all} \ t\in\{1,\dots,d\} \ \text{and} \ k\in\{1\dots,n_t-1\},\label{eq:Pricing_constraint_5}\\
&\scriptstyle\delta_i,\gamma_t^k,p_t^k,q_t^k\in\{0,1\}.\label{eq:Pricing_constraint_6}
\end{align}}%
Thereby, the following definitions and assumptions have been made. In each dimension $t\in\{1,\dots,d\}$, we sort the covariates of patients in increasing order, \ie, sort the set $\{X_{i,t}\}_{i=1}^n$. If some patients have the same covariate in dimension $t$, we remove it from the set. This results in a list of $n_t$ distinct and sorted values for each dimension. For a patient $i$, we then define the index in that list in dimension $t$ by $r^i_t$. Furthermore, let $\gamma_t^k$ denote whether or not the $k$-th value in this list lies within the projected interval of the candidate hyperbox onto dimension $t$. Now, constraint \labelcref{eq:Pricing_constraint_1} denotes that an individual with positive coefficient, \ie, $i\in(I_1\cap\mathcal{P})\cup (I_{-1}\cap\mathcal{N})$, can be selected if the respective covariate lies within the projected interval of the candidate hyperbox in each dimension. Constraint \labelcref{eq:Pricing_constraint_2} says that a patient with negative coefficient, \ie, $i\in(I_{-1}\cap\mathcal{P})\cup (I_1\cap\mathcal{N})$, can be excluded if, in at least one dimension, the covariate lies outside the projected interval of the candidate hyperbox. Constraints \labelcref{eq:Pricing_constraint_3,eq:Pricing_constraint_4,eq:Pricing_constraint_5} encode the consecutive property of $\gamma_t^k$ with auxiliary variables $p_t^k$ and $q_t^k$; see \cite{Louveaux.2014} for details. Finally, the resulting box can be built from the values of $\delta$, \ie, for each dimension $t$ we set $l_{t}=\min_{\{i:\delta_i=1\}} X_{i,t}$ and $u_{t}=\max_{\{i:\delta_i=1\}} X_{i,t}$.

\noindent\textbf{Computational complexity.} The computational complexity of IOPL is mainly driven by the cost for solving the pricing problem. While this is known to be NP-hard \cite{Louveaux.2014}, we find that we can efficiently solve real-world problem instance with $n\approx1,000$, which is a typical problem size in clinical practice (see \Cref{sec:clinical_data}). One reason might be that a large part of the dual variables $\mu$ are zero during optimization.

\section{Experiments}
\label{sec:experiments}
In this section, we investigate IOPL in a variety of settings. Using simulated data, we illustrate how IOPL approximates the optimal decision regions and compare the regret against state-of-the-art methods from \emph{interpretable} off-policy learning. The latter shows that IOPL yields regrets that are substantially lower than state-of-the-art methods. Furthermore, IOPL performs on par with state-of-the-art ``black box'' methods in off-policy learning. Using real-world clinical data, we asses the interpretability of our policies. For this, we conduct a user study with clinical experts to rate our policies in terms of interpretability, which demonstrates that IOPL indeed yields policies that are perceived as being interpretable in practice.

\noindent\textbf{Baselines.} We consider three baselines from state-of-the-art interpretable off-policy learning: (i)~\emph{Linear eligibility score}~(ES): For this baseline, we set $\Pi$ to the class of linear functions decoding a treatment ($T=1$) if the value is greater or equal than zero and no treatment ($T=-1$) otherwise \cite{Kitagawa.2018}. (ii)~\emph{Decision lists}~(DL): For this baseline, we consider decision lists as in \cite{Zhang.2018}. (iii)~\emph{Tree-based policy learner}~(DT): For this baseline, we consider a tree-based approach with doubly robust correction techniques \cite{Zhang.2012}. All baselines are considered as \emph{interpretable} \cite{Rudin.2019}.

For comparison, we also report the performance of a ``black box'' model, however, for which interpretability is precluded. We consider the class of \emph{deep neural networks}~(DNN), which allows for complex decision boundaries but is considered non-intelligible \cite{Rudin.2019}. This allows to give a lower bound on the performance, when dropping the interpretability requirement from clinical practice. Details on all baselines and hyperparameters are given in \Cref{app:baselines}.

\noindent\textbf{Performance.} We evaluate the performance in our simulation study by reporting the regret, defined as $\min_{\pi\in\{-1,1\}^\mathcal{X}} V(\pi) - V(\hat\pi^\text{opt})$. To do so, we sample i.i.d. observations and use a subset for training. Afterward, the regret is computed on all sampled observations. We use the exact nuisance functions $\mu_t$ and $e_t$ in the simulation study for all baselines and IOPL. For the real-world clinical data, we use a kernel regression for $\mu_t$ and logistic regression for $e_t$.

\noindent\textbf{Implementation of IOPL.} We provide a publicly available implementation of IOPL in Python. For solving the LP (linear program) relaxations and the pricing problem, we use Gurobi 9.0. We stop IOPL if it exceeds $l=50$ branch-and-bound iterations as given in \Cref{alg:branch_and_bound}. We set a maximum time limit of 180 seconds for solving the pricing problem in our experiments. We emphasize that this time limit was never exceeded in our experiments, including the experiments with the real-world clinical data (see \Cref{sec:branch_and_bound_subproblems} for a discussion of the reasons).\footnote{Code available at \url{https://github.com/DanielTschernutter/IOPL}}

\noindent\textbf{Hardware.} We run all of our experiments on a server with two 16-core Intel Xeon Gold 6242 processors each with 2.8GHz and 192GB of RAM.

\subsection{Simulation Study}
\label{sec:simulation_study}
We perform a simulation study to further investigate IOPL. In particular, our objective is to (i)~illustrate how our policies approximate the optimal decision region and (ii)~compare the performance to state-of-the-art interpretable off-policy learning methods.

\noindent\textbf{Illustration.} For illustrative purposes, we generate 2-dimensional patient covariates $X_0,X_1\sim U[-1,1]$, a treatment assignment which is independent of $X$ with $\Prb{T=1}=\nicefrac{1}{2}$, and an outcome $Y\mid X, T\sim \mathcal{N}(m(X, T), \sigma^2)$, where  $m(X,T)=1+2X_0+X_1+f(X,T)$ and $\sigma=0.1$. The experimental setup is based on \citet{Zhao.2012}. The benefit of a 2-dimensional setting is that it allows to visualize the true decision regions. The function $f$ introduces interactions between treatments and the outcome variable. We investigate three different scenarios of increasing complexity, see \Cref{app:illustration_of_IOPL} for detailed definitions. The first scenario (``basic'') corresponds to a parabola as the decision boundary. The second scenario (``complex'') describes a complex decision boundary for which it is optimal to treat individuals with covariates inside a ring. The third scenario (``very complex'') corresponds to the most complex scenario with two disconnected decision regions and highly nonlinear decision boundaries.

In Figure~\labelcref{fig:approximation} from \Cref{app:universal_approximation_theorem_visualization}, we observe that, for smaller numbers of hyperboxes, \ie smaller $M$, IOPL yields a coarse covering of the decision region, while an increasing number of hyperboxes yields a finer approximation of the decision region. We also report the resulting regrets for all three methods (DR, DM, and IPS) in \Cref{app:performance_low_dimensional}, in which the doubly robust approach performs best. Thus, we restrict the following experiments to the use of the doubly robust method. This goes in line with earlier research \cite{Bennett.2020}.

\noindent\textbf{Regret analysis.} In order to compare the performance of IOPL against our baselines, we generate 4-dimensional patient covariates $X_0,X_1,X_2,X_3\sim U[-1,1]$, a treatment assignment which is independent of $X$ with $\Prb{T=1}=\nicefrac{1}{2}$, and an outcome $Y\mid X, T\sim \mathcal{N}(m(X, T), \sigma^2)$ where  $m(X,T)=\max(X_2+X_3,0)+\nicefrac{1}{2} \ T \ \text{sign}(\prod_{i=0}^3 X_i)$, and $\sigma=0.1$. The experimental setup is based on \citet{Athey.2021}. See \Cref{app:regret_analysis} for details.

\begin{figure}
	\caption{Regret analysis}
	\begin{center}
		\includegraphics*[scale=0.12]{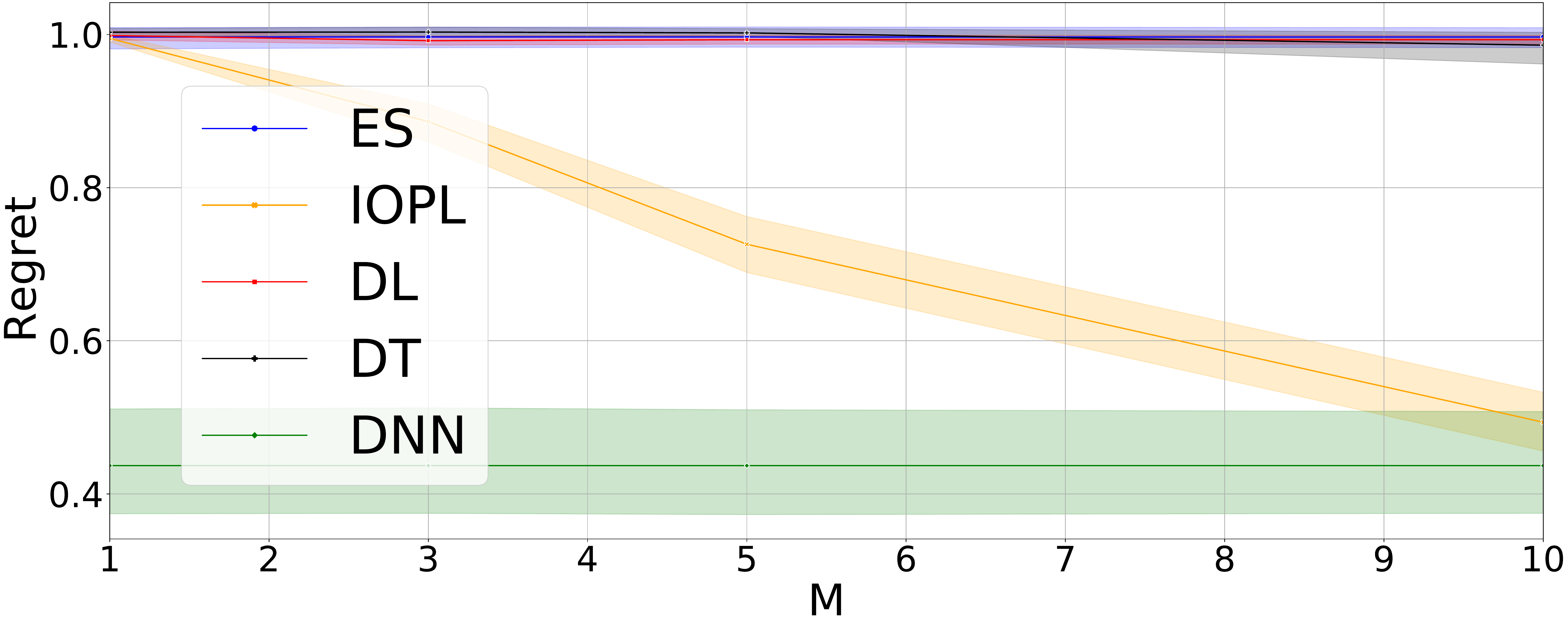}
	\end{center}
	{\scriptsize{\emph{Notes}. We sample 10 different datasets using the described data-generating processes. For IOPL, DL, and DT, we vary the parameter $M$ in $\{1,3,5,10\}$; see \Cref{app:baselines} for details. The shaded areas are the 95\% confidence intervals.}}
	\label{fig:decision_lists}
\end{figure}

The results are reported in \Cref{fig:decision_lists}. First, IOPL outperforms all baselines from interpretable off-policy learning. Second, the decision list and tree-based baselines perform only slightly better than the eligibility score, \ie, the linear baseline. This is in line with the fact that both do not offer theoretical approximation guarantees. For additional experiments with a simpler data generating process, see \Cref{app:additional_experiments}. Third, for larger values of $M$, IOPL yields a regret that is on par with the ``black box'' method (DNN).

\subsection{Real-World Clinical Data}
\label{sec:clinical_data}

We further assess the interpretability of our policies in a user study based on real-world clinical data. We draw upon the AIDS Clinical Trial Group (ACTG) study 175 \cite{Hammer.1996}. The ACTG 175 study assigned treatments randomly to patients with human immunodeficiency virus (HIV) type 1. We consider two treatment arms: (i)~both zidovudine (ZDV) and zalcitabine (ZAL) ($T=1$) vs. (ii)~ZDV only ($T=-1$). This corresponds to $n=1,056$ patients. We consider $d=12$ covariates similar to \cite{Lu.2013}. Further details on the ACTG study are in \Cref{app:actg_study}. We trained ES, as well as DT and IOPL, with $M$ varying in $\{1,3,5\}$. For $M=5$, IOPL returned the policy in Figure~\labelcref{fig:example_policy}. As we use real clinical data, we do not have access to the true policy value, which impedes a regret analysis in this setting. For the sake of completeness, the empirical policy values can be found in \Cref{app:empirical_policy_values_clinical_data}.

\noindent\textbf{User study.} In a user study, we asked 10 practitioners to rate the interpretability of our policy and the baseline policies on a scale between 0 (black box) and 10 (fully transparent) and whether they would use our policies in practice (see \Cref{app:user_study}). All but one practitioner perceived our policies as interpretable (\ie, a rating larger than 5, which denotes neutral). Furthermore, 9 out of 10 would consider using our treatment decisions in practice. This demonstrates that our policies fulfill interpretability demands from clinical practice. A detailed comparison to our baselines can be found in \Cref{app:user_study}.

\section{Conclusion}
\label{sec:conclusion}

We propose IOPL, a novel algorithm for interpretable off-policy learning. It is based on a policy class $\Pi_{H}^M$ that assumes a representation of the decision region as a union of hyperboxes. As a result, our treatment decisions are in disjunctive normal form. To optimize over $\Pi_{H}^M$, we develop a tailored branch-and-price framework that allows for efficient training. In addition, we prove that our policy class is sufficiently rich and thus flexible enough to approximate any measurable policy arbitrarily well.

\section{Acknowledgements}
We acknowledge funding from the Swiss National Science Foundation (Grant 186932).

\bibliography{literature}
\bibliographystyle{icml2022}

\newpage
\appendix
\onecolumn

\section{Interpretability of IOPL}
\label{app:explainability}

In this section, we review IOPL in terms of interpretability with respect to the \emph{AAAI 2021 tutorial} on ``Explaining Machine Learning Predictions: State-of-the-art, Challenges, and Opportunities''.\footnote{AAAI 2021 tutorial on ``Explaining Machine Learning Predictions: State-of-the-art, Challenges, and Opportunities'', \url{https://explainml-tutorial.github.io/aaai21}, last accessed 01/25/22.}

\begin{itemize}
\item\textbf{Interpretability.} IOPL can be viewed as an inherently interpretable predictive model. This is in contrast to post hoc explainability approaches, where the goal is to explain decisions from black box models in retrospect to benefit from their predictive performance. To the contrary, IOPL offers an interpretable machine learning approach with universal approximation guarantees itself. That is, black box models usually offer better performance, however finding an interpretable representation afterwards still goes along with an increase in the regret due to a lack of flexibility (for instance, when explaining neural networks with fixed depth decision trees). As a remedy, our approach combines both worlds with an easy to use and understandable parameter $M$, which controls the degree of interpretability.

\item\textbf{Global vs. local explainability.} As already mentioned in the main paper, IOPL offers policies in disjunctive normal form that allow for global explainability. That is, a comprehensible representation of the treatment groups as in Figure~\labelcref{fig:example_policy} allows clinical experts to critically assess the learned policy, \ie, the complete behavior of the model. If the parameter $M$ is set to very large numbers, a compact representation as in Figure~\labelcref{fig:example_policy} might not be possible anymore. Nevertheless, IOPL still offers local explainability, as individual predictions can be explained by displaying the single hyperbox that led to the corresponding decision. In case that one patient fulfills multiple conditions, \ie, the covariates lie in multiple hyperboxes, either all or the one with the largest volume can be considered. In that way, clinical practitioners can still review if the treatment decision is made on a reasonable basis.

\item\textbf{Utility.} IOPL yields policies that can help to critically assess the behavior of the model in the following ways: (i)~It allows for debugging as medical professionals are offered a comprehensible representation of the model behavior. (ii)~Biases can be detected and ruled out, as our policies allow to quickly recognize if, for instance, only men a treated with a certain medication. (iii)~The treatment decisions of IOPL are completely transparent and, hence, allows medical practitioners to decide when to trust the model predictions and when they are suitable for deployment.
\end{itemize}

\newpage
\section{Details on Binary Classification Reformulation}
\label{app:classification_reformulation}

We now show that problem \labelcref{eq:policy_value_optimization} and problem \labelcref{eq:policy_value_optimization_psi} are equivalent. For this, we proceed as follows. 

First, we recognize that \labelcref{eq:policy_value_optimization} is equivalent to 
\begin{align}
\pi^{\text{opt}} \in \argmin_{\pi\in\Pi}{V(\pi)} - \frac{1}{2}\E[Y(+1)-Y(-1)],
\end{align}
since we only subtract a constant. Similar to \cite{Bennett.2020}, we see that the above problem is equivalent to
\begin{align}
\pi^{\text{opt}} \in \argmin_{\pi\in\Pi}{\E[\psi \ \pi(X)]},
\end{align}
where $\psi$ corresponds to one of the three standard methods for policy learning: direct method $\psi^{\mathrm{DM}}$, inverse propensity score weighted method $\psi^{\mathrm{IPS}}$, and doubly robust method $\psi^{\mathrm{DR}}$ defined as
\begin{align}
\psi^{\mathrm{DM}}=\mu_{1}(X)-\mu_{-1}(X), \qquad \hfill \psi^{\mathrm{IPS}}=\frac{TY}{e_T(X)}, \qquad \hfill \psi^{\mathrm{DR}}=\psi^{\mathrm{DM}}+\psi^{\mathrm{IPS}}-\frac{T\mu_T(X)}{e_T(X)}.
\end{align}
Now, we have that
\begin{align}
\E[\psi \ \pi(X)]=\E[\psi T^2 \pi(X)]=\E[-\psi T (1-T\pi(X))+\psi T],
\end{align}
and
\begin{align}
1-T\pi(X)=2I(T\neq \pi(X)).
\end{align}
Hence,
\begin{align}
\E[\psi \ \pi(X)]=\E[-2T\psi I(T\neq \pi(X))]+\E[\psi T],
\end{align}
and thus
\begin{align}
\pi^{\text{opt}} \in \argmin_{\pi\in\Pi}{\E[\psi \ \pi(X)]}\Leftrightarrow
\pi^{\text{opt}} \in \argmin_{\pi\in\Pi}{\E[-T\psi I(T\neq \pi(X))]},
\end{align}
which proves the claim as $-T\psi$ with $\psi$ given as above corresponds exactly to the definitions of $\psi$ in the main paper.

\newpage
\section{Equivalence of \MILP and \labelcref{eq:empirical_policy_value}}
\label{app:equivalence_MILP}
The empirical binary classification problem~\labelcref{eq:empirical_policy_value} is given by
\begin{align}
\label{eq:EBCP_detailed}
\min\limits_{\pi\in\Pi} \frac{1}{n}\sum_{i=1}^n \psi_i I(T_i\neq \pi(X_i)).
\end{align}
By setting $\Pi=\Pi_{H}^M$ and $\xi_i=I(T_i\neq \pi_{\mathcal{D}}(X_i))$ in problem~\labelcref{eq:EBCP_detailed}, we have to model: (i)~the form of the policy $\pi_{\mathcal{D}}$, \ie, which of the $N$ hyperboxes belongs to the policy and (ii)~the indicator variables $\xi_i$ given a specific policy $\pi_{\mathcal{D}}$, \ie, a specific choice of hyperboxes. 

First, the objective becomes
\begin{align}
\label{eq:EBCP_detailed_xi}
\min\limits_{s,\xi} \frac{1}{n}\sum_{i=1}^n \psi_i \xi_i.
\end{align}

For (i), we introduce binary decision variables $s_j\in\{0,1\}$ for $j\in\{1,\dots,N\}$ indicating whether or not box $j$ is used in the optimal policy. To ensure that the maximal number of hyperboxes $M$ is not exceeded, we use the constraint
\begin{align}
\sum\limits_{j=1}^N s_j\le M,
\end{align}
which encodes that the number of used hyperboxes is bounded by $M$.

For (ii), we proceed via a case distinction on the samples $i$.

\underline{Case 1}: $i\in I_1\cap\mathcal{P}$\\
In this case, patient $i$ was treated, \ie, $T_i=1$, and the corresponding weight $\psi_i>0$. From $T_i=1$, we get that
\begin{align}
\label{eq:xi_definition}
\xi_i=I(1\neq\pi_{\mathcal{D}}(X_i))=\begin{cases}1, & \text{ if } \pi_{\mathcal{D}}(X_i)=-1,\\0, & \text{ if } \pi_{\mathcal{D}}(X_i)=1.\end{cases}
\end{align}
Furthermore, from $\psi_i>0$, we get that, in an optimal solution, $\xi_i$ will strive to zero whenever possible (see objective in \labelcref{eq:EBCP_detailed_xi}). Thus, we need a constraint that forces $\xi_i$ to one if patient $i$ is not treated according to the optimal policy to fulfill \labelcref{eq:xi_definition}, and is switched off otherwise to allow $\xi_i$ to strive to zero. This is exactly encoded in the constraint
\begin{align}
\label{eq:constraint_expl}
\xi_i+\sum\nolimits_{j\in\mathcal{K}_i}s_j\ge 1.
\end{align}
Note that $\mathcal{K}_i$ is the set of all indices of hyperboxes containing the covariates of patient $i$. That is, if on the one hand no hyperbox that contains the covariates of patient $i$ is selected we get
\begin{align}
\sum\nolimits_{j\in\mathcal{K}_i}s_j=0,
\end{align}
and, thus, constraint \labelcref{eq:constraint_expl} becomes $\xi_i\ge1$, yielding $\xi_i=1$. If, on the other hand, a hyperbox is selected that contains the covariates of patient $i$, we get
\begin{align}
\sum\nolimits_{j\in\mathcal{K}_i}s_j\ge 1,
\end{align}
and, thus, that constraint \labelcref{eq:constraint_expl} is fulfilled for any $\xi_i$, yielding $\xi_i=0$.

\underline{Case 2}: $i\in I_1\cap\mathcal{N}$\\
In this case, $\xi_i$ is again given by \labelcref{eq:xi_definition}. However, now we have that $\psi_i<0$, and, thus, we get that, in an optimal solution, $\xi_i$ will strive to one whenever possible (see objective in \labelcref{eq:EBCP_detailed_xi}). Thus, we need a constraint that forces $\xi_i$ to zero if patient $i$ is treated according to the optimal policy to fulfill \labelcref{eq:xi_definition}, and is switched off otherwise to allow $\xi_i$ to strive to one. This is encoded in the constraint
\begin{align}
\label{eq:constraint_expl_2}
\xi_i\le 1-s_j \text{ for } j\in\mathcal{K}_i.
\end{align}
Note that if on the one hand any hyperbox containing the covariates of patient $i$ is selected constraint \labelcref{eq:constraint_expl_2} enforces $\xi_i\le0$ and, thus, $\xi_i=0$. If, on the other hand, no hyperbox that contains the covariates of patient $i$ is selected, constraint \labelcref{eq:constraint_expl_2} becomes $\xi_i\le1$ and is thus fulfilled for any $\xi_i$, yielding $\xi_i=1$.

The cases $i\in I_{-1}\cap\mathcal{P}$ and $i\in I_{-1}\cap\mathcal{N}$ follow the exact same logic with the difference that we have
\begin{align}
\xi_i=I(-1\neq\pi_{\mathcal{D}}(X_i))=\begin{cases}0, & \text{ if } \pi_{\mathcal{D}}(X_i)=-1,\\1, & \text{ if } \pi_{\mathcal{D}}(X_i)=1,\end{cases}
\end{align}
in this case.

Again, we note that the integrality of $\xi_i$ follows by the integrality constraint of $s_j$. That is, due to the objective and our constraints, the values of $\xi_i$ always strive to the extreme points of the interval $[0,1]$. Hence, the constraint $\xi_i\in[0,1]$ is sufficient and reduces the number of integer variables significantly compared to $\xi_i\in\{0,1\}$ for all $i\in\{1,\dots,n\}$.

\newpage
\section{Universal Approximation Theorem}
\label{app:universal_approximation_theorem}

\subsection{Proof of \Cref{thm:universal_approximation_theorem}}
First, we observe that
\begin{align}
\pi^\ast(x)=\begin{cases}
1 & \text{ if } \ x\in\mathcal{A},\\
-1 & \text{ if } \ x\in\mathcal{A}^\complement.\\
\end{cases}
\end{align}
As $\pi^\ast$ is assumed to be Lebesgue measurable, we have that the decision region $\mathcal{A}$ is Lebesgue measurable with $\lambda(\mathcal{A})<\infty$, where $\lambda$ denotes the Lebesgue measure. Note that this implies $\pi^\ast\in L^p(\mathcal{X})$ for all $1\le p<\infty$ as $\pi^\ast$ is a simple function.

Now, from Theorem 3.4 in \cite{Stein.2009}, we know that, for every $\epsilon>0$, there exists a finite union of boxes $\tilde{\mathcal{A}}=\bigcup\limits_{j=1}^{m} \tilde{S}^j$ such that
\begin{align}
\lambda(\mathcal{A}\triangle\tilde{\mathcal{A}})\le \frac{\epsilon^p}{2^{p+1}},
\end{align}
where $\mathcal{A}\triangle\tilde{\mathcal{A}}$ denotes the symmetric difference between two sets, \ie, $\mathcal{A}\triangle\tilde{\mathcal{A}}=\mathcal{A}\setminus\tilde{\mathcal{A}}\cup\tilde{\mathcal{A}}\setminus\mathcal{A}$.

For the next part of the proof, we need a technical lemma (the proof is given in \Cref{app:techincal_lemma}) given as follows.
\begin{lemma}[Technical lemma]
	\label{lemma:technical_lemma}
	For every $\delta\in(0,1)$, $\epsilon>0$, and box $\tilde{S}$, there exists a sample size $n_{\delta,\epsilon}\in\mathbb{N}$ and an i.i.d. sample $\{(X_i,T_i,Y_i)\}_{i=1}^{n_{\delta,\epsilon}}$ with $(X_i,T_i,Y_i)\sim(X,T,Y)$, such that there exists a box $\tilde{\tilde{S}}$ generated by a subset $\{X_i\}_{i\in\mathcal{I}}\subseteq\{X_i\}_{i=1}^{n_{\delta,\epsilon}}$, \ie,
	\begin{align}
	\tilde{\tilde{S}}=[\tilde{\tilde{l}}_{1},\tilde{\tilde{u}}_{1}]\times[\tilde{\tilde{l}}_{2},\tilde{\tilde{u}}_{2}]\times\dots\times[\tilde{\tilde{l}}_{d},\tilde{\tilde{u}}_{d}]
	\ \text{ with } \ \tilde{\tilde{l}}_{t}=\min_{i\in \mathcal{I}} X_{i,t} \ \text{ and } \ \tilde{\tilde{u}}_{t}=\max_{i\in \mathcal{I}} X_{i,t},
	\end{align}
	that fulfills
	\begin{align}
	\lambda(\tilde{S}\triangle\tilde{\tilde{S}})\le\frac{\epsilon^p}{2^{p+1}m},
	\end{align}
	with probability at least $1-\delta$.
\end{lemma}
By iteratively applying Lemma~\ref{lemma:technical_lemma}, there exists an $n_{\delta,\epsilon}\in\mathbb{N}$ large enough and boxes $\tilde{\tilde{S}}^j$, such that
\begin{align}
\lambda(\tilde{S}^j\triangle\tilde{\tilde{S}}^j)\le\frac{\epsilon^p}{2^{p+1}m},
\end{align}
for all $j\in\{1,\dots,m\}$ with probability at least $1-\delta$. Now let $\tilde{\tilde{\mathcal{A}}}=\bigcup\limits_{j=1}^{m} \tilde{\tilde{S}}^j$. We yield
\begin{align}
\lambda(\mathcal{A}\triangle\tilde{\tilde{\mathcal{A}}})&=\lambda\left(\mathcal{A}\triangle\bigcup\limits_{j=1}^{m} \tilde{\tilde{S}}^j\right)\\
&\le \lambda\left(\mathcal{A}\triangle\bigcup\limits_{j=1}^{m} \tilde{S}^j\right)+\lambda\left(\bigcup\limits_{j=1}^{m} \tilde{S}^j\triangle\bigcup\limits_{j=1}^{m} \tilde{\tilde{S}}^j\right)\label{eq:explanation_1}\\
&\le\frac{\epsilon^p}{2^{p+1}}+\sum\limits_{j=1}^m \lambda\left(\tilde{S}\triangle\tilde{\tilde{S}}\right)\label{eq:explanation_2}\\
&\le\frac{\epsilon^p}{2^p},
\end{align}
where \Cref{eq:explanation_1} follows by a generalized triangle inequality for the symmetric difference and \Cref{eq:explanation_2} follows by the inclusion property of the symmetric difference of unions and the sub-additivity of the Lebesgue measure.

Now, we can write $\pi^\ast=\chi_{\mathcal{A}}-\chi_{\mathcal{A}^\complement}$, where $\chi$ denotes the characteristic function and define $\pi_{\mathcal{D}^\ast}=\chi_{\tilde{\tilde{\mathcal{A}}}}-\chi_{\tilde{\tilde{\mathcal{A}}}^\complement}$ by setting $\mathcal{D}^\ast$ appropriately and set $M=\lvert\mathcal{D}^\ast\rvert$. We then yield
\begin{align}
\lVert \pi^\ast - \pi_{\mathcal{D}^\ast} \rVert_p&=\lVert \chi_{\mathcal{A}}-\chi_{\tilde{\tilde{\mathcal{A}}}}+\chi_{\tilde{\tilde{\mathcal{A}}}^\complement}-\chi_{\mathcal{A}^\complement} \rVert_p\\
&\le \lVert \chi_{\mathcal{A}}-\chi_{\tilde{\tilde{\mathcal{A}}}}\rVert_p+\lVert\chi_{\tilde{\tilde{\mathcal{A}}}^\complement}-\chi_{\mathcal{A}^\complement} \rVert_p.
\end{align}
and
\begin{align}
\lVert \chi_{\mathcal{A}}-\chi_{\tilde{\tilde{\mathcal{A}}}}\rVert_p^p&=\int\limits_{\mathcal{X}} \lvert \chi_{\mathcal{A}}-\chi_{\tilde{\tilde{\mathcal{A}}}}\lvert^p \dd{x}\\
&=\int\limits_{\mathcal{A}\setminus\tilde{\tilde{\mathcal{A}}}} 1 \dd{x}+\int\limits_{\tilde{\tilde{\mathcal{A}}}\setminus\mathcal{A}} 1 \dd{x} = \lambda(\mathcal{A}\triangle\tilde{\tilde{\mathcal{A}}}).
\end{align}
Analogously, we get
\begin{align}
\lVert \chi_{\mathcal{A}^\complement}-\chi_{\tilde{\tilde{\mathcal{A}}}^\complement}\rVert_p^p= \lambda(\mathcal{A}^\complement\triangle\tilde{\tilde{\mathcal{A}}}^\complement)=\lambda(\mathcal{A}\triangle\tilde{\tilde{\mathcal{A}}}).
\end{align}
Combining the above we thus get
\begin{align}
\lVert \pi^\ast - \pi_{\mathcal{D}^\ast} \rVert_p\le 2\lambda(\mathcal{A}\triangle\tilde{\tilde{\mathcal{A}}})^{\frac{1}{p}}\le\epsilon,
\end{align}
which proves the claim.

\subsection{Proof of Lemma~\ref{lemma:technical_lemma}}
\label{app:techincal_lemma}

Let $\mathcal{V}$ be the set of vertices of $\tilde S$. W.l.o.g we assume that $\text{supp}(X)=\mathcal{X}$, otherwise we can redefine $\mathcal{X}$ to contain only $X$ with positive probability. Now, we have that $\Prb{\lVert X-v\rVert<\tilde \delta}>0$ for all $v\in\mathcal{V}$ and $\tilde \delta>0$. That is, when sampling long enough, we will reach a sample with $X_{i_v}\in B_{\tilde\delta}(v)$ for all vertices $v\in\mathcal{V}$ and for indices $i_v\in\{1,\dots,n_{\delta,\epsilon}\}$. By choosing $\tilde\delta$ and the sample size $n_{\delta,\epsilon}$ appropriately, the claim follows.

\newpage
\section{Asymptotic Estimation Properties}
\label{app:asymptotic_properties}

\subsection{Proof of \Cref{thm:rademacher_complexity}}
To prove \Cref{eq:rademacher_complexity}, we make use of Massart's Lemma\footnote{see for instance Theorem 4.3 in \url{http://www.cs.toronto.edu/~rjliao/notes/Notes_on_Rademacher_Complexity.pdf}, last accessed 01/25/22.}.
\begin{lemma}
	\label{lemma:massart}
	Assume $\lvert \Pi\rvert$ is finite. Let $\{(X_i)\}_{i=1}^n$ be a random i.i.d. sample, and let
	\begin{align}
	B=\max\limits_{\pi\in\Pi} \left(\sum\limits_{i=1}^{n}\pi(X_i)^2\right)^{\frac{1}{2}},
	\end{align}
	then
	\begin{align}
	R_{n}(\Pi)\le\frac{B\sqrt{2\log{\lvert \Pi\rvert}}}{n}.
	\end{align}
\end{lemma}
First, as $\pi\in\Pi_H^M$ are binary valued functions we immediately have that $B=\sqrt{n}$ in our case. Second, we make the following observation
\begin{align}
\lvert \Pi_H^M\rvert = \sum\limits_{k=0}^M \binom{2^n}{k} \le \left(\frac{e 2^n}{M}\right)^M.
\end{align}
By applying Lemma~\ref{lemma:massart}, we yield
\begin{align}
R_{n}(\Pi_H^M) &\le \sqrt{\frac{2\log{\lvert \Pi_H^M\rvert}}{n}}\le \sqrt{\frac{2M(1+n\log{2}-\log{M})}{n}}\\
&\le \sqrt{2M + 2M\log{2}}= C\sqrt{2M},
\end{align}
where $C=\sqrt{1+\log{2}}$.

\subsection{Proof of Corollary~\ref{cor:asymptotic_behavior}}

The proof follows the elaborations in Section 2 in \cite{Bennett.2020}. For the sake of completeness, we restate the proof in our setting and notation. \cite{Athey.2017}, \cite{Kitagawa.2018}, and \cite{Zhou.2018} gave bounds of the form
\begin{align}
\sup_{\pi\in\Pi}\lvert \mathcal{J}(\pi)-\mathcal{J}_n(\pi) \rvert=\mathcal{O}_p(\nicefrac{1}{\sqrt{n}}),
\end{align}
given that the policy class $\Pi$ has bounded complexity. Hence, by \Cref{thm:rademacher_complexity}, one yields
\begin{align}
\mathcal{J}(\pi^\mathrm{opt})-\mathcal{J}(\hat\pi^\mathrm{opt}_n)\le
\mathcal{J}(\pi^\mathrm{opt})-\mathcal{J}(\hat\pi^\mathrm{opt}_n)+\mathcal{J}_n(\hat\pi^\mathrm{opt}_n)-\mathcal{J}_n(\pi^\mathrm{opt})\le 2\sup_{\pi\in\Pi}\lvert \mathcal{J}(\pi)-\mathcal{J}_n(\pi) \rvert=\mathcal{O}_p(\nicefrac{1}{\sqrt{n}}).
\end{align}

\newpage
\section{Expert-Informed Penalty Method}
\label{app:algorithmic_calibration}

To estimate the number of hyperboxes, \ie, the number of rules, in a fully automated manner we propose a penalty method for \MILP in the following. To do so, we add a penalty term $\omega \ \sum_{j=1}^N s_j$ to the objective function. Thereby, the penalty parameter $\omega>0$ can be set by a clinical practitioner and describes the value one is willing to sacrifice in the policy value in order to have one fewer rule in the resulting model \cite{Rudin.2019}. In our case study with real-world data, for instance, $Y$ decodes the increase in the CD4 cell count after 20 weeks. That is, setting for example $\omega=25$, one is willing to sacrifice a decrease of the population wide CD4 cell count by 25 cells/$\text{mm}^3$ for every rule that is eliminated in the final model.\footnote{Any scaling of $Y$ must be taken into account, of course.} After that, the parameter $M$ can be set to an absolute maximal number, \eg, $M=100$, and IOPL determines the number of rules automatically.

For our expert-informed penalty method, the framework of IOPL is adapted as follows: (i)~we alter \MILP by adding the discussed penalty term and (ii)~we update the reduced cost for the pricing problem accordingly. All other steps in \Cref{alg:branch_and_bound} remain unchanged.

For (i), \MILP becomes
\begin{align}
&\min \frac{1}{n}\sum\limits_{i=1}^n \psi_i \xi_i + \omega \ \sum\limits_{j=1}^N s_j&&\\
\textrm{s.t.} \ & \xi_i+\sum\nolimits_{j\in\mathcal{K}_i}s_j\ge 1 &&\hspace{-1em}\text{ for } i\in I_1\cap\mathcal{P},\\
&\xi_i\ge s_j &&\hspace{-1em}\text{ for } i\in I_{-1}\cap\mathcal{P} \text{ and } j\in\mathcal{K}_i,\\
&\xi_i\le 1-s_j &&\hspace{-1em}\text{ for } i\in I_1\cap\mathcal{N} \text{ and } j\in\mathcal{K}_i,\\
&\xi_i\le\sum\nolimits_{j\in\mathcal{K}_i}s_j &&\hspace{-1em}\text{ for } i\in I_{-1}\cap\mathcal{N},\\
&\sum\nolimits_{j=1}^N s_j\le M,&&\\
&s_j\in\{0,1\}, \xi_i\in[0,1].
\end{align}
Note that the penalized objective is still linear. Hence, the computational complexities for the relaxed problem \RMILP and the restricted \MILP remain the same.

For (ii), we have to update the reduced cost accordingly. Based on the form of the penalty term it changes merely by an additional additive constant $\omega$, \ie, the reduced cost becomes
\begin{align}
\begin{split}
&\omega-\sum\limits_{i\in I_1\cap\mathcal{P}} \mu_i^1\delta_i+\sum\limits_{i\in I_{-1}\cap\mathcal{P}} \Bigg(\sum\limits_{j\in\mathcal{K}_i}\mu_{i,j}^2\Bigg)\delta_i\\
&+\sum\limits_{i\in I_1\cap\mathcal{N}} \Bigg(\sum\limits_{j\in\mathcal{K}_i}\mu_{i,j}^3\Bigg)\delta_i
-\sum\limits_{i\in I_{-1}\cap\mathcal{N}} \mu_i^4\delta_i+\lambda.
\end{split}
\end{align}

In summary, the algorithm changes merely in an additional linear term in the objective functions of the relaxed problem \RMILPK{$\mathcal{W}$} and the restricted problem \MILPK{$\mathcal{W}$}. The pricing problem remains unchanged, as we only have to terminate the column generation procedure if the minimal value of the pricing problem is greater or equal to $-\omega$ (instead of zero).

\newpage
\section{Baselines}
\label{app:baselines}

\subsection{Surrogate-Loss Classification}

For the ES and DNN baselines, we follow the literature on off-policy learning and reformulate \labelcref{eq:empirical_policy_value} equivalently via a convex surrogate loss that replaces the discontinuous and non-convex 0-1 loss with the well-established hinge loss \citep{Beygelzimer.2009,Jiang.2019,Zhao.2012}.

\subsubsection{Training}
For training, we first note that a straightforward replacement of the 0-1 loss with the hinge-loss in \labelcref{eq:empirical_policy_value} is not possible. This is because the weights $\psi_i$ can be both positive and negative. Hence, the function
\begin{align}
\frac{1}{n}\sum\limits_{i=1}^n \psi_i \max(1-T_i\pi(X_i),0)
\end{align}
might not be convex. As a remedy, we rewrite \labelcref{eq:empirical_policy_value} as follows. Note that
\begin{align}
\psi_i 2I(T_i\neq \pi(X_i))&=\psi_i(1-T_i\pi(X_i))=\psi_i-\lvert\psi_i\rvert\text{sign}(\psi_i)T_i\pi(X_i)\\
&=\psi_i-\lvert\psi_i\rvert+\lvert\psi_i\rvert-\lvert\psi_i\rvert\text{sign}(\psi_i)T_i\pi(X_i)\\
&=\psi_i-\lvert\psi_i\rvert+\lvert\psi_i\rvert(1-\text{sign}(\psi_i)T_i\pi(X_i))\\
&=\psi_i-\lvert\psi_i\rvert+2\lvert\psi_i\rvert I(\text{sign}(\psi_i)T_i\neq \pi(X_i))
\end{align}
Thus, \labelcref{eq:empirical_policy_value} can be reformulated, using the hinge loss as a surrogate, as
\begin{align}
\min\limits_{\pi\in\Pi} \ \frac{1}{n}\sum\limits_{i=1}^n \lvert\psi_i\rvert \max(1-\text{sign}(\psi_i)T_i\pi(X_i),0).
\end{align}
To control for over-fitting, we add a regularization term $\rho \ \mathrm{Reg}(\pi)$ resulting in
\begin{align}
\label{eq:hinge_loss_surrogate}
\min\limits_{\pi\in\Pi} \ \frac{1}{n}\sum\limits_{i=1}^n \lvert\psi_i\rvert \max(1-\text{sign}(\psi_i)T_i\pi(X_i),0)+\rho \ \mathrm{Reg}(\pi).
\tag{H-EBCP}
\end{align}

\textbf{Linear eligibility score.} For the linear model, we proceed as follows. We set $\pi(X_i)=\langle w,X_i\rangle -b$ and the regularization term is set to $\mathrm{Reg}(\pi)=\lVert w\rVert^2$. Thus, we can rewrite \labelcref{eq:hinge_loss_surrogate} as
\begin{align}
&\min\limits_{w\in\mathbb{R}^d, b\in\mathbb{R}, \xi\in\mathbb{R}^n} \ \frac{1}{n}\sum\limits_{i=1}^n \lvert\psi_i\rvert \ \xi_i + \rho \ \lVert w\rVert^2\\
\textrm{s.t.} \quad  & \xi_i\ge 1-\text{sign}(\psi_i)T_i(\langle w,X_i\rangle-b) \ \text{ for all } i\in\{1,\dots,n\},\\
& \xi_i\ge 0 \ \text{ for all } i\in\{1,\dots,n\}.
\end{align}
The resulting problem is a weighted linear support vector classification problem that is solved via the LinearSVC class of the sklearn python package.

\textbf{Deep neural network.} For our neural network baseline, we set $\mathrm{Reg}$ to the \mbox{$\ell_2$-regularization}. We use PyTorch with a custom loss function and the Adam optimizer to train the neural network parameters. In particular, we use early stopping with a patience of 10 and 20 warmup epochs. The structure of the neural network is summarized in \Cref{tab:NN_configuration}.

\FloatBarrier

\begin{table}[h]
	\caption{Deep Neural Network Structure}
	\label{tab:NN_configuration}
	\centering
	\begin{tabular*}{.5\textwidth}{cccc l}
		\toprule
		\multicolumn{4}{c}{\textbf{Input neurons per layer}}  & \textbf{Activation}\\
		\cmidrule(lr){1-5} 
		\multicolumn{1}{c}{Layer 1} & \multicolumn{1}{c}{Layer 2} & \multicolumn{1}{c}{Layer 3} & \multicolumn{1}{c}{Layer 4} & (output-layer) \\
		\midrule
		$d$ & 128 & 128 & 64 & identity\\
		\bottomrule
	\end{tabular*}
\end{table}

\FloatBarrier

\subsection{Decision Lists}

For our decision list baseline, we follow \cite{Zhang.2018}. That is, we define the set of potential clauses in the decision list as
\begin{align}
\mathcal{R}=&\{\mathcal{X},\{x\in\mathcal{X}: x_{j_1}\le\tau_1\},\{x\in\mathcal{X}: x_{j_1}>\tau_1\},\\
&\{x\in\mathcal{X}: x_{j_1}\le\tau_1 \ \text{and} \ x_{j_2}\le\tau_2\},\\
&\{x\in\mathcal{X}: x_{j_1}\le\tau_1 \ \text{and} \ x_{j_2}>\tau_2\},\\
&\{x\in\mathcal{X}: x_{j_1}>\tau_1 \ \text{and} \ x_{j_2}\le\tau_2\},\\
&\{x\in\mathcal{X}: x_{j_1}>\tau_1 \ \text{and} \ x_{j_2}>\tau_2\}:\\
&1\le j_1<j_2\le d, \tau_1,\tau_2\in\mathbb{R}\},
\end{align}
where $j_1$ and $j_2$ are indices and $\tau_1$ and $\tau_2$ are thresholds. For optimization, we analogously define the optimal estimated decision rule $\hat\pi$ using the outcome function as $x\mapsto \argmin_{t\in\{-1,1\}} \mu_{t}(x)$. The estimation then follows \Cref{alg:DL_optimization} as proposed in \cite{Zhang.2018}.

\begin{algorithm}[H]
	\caption{Estimation of Clauses for Decision List Baseline}
	\label{alg:DL_optimization}
	\scriptsize
	\begin{algorithmic}
		\STATE {\bfseries Input:} Maximal number of clauses $M$, hyperparameter $\xi$, hyperparameter $\eta$
		\STATE {\bfseries Output:} Optimal decision list
		\FOR{$l\in\{1,\dots,M\}$}
		\IF{$l<M$}
		\STATE Compute new clause via
		\begin{align*}
		(R_l,t_l)=&\argmin_{R\in\mathcal{R},t\in\{-1,1\}} \frac{1}{n}\sum\limits_{i=1}^n I(X_{i}\in G_l, X_{i}\in R) \mu_t(X_i)\\
		&+I(X_{i}\in G_l, X_{i}\notin R) \min\{\mu_1(X_i),\mu_{-1}(X_i)\}\\
		&-\xi\left(\frac{1}{n}\sum\limits_{i=1}^n I(X_{i}\in G_l, X_{i}\in R)\right)-\eta\left(2-V(R)\right),
		\end{align*}
		subject to $n^{-1}I(X_{i}\in G_l, X_{i}\in R)>0$, where $G_1=\mathcal{X}$, $G_l=\mathcal{X}\setminus\bigcup\limits_{k<l}R_k$ for $l\ge 2$, and $V(R)\in\{0,1,2\}$ is the number of parameters used to define clause $R$.
		\IF{$R_l=\mathcal{X}$}
			\STATE {\bfseries Return} $((R_1,t_1),\dots,(R_l,t_l))$
		\ENDIF
		\ELSE
		\STATE Compute $(R_M,t_M)=\argmin_{R\in\mathcal{R},t\in\{-1,1\}} \frac{1}{n}\sum\limits_{i=1}^n I(X_{i}\in G_M) \mu_t(X_i)-\eta\left(2-V(R)\right)$.
		\STATE {\bfseries Return} $((R_1,t_1),\dots,(R_M,t_M))$
		\ENDIF
		\ENDFOR
	\end{algorithmic}
\end{algorithm}
If $l=M$ the solution satisfies $V(R_M)=0$, and, hence, $R_M=\mathcal{X}$. The estimated decision lists thus have the form
\begin{align*}
\text{If} \quad &x\in R_1 \quad \text{then} \quad t_1;\\
\text{else if} \quad &x\in R_2 \quad \text{then} \quad t_2;\\
\dots\\
\text{else if} \quad &x\in \mathcal{X} \quad \text{then} \quad t_{\hat M},
\end{align*}
where $\hat M\le M$. For optimizing over $R$ and $t$, \cite{Zhang.2018} proposed an algorithm with $\mathcal{O}(n\log n 2 d)$ operations. In our implementation, we used a brute-force approach. The hyperparameter grids are given in \Cref{tab:hp_grids}.

\subsection{Tree-Based Policy Learner}

For our tree-based baseline, we make use of the freely available python package EconML \cite{econml}. Therein, the idea is to first derive doubly robust estimates of the counterfactual outcomes
\begin{align}
Y^{\mathrm{DR}}_{i,t}=\mu_{t}(X_i)+\frac{Y_i-\mu_{t}(X_i)}{e_{t}(X_i)} 1\{T_i=t\}
\end{align}
and then optimize the policy value in the form
\begin{align}
\sum\limits_{i=1}^n \left(\frac{1}{2}\pi(X_i)+\frac{1}{2}\right)\left(Y^{\mathrm{DR}}_{i,1}-Y^{\mathrm{DR}}_{i,-1}\right).
\end{align}
For this baseline, the parameter $M$ describes the maximal depth of the tree.

\subsection{Hyperparameters}
\label{app:hyperparameters}

For all baselines, we use 80\% of the data for training and 20\% for validation. The hyperparameters are given in \Cref{tab:hp_grids}.

\FloatBarrier

\begin{table}[h]
	\caption{Hyperparameter Grids}
	\label{tab:hp_grids}
	\centering
	\begin{tabular}{l l l}
		\toprule
		\textbf{Model} & \textbf{Hyperparameters} & \textbf{Tuning range}\\
		\midrule
		Linear eligibility score & regularization parameter $\rho$ & $\{10^{-k}: k\in\{0,\dots,9\}\}$\\
		Decision list & hyperparameter $\xi$ & $\{0.1,0.5,1.0\}$\\
		& hyperparameter $\eta$ & $\{0.01,0.1,1.0\}$\\
		Tree-based policy learner & minimum samples leaf & $\{1,3,5\}$\\
		& minimum samples split & $\{5,10,20\}$\\
		Deep neural network & initial learning rate & $\{10^{-1},10^{-2},10^{-3},10^{-4}\}$\\
		& batch size & $\{128,\text{full}\}$\\
		& momentum parameter $\beta_1$ & $\{0.9,0.99\}$\\
		& regularization parameter $\rho$ & $\{10^{-2},10^{-3},10^{-4}\}$\\
		\bottomrule
	\end{tabular}
\end{table}

\FloatBarrier

\subsection{Scaling}

For our baselines, we rescale and shift $\psi_i$ via the formula $\hat\psi_i=\frac{\psi_i-\mu}{\sigma}$, where $\mu$ is the mean value of $\psi$ and $\sigma$ the standard deviation. Note that
\begin{align}
\min\limits_{\pi\in\Pi}\frac{1}{n}\sum\limits_{i=1}^n \hat\psi_i I(T_i\neq \pi(X_i))
\end{align}
is equivalent to \labelcref{eq:empirical_policy_value}.

\newpage
\section{Illustration of IOPL}
\label{app:illustration_of_IOPL}

\FloatBarrier

\subsection{Illustration of Universal Approximation Theorem}
\label{app:universal_approximation_theorem_visualization}

\begin{figure}[h]
	\caption{Illustration of Universal Approximation Theorem}
	\begin{center}
		\begin{tabular}{cccc}
			\subfloat[Basic, $M=1$]{\includegraphics*[scale=0.17]{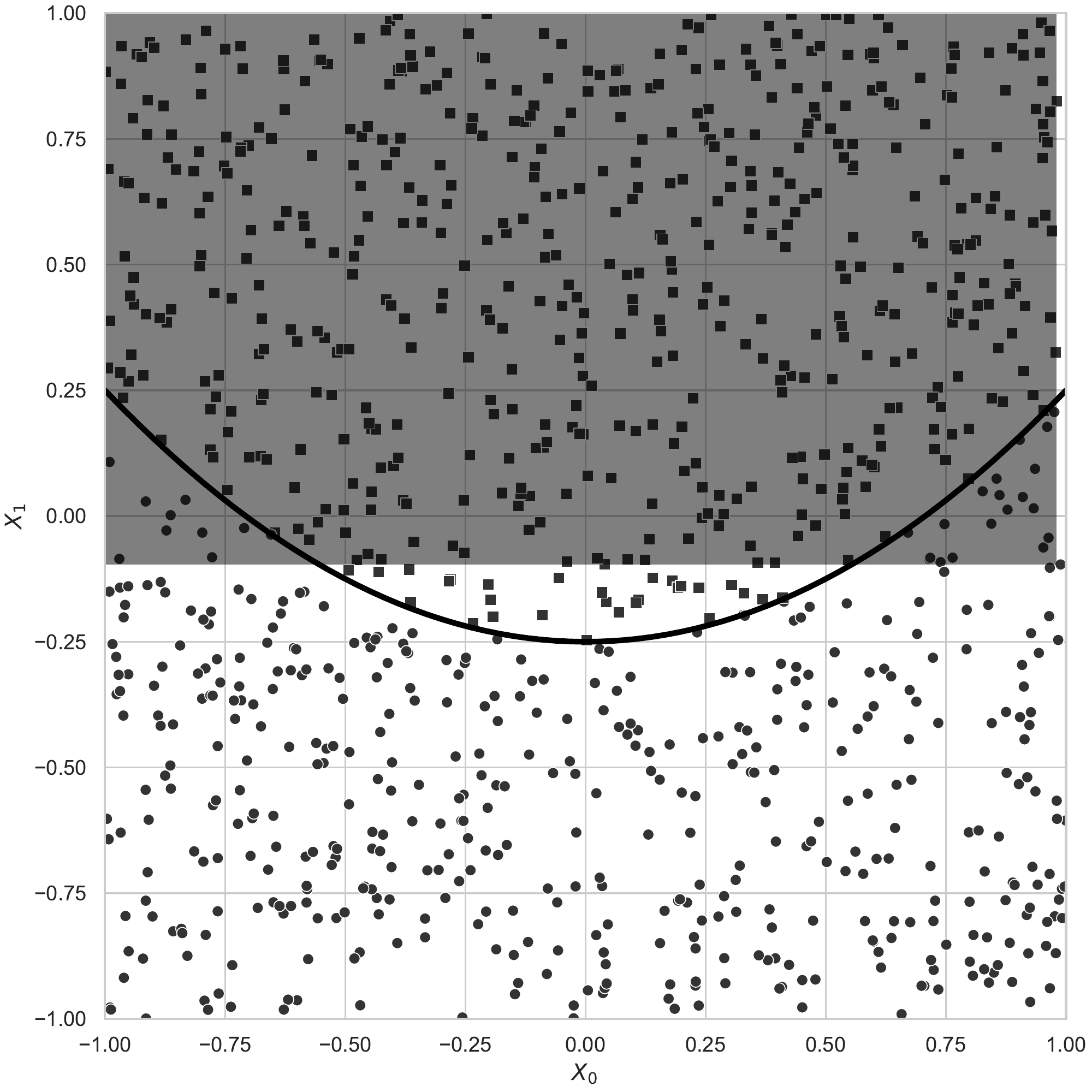}}&
			\subfloat[Basic, $M=3$]{\includegraphics*[scale=0.17]{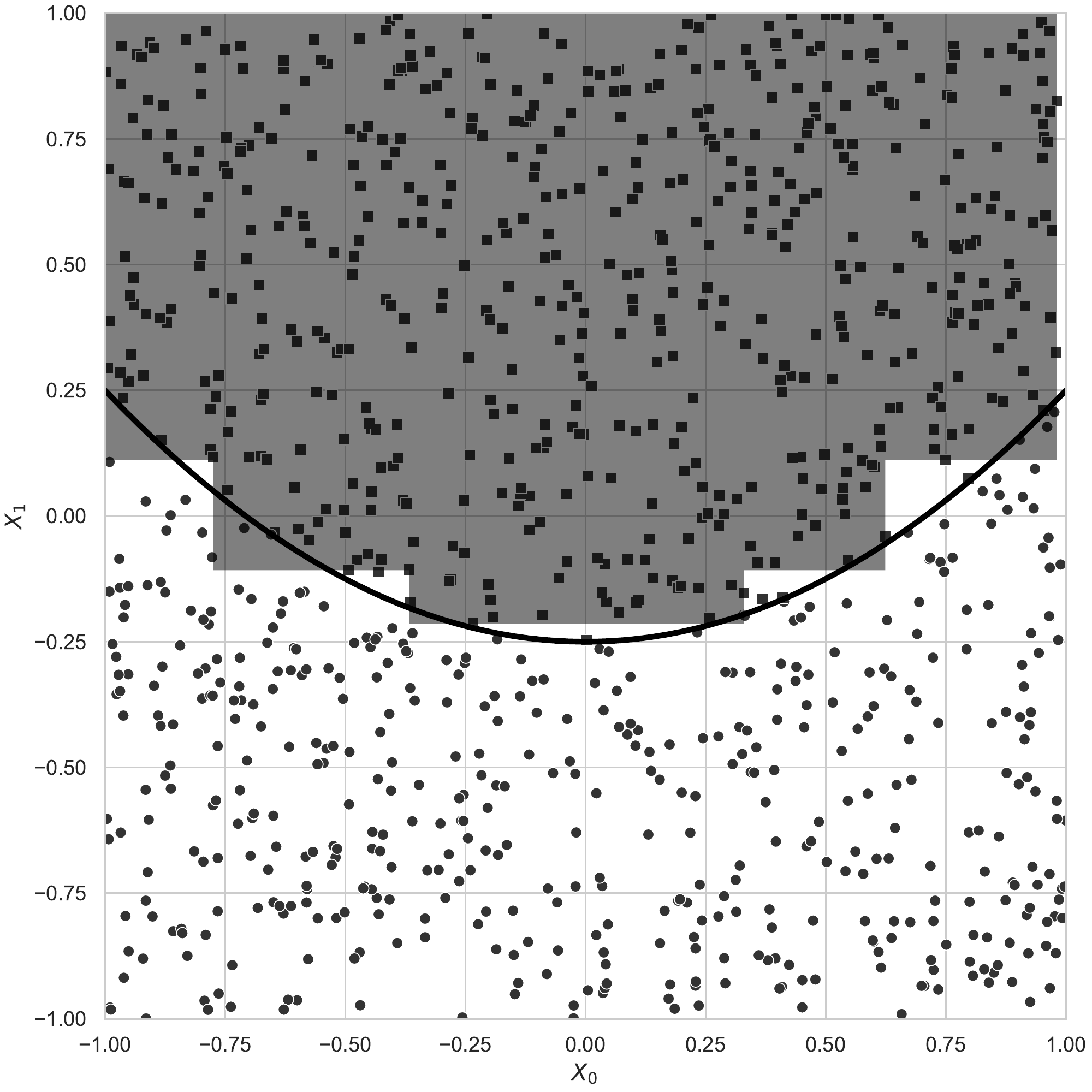}}&
			\subfloat[Basic, $M=5$]{\includegraphics*[scale=0.17]{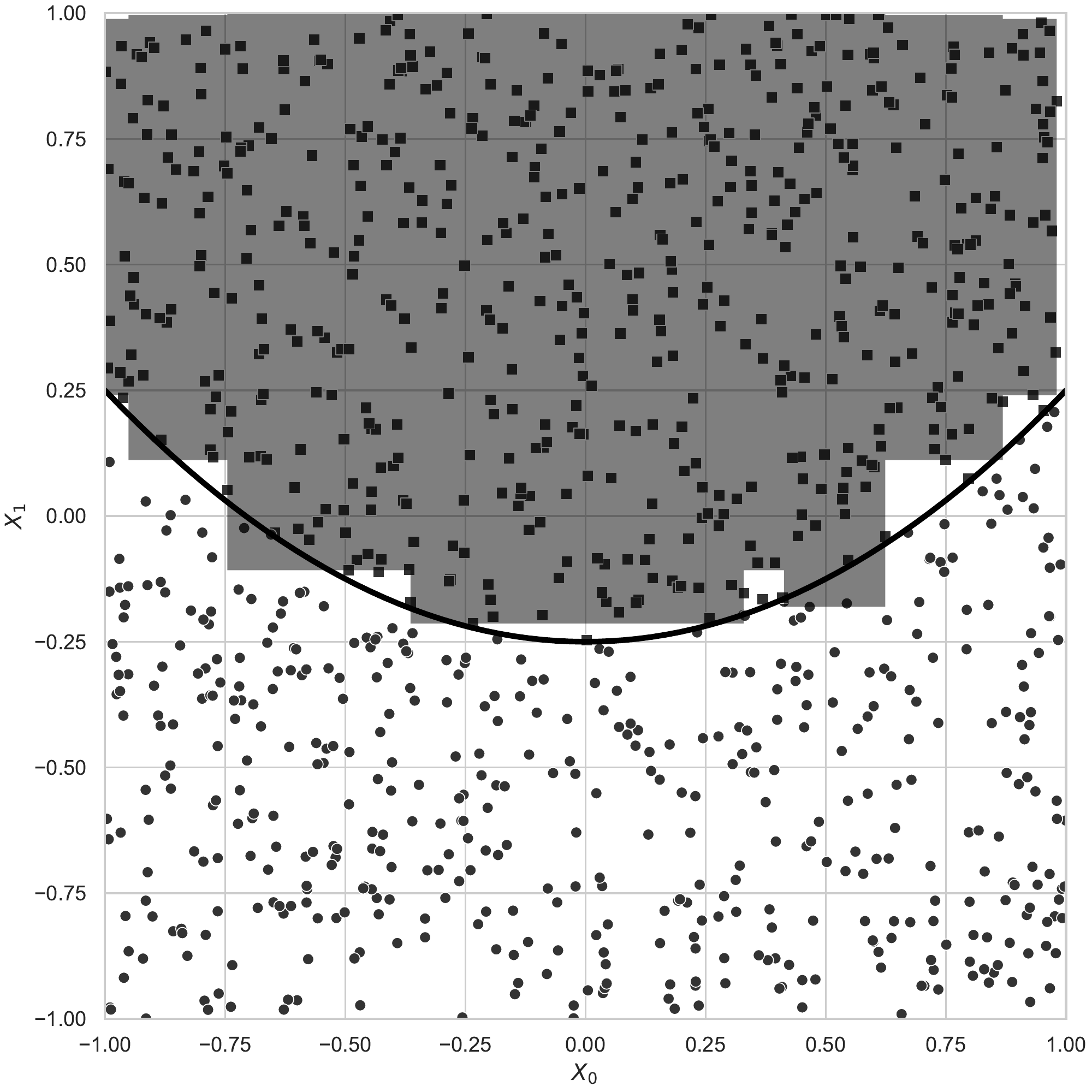}\label{fig:basic_M_5}}&
			\subfloat[Basic, $M=10$]{\includegraphics*[scale=0.17]{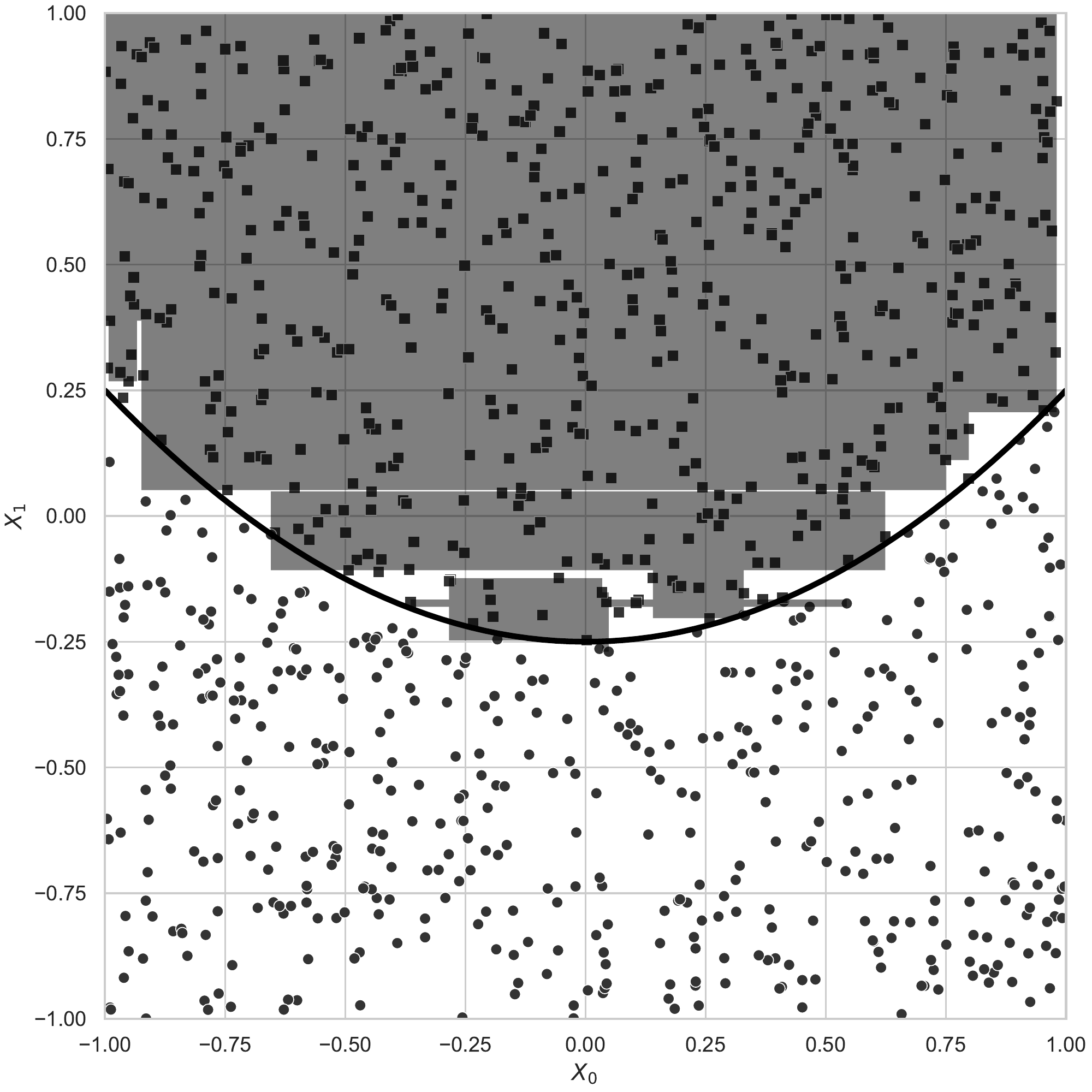}\label{fig:basic_M_10}}\\
			
			\subfloat[Complex, $M=1$]{\includegraphics*[scale=0.17]{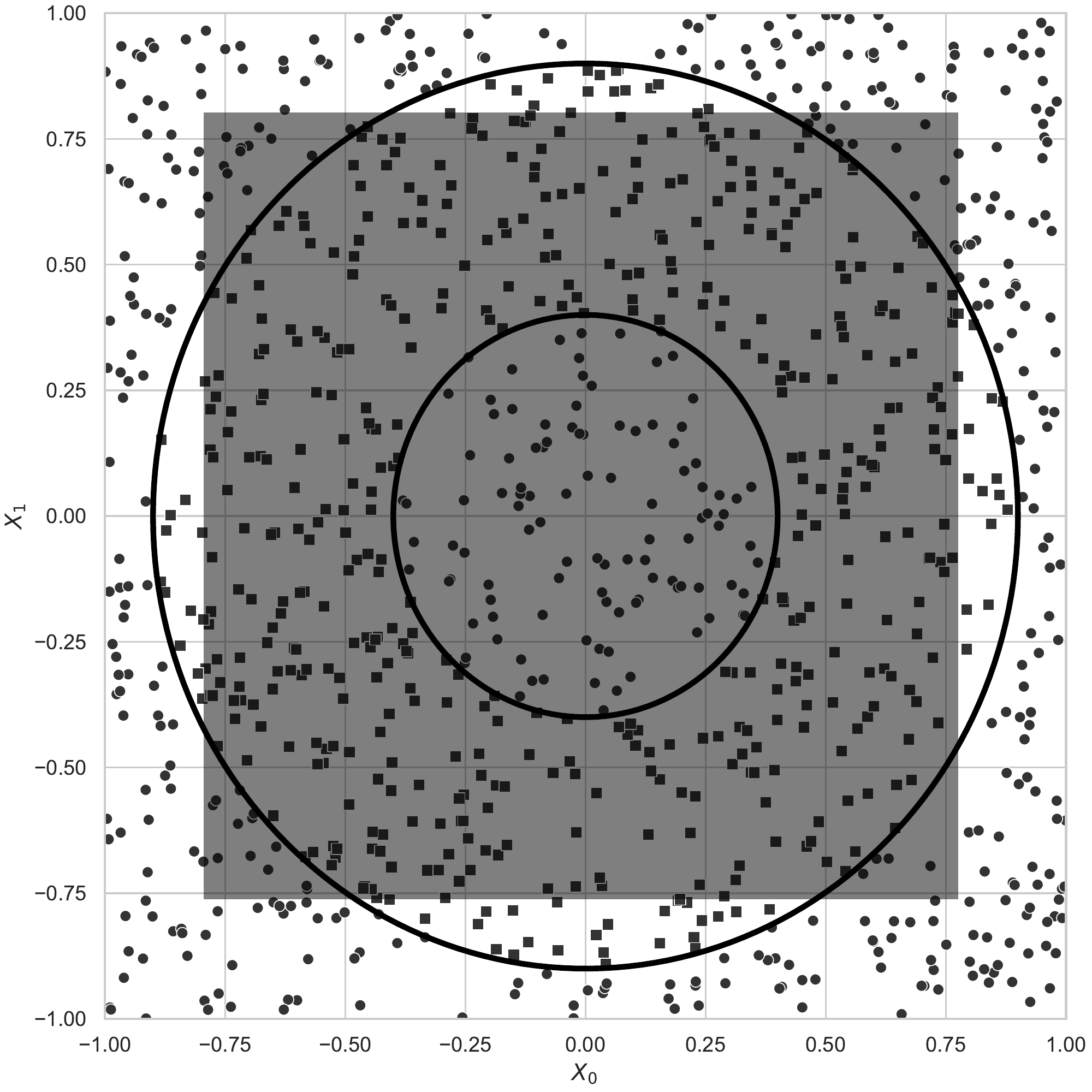}}&
			\subfloat[Complex, $M=3$]{\includegraphics*[scale=0.17]{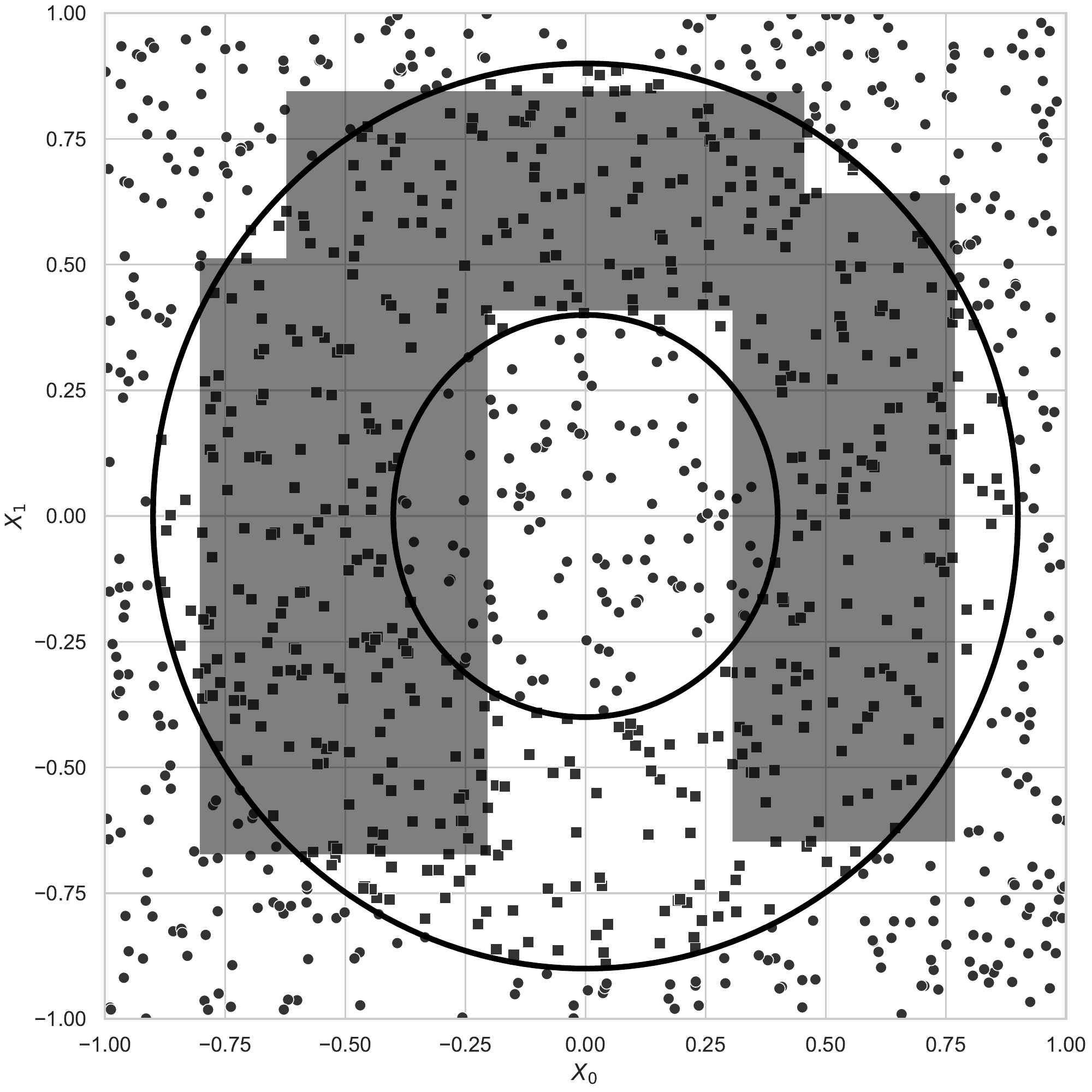}}&
			\subfloat[Complex, $M=5$]{\includegraphics*[scale=0.17]{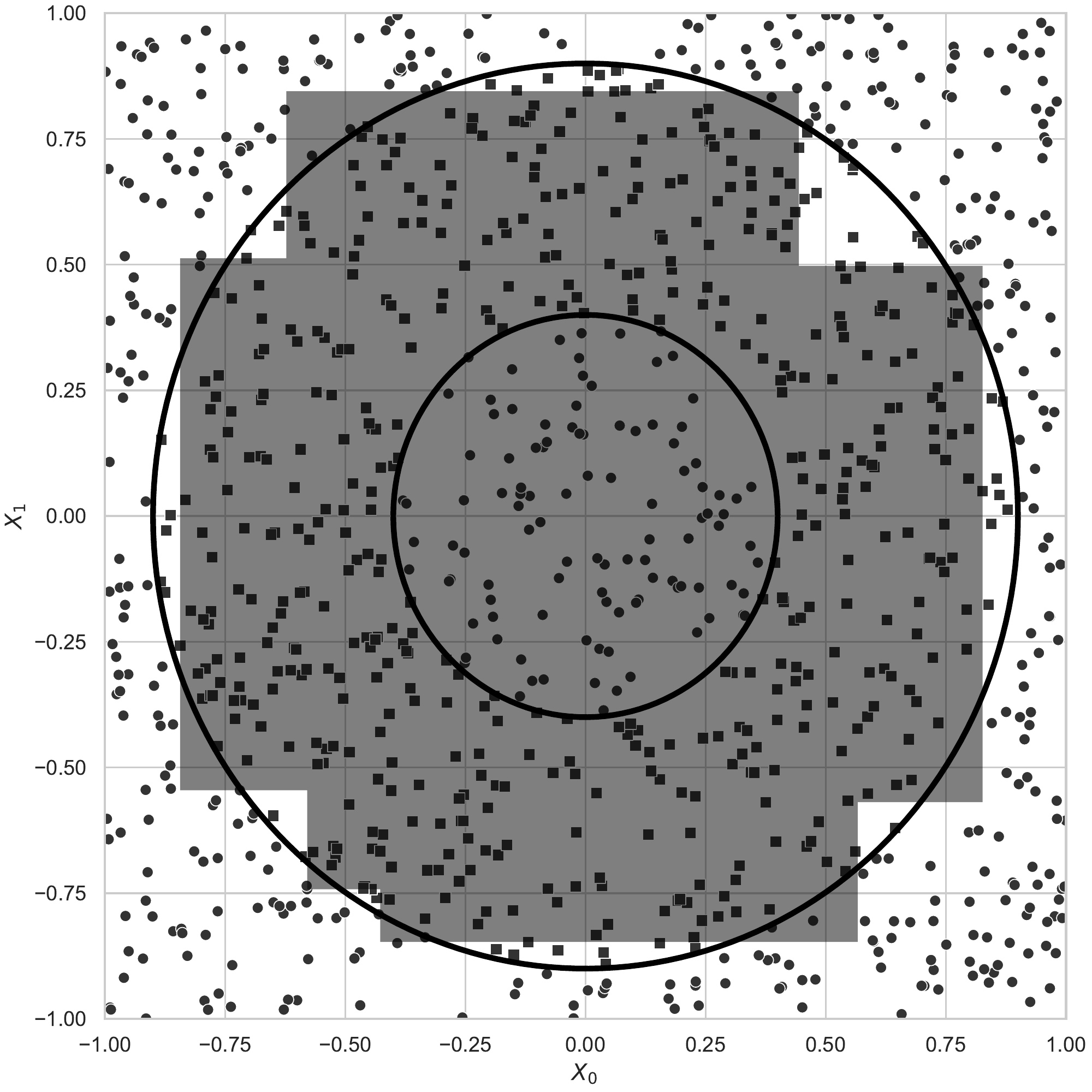}}&
			\subfloat[Complex, $M=10$]{\includegraphics*[scale=0.17]{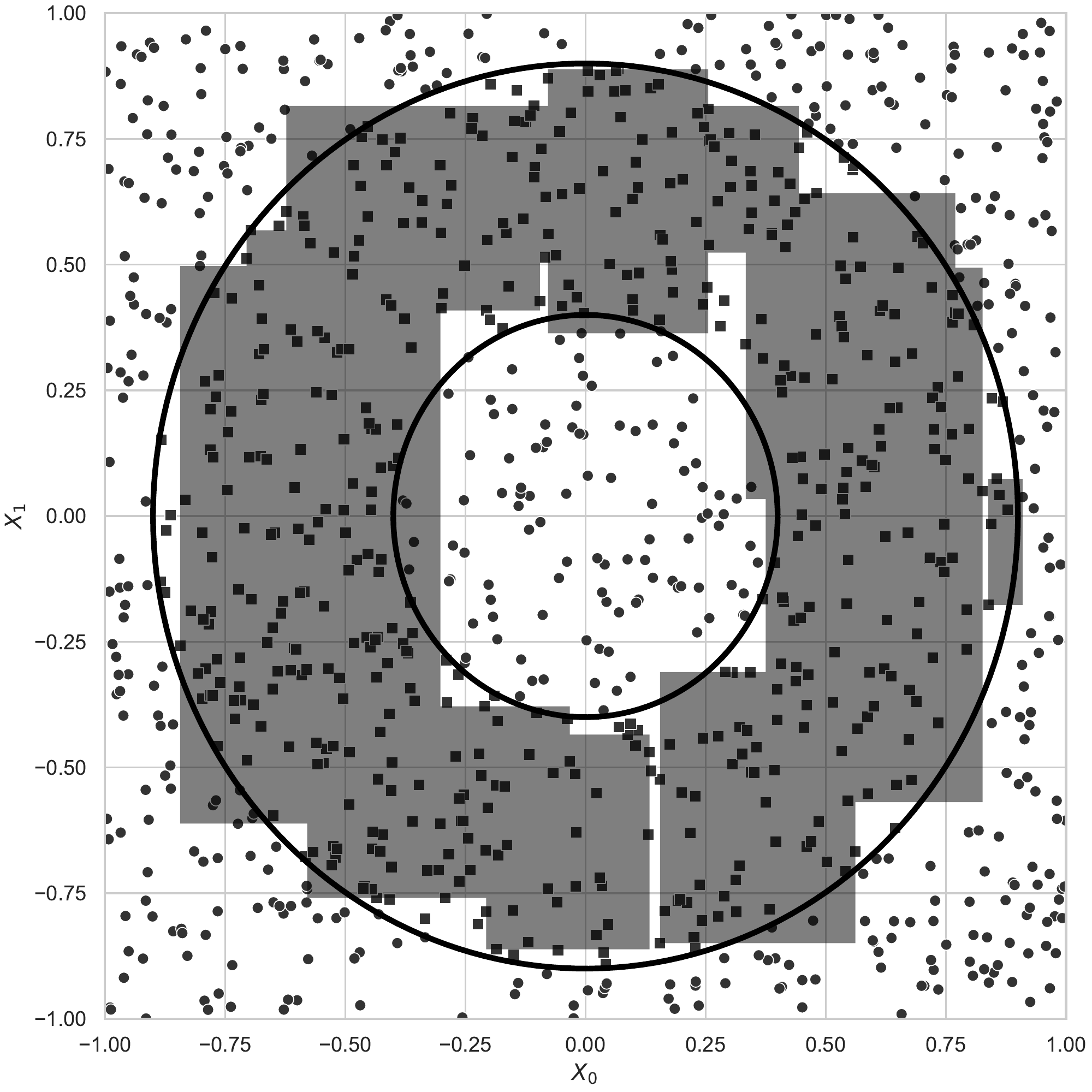}}\\
			
			\subfloat[Very complex, $M=1$]{\includegraphics*[scale=0.17]{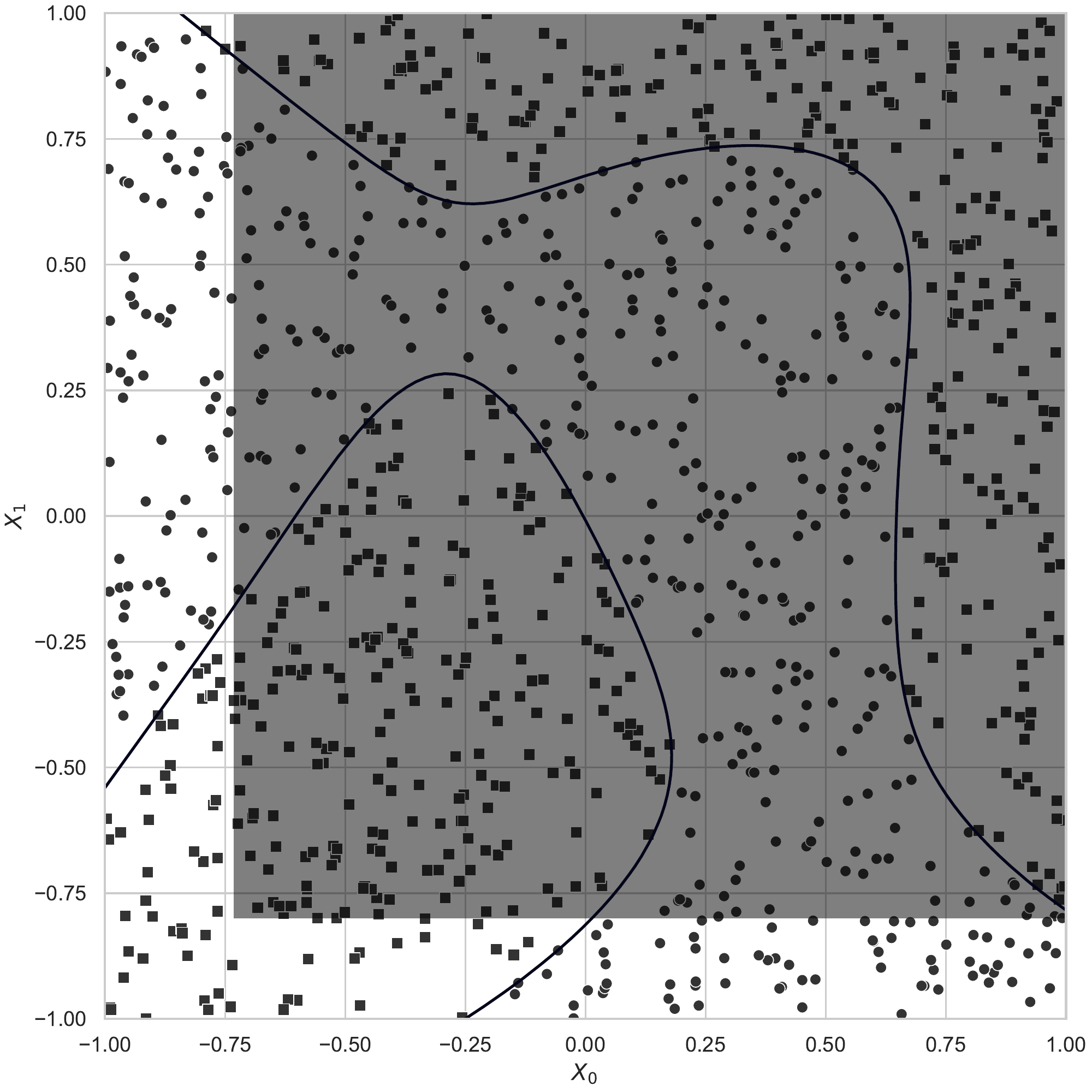}}&
			\subfloat[Very complex, $M=3$]{\includegraphics*[scale=0.17]{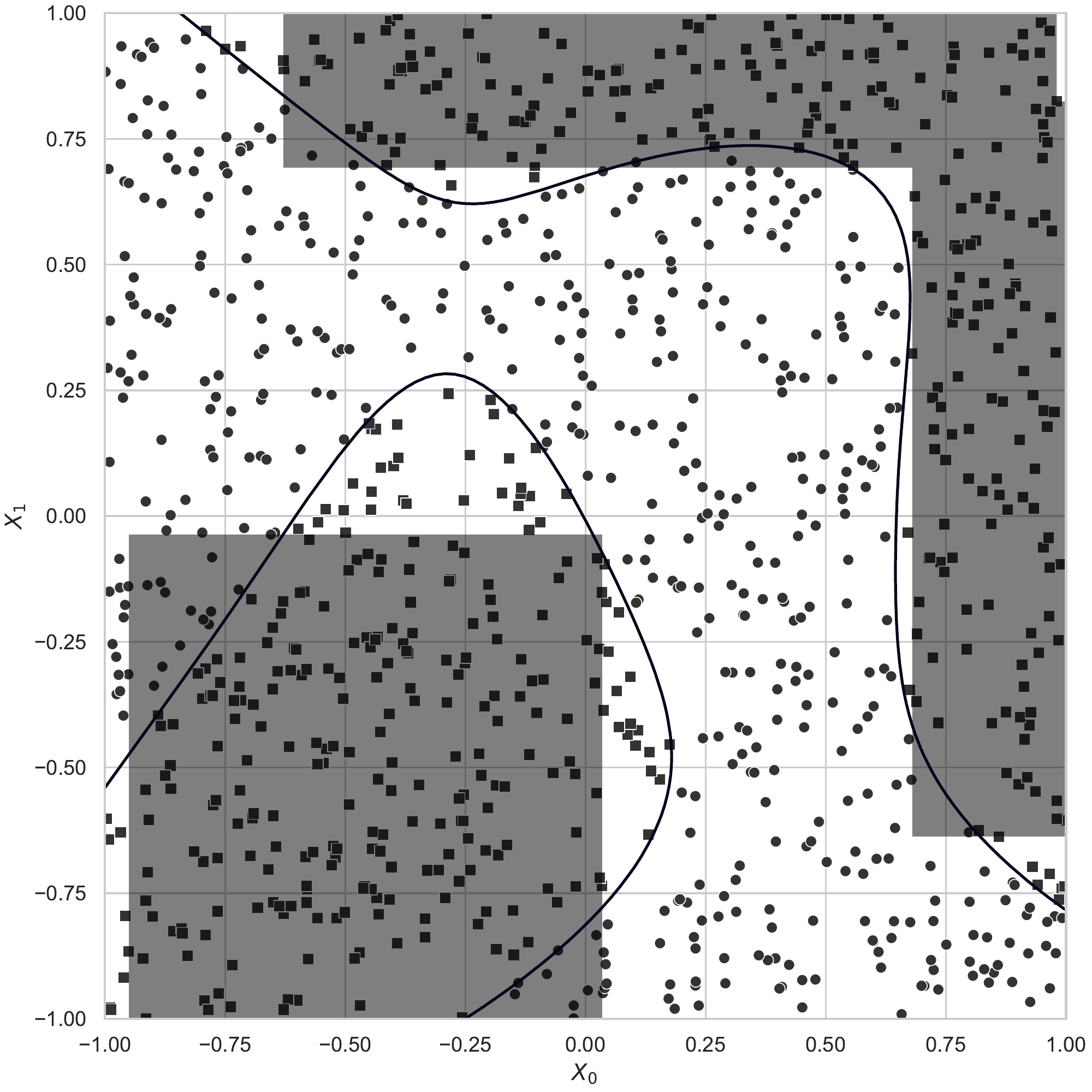}}&
			\subfloat[Very complex, $M=5$]{\includegraphics*[scale=0.17]{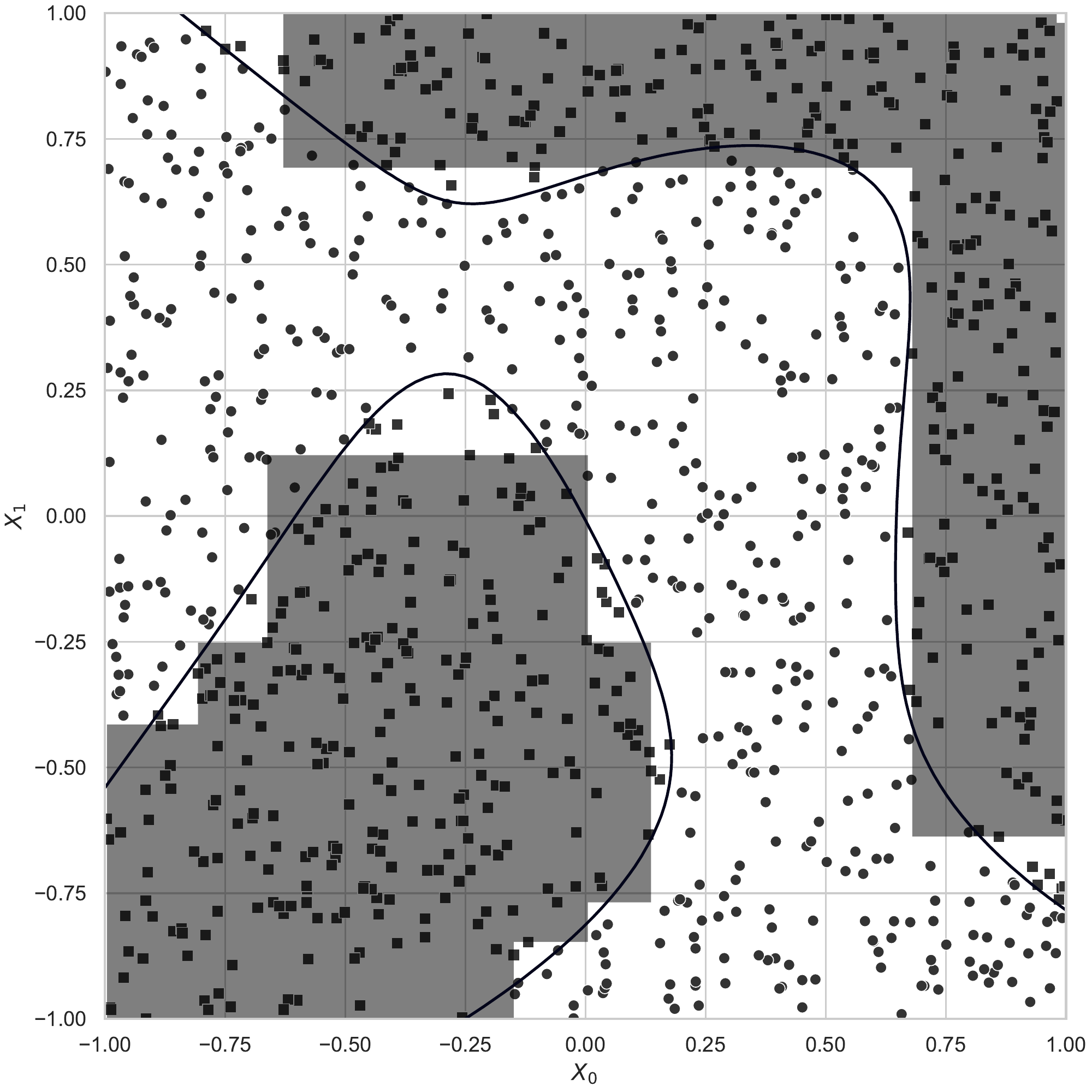}}&
			\subfloat[Very complex, $M=10$]{\includegraphics*[scale=0.17]{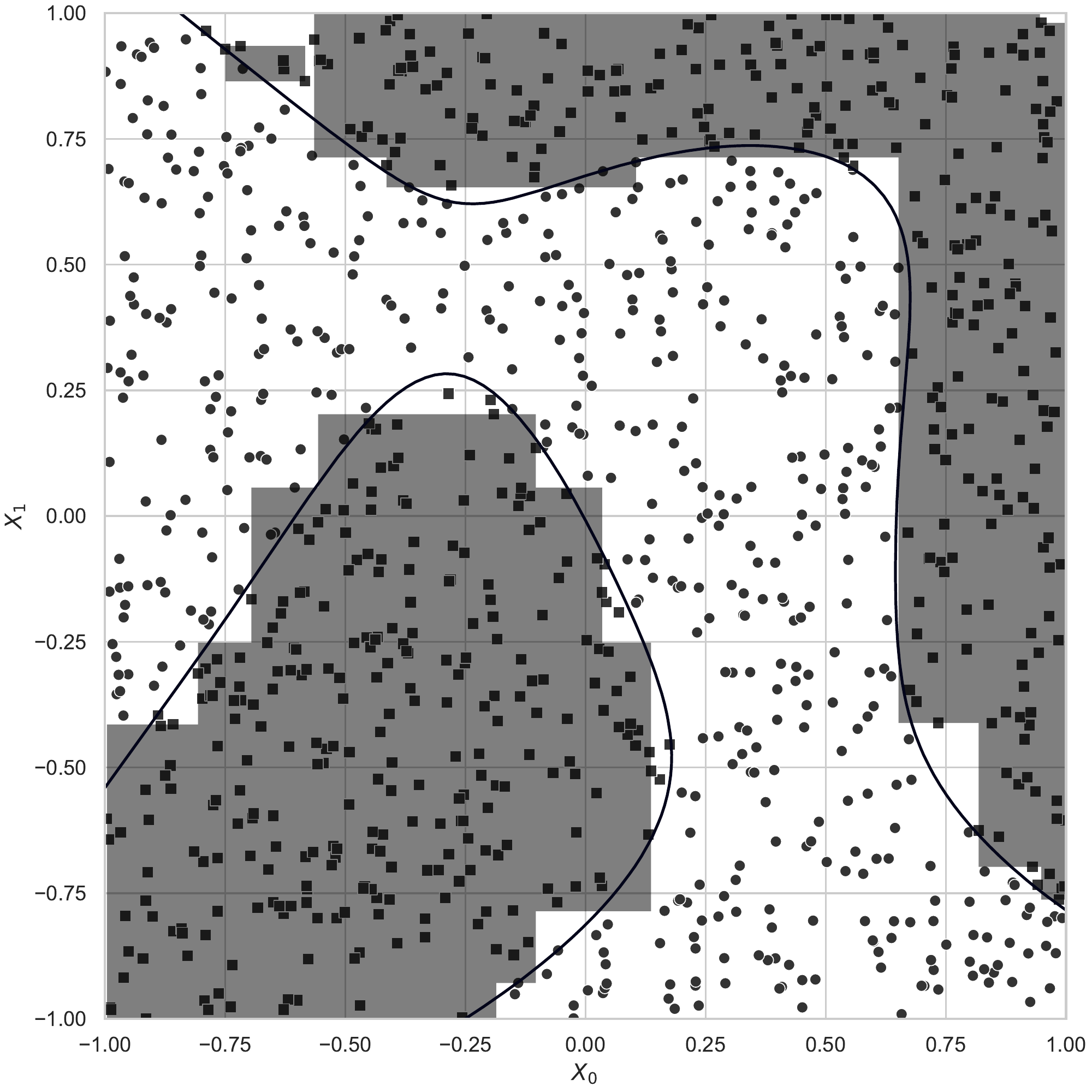}}\\			
		\end{tabular}
	\end{center}
	{\scriptsize{\emph{Notes}. For demonstration, we used the results from the illustration setting in \Cref{sec:simulation_study} (seed=0) with the DR method. The figure shows the hyperboxes returned from IOPL in gray together with the true decision boundaries (black lines). From left to right, one can observe how the parameter $M$, which controls the number of hyperboxes used in the policy, effects the fit. For $M=10$, the hyperboxes almost follow the true decision region.}}
	\label{fig:approximation}
\end{figure}

The definitions of the function $f$ for each of the three scenarios are
\begin{align*}
&\scriptstyle\text{(1) Basic: }f(X,T)=-2T(X_1-\frac{1}{2}X_0^2+\frac{1}{4})\\
&\scriptstyle\text{(2) Complex: }f(X,T)=-2T\left(\left(\frac{4}{10}\right)^2-X_0^2-X_1^2\right)\left(X_0^2+X_1^2-\left(\frac{9}{10}\right)^2\right)\\
&\scriptstyle\text{(3) Very complex: }f(X,T)=-T\left(3\left(1-X_0\right)^2e^{-X_0^2-\left(X_1+1\right)^2}-10\left(X_0 \frac{1}{5}-X_0^3-X_1^5\right)e^{-X_0^2-X_1^2}-\frac{1}{3}e^{-\left(X_0+1\right)^2-X_1^2}-1\right)
\end{align*}
Figure \labelcref{fig:approximation} visualizes the true decision boundaries of the outcome functions in \Cref{sec:simulation_study} for all three scenarios. For smaller numbers of hyperboxes, \ie smaller $M$, one observes a coarse covering of the decision region, while an increasing number of hyperboxes yields a finer approximation of the decision region. This goes in line with our theoretical findings in \Cref{sec:universal_approximation_theorem}.

\FloatBarrier

\subsection{Performance}
\label{app:performance_low_dimensional}

\begin{figure}[h]
	\caption{Regret analysis with DR}
	\begin{center}
		\subfloat[Basic]{\includegraphics*[scale=0.13]{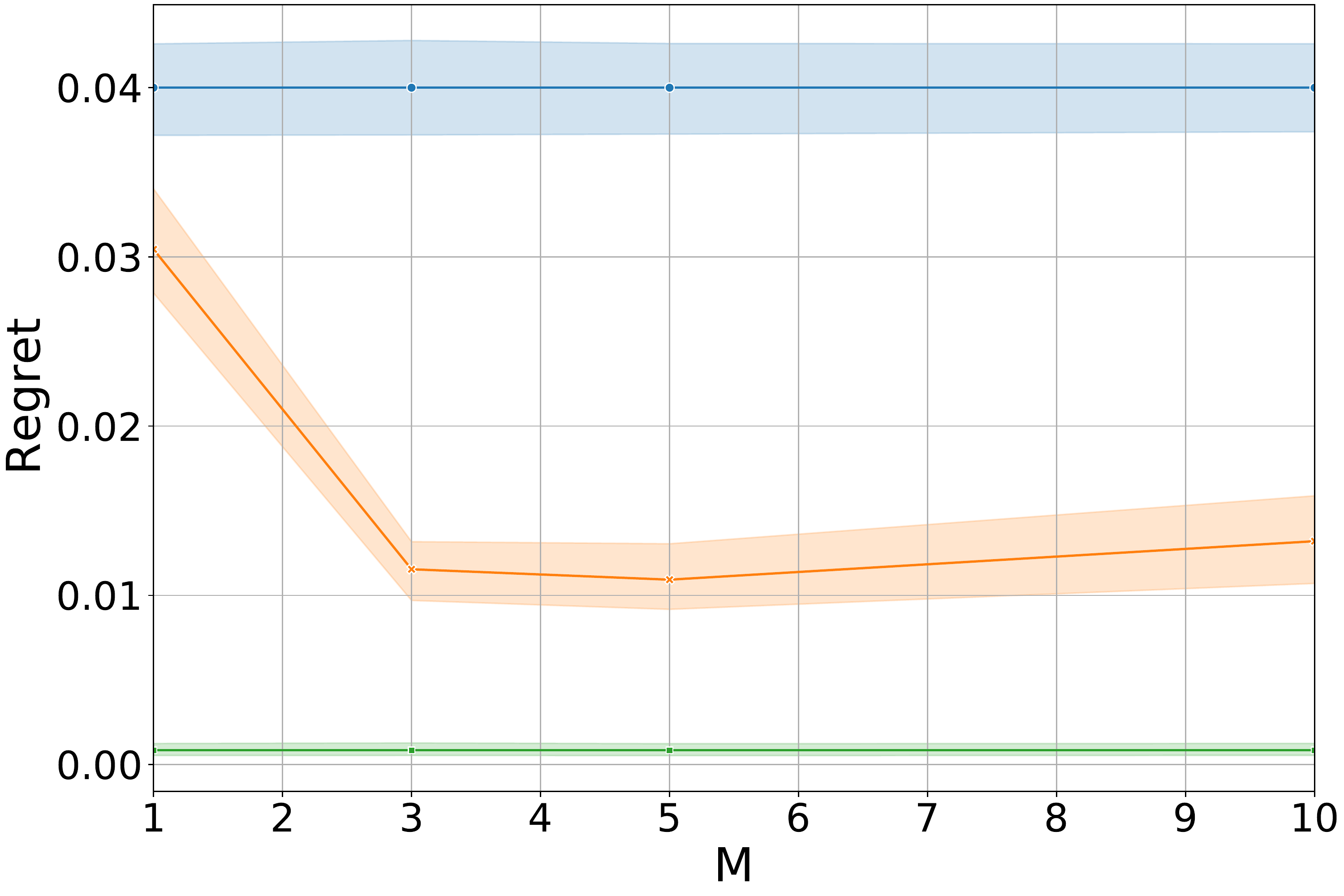}}
		\subfloat[Complex]{\includegraphics*[scale=0.13]{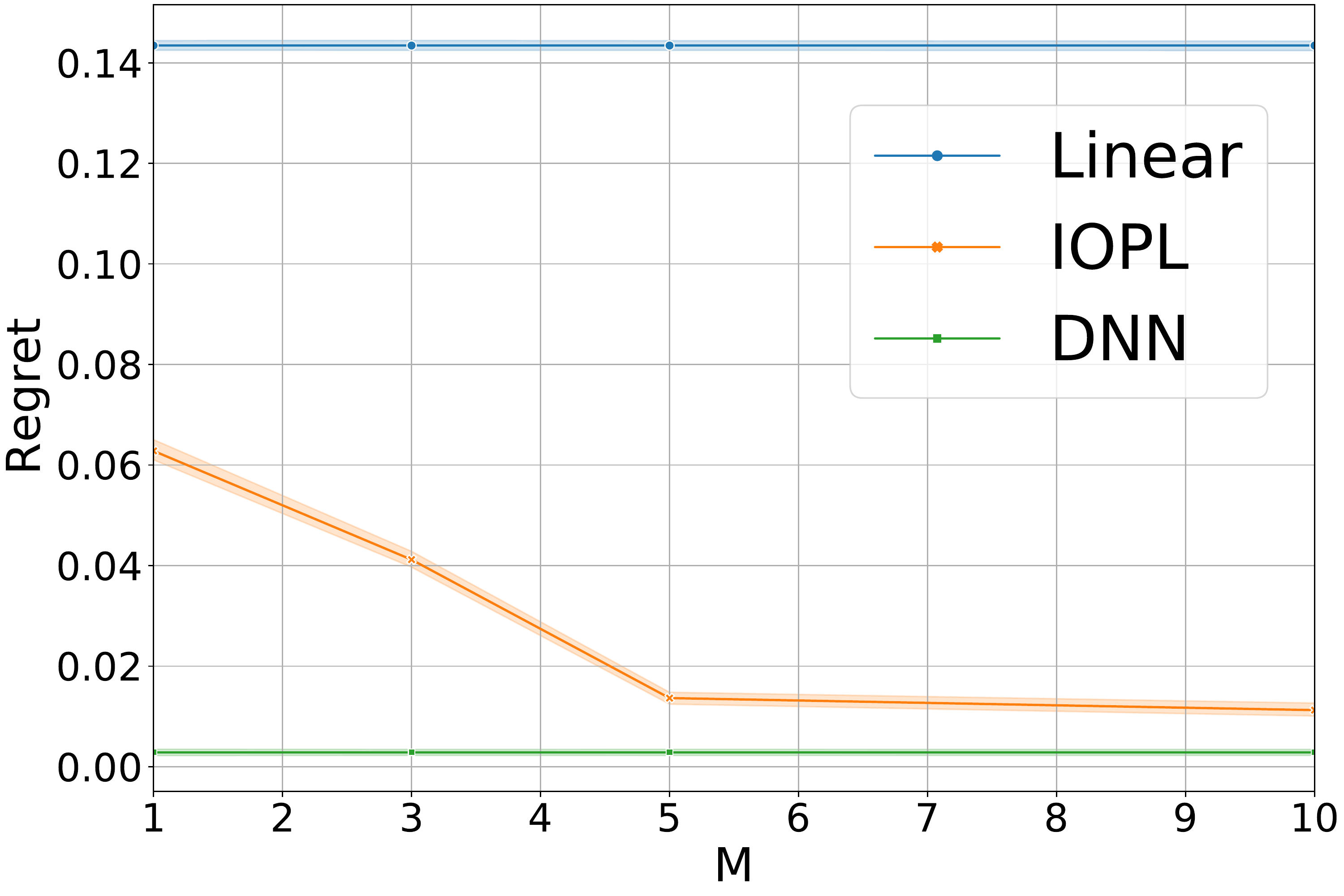}}
		\subfloat[Very complex]{\includegraphics*[scale=0.13]{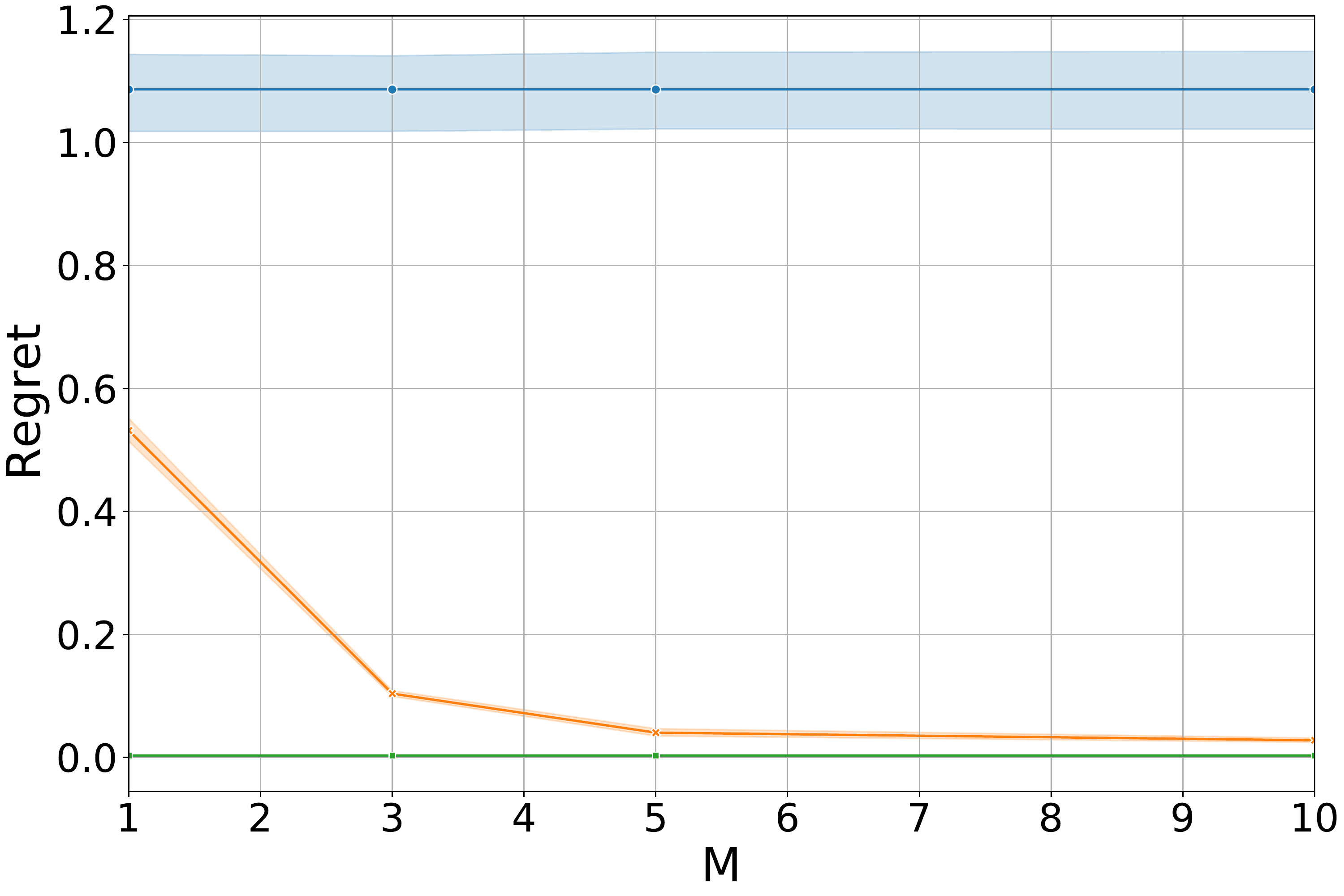}}	
	\end{center}
	{\scriptsize{\emph{Notes}. We set the method to DR. We sample 10 different datasets using the described data-generating processes, which are used for training IOPL and all baselines. For IOPL, we vary the parameter $M$ in $\{1,3,5,10\}$. The shaded areas correspond to the 95\% confidence intervals around the mean.}}
	\label{fig:results}
\end{figure}

In this section, we report the regret as defined in \Cref{sec:experiments} for our illustration setting. We sample 10,000 i.i.d. observations according to the three given data-generating processes and use $n=1,000$ for training. Evidently, DR performs best as can be seen in Figure \labelcref{fig:results}. In the basic scenario, we observe that the linear baseline already performs sufficiently well. Furthermore, we see that the simple form of the decision region in the basic scenario leads to over-fitting for IOPL, which can be seen from the stagnation of the regret for $M>3$. This goes in line with the observations in Figure \labelcref{fig:basic_M_5,fig:basic_M_10}.

\begin{figure}[h]
	\caption{Regret analysis with DM}
	\begin{center}
		\subfloat[Basic]{\includegraphics*[scale=0.13]{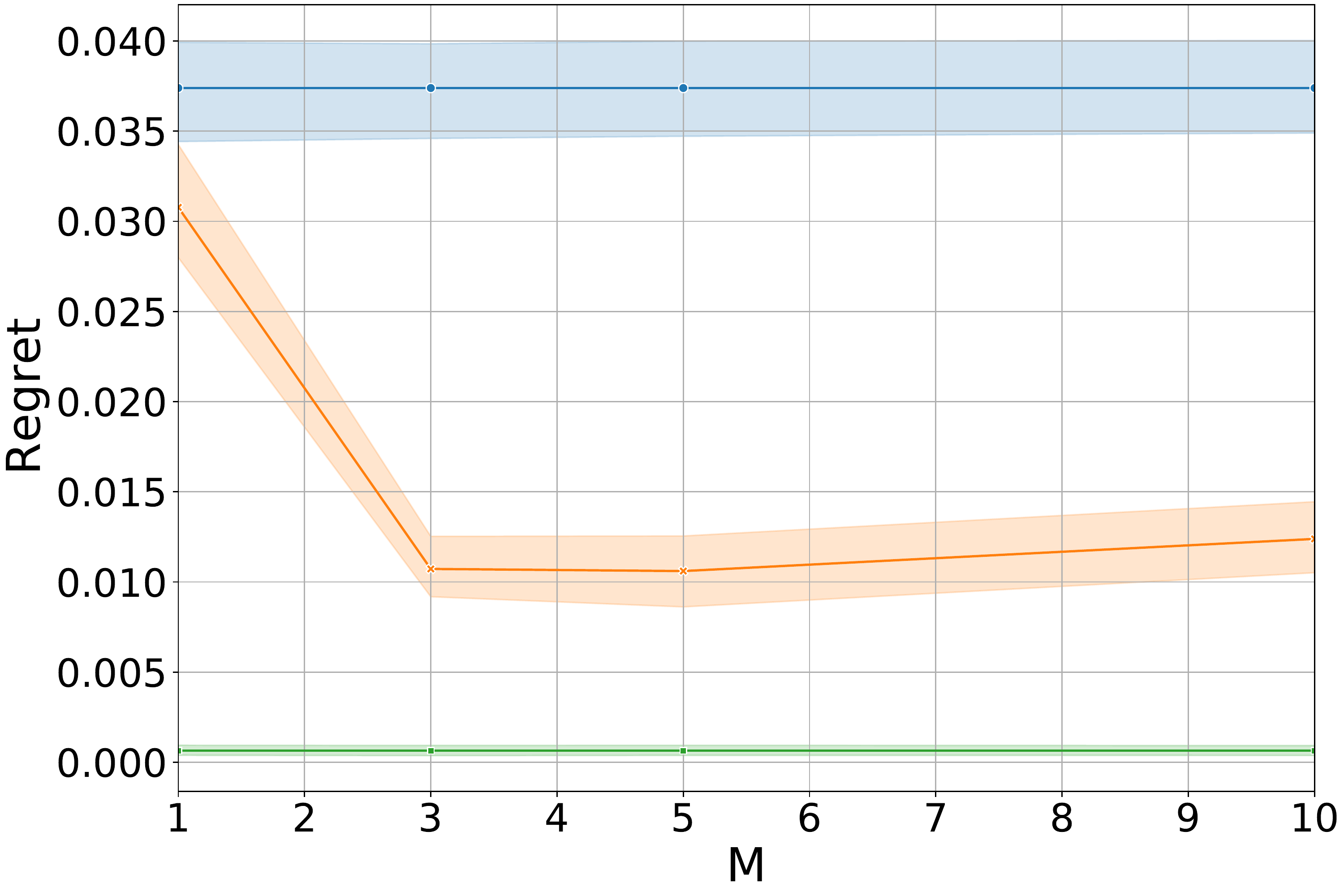}}
		\subfloat[Complex]{\includegraphics*[scale=0.13]{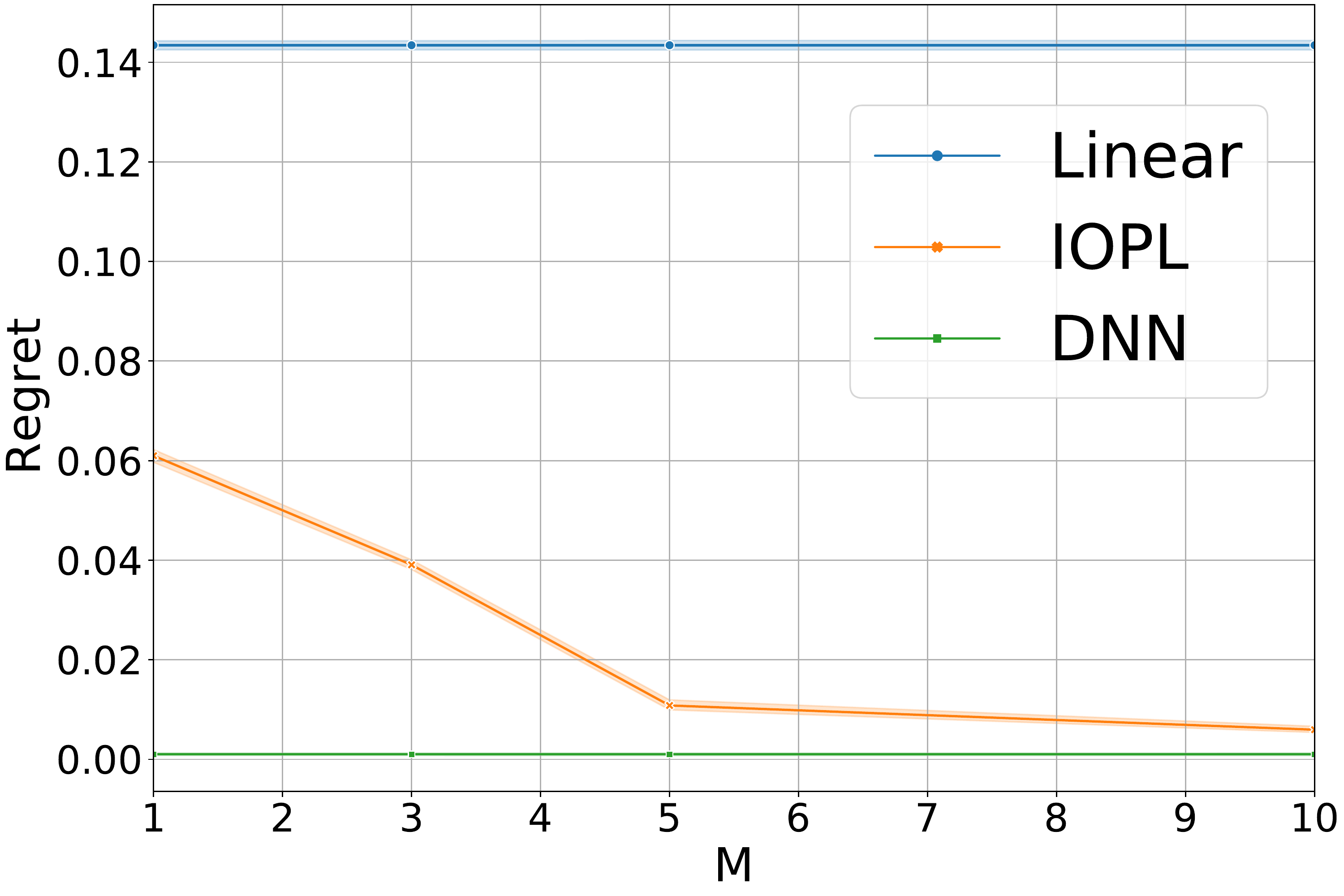}}
		\subfloat[Very complex]{\includegraphics*[scale=0.13]{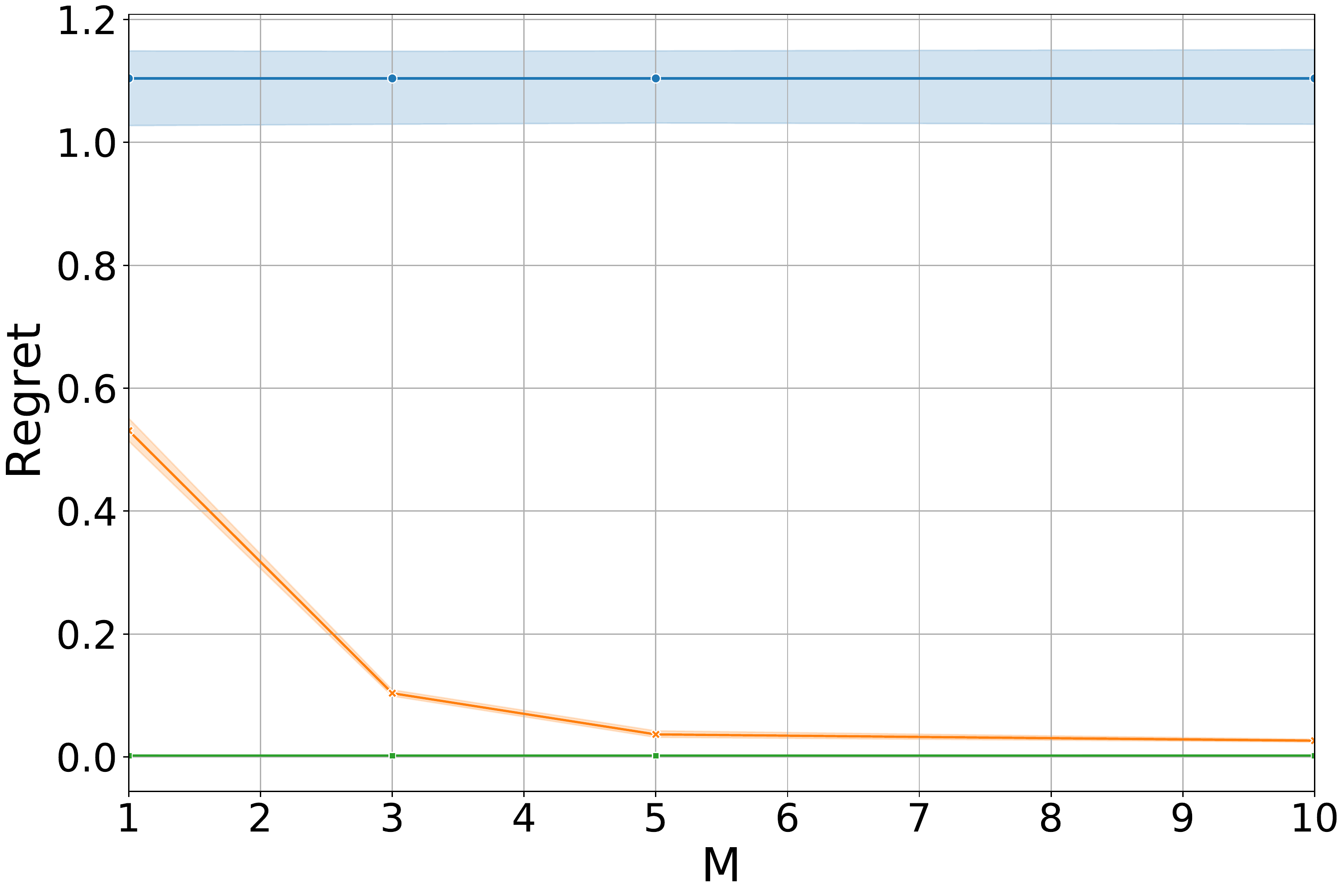}}	
	\end{center}
	{\scriptsize{\emph{Notes}. We set the method to DM. We sample 10 different datasets using the described data generating processes. All baselines and IOPL are trained on all 10 datasets. For IOPL, we vary the parameter $M$ in $\{1,3,5,10\}$. The shaded areas correspond to the 95\% confidence intervals around the mean.}}
	\label{fig:results_DM}
\end{figure}

\begin{figure}[h]
	\caption{Regret analysis with IPS}
	\begin{center}
		\subfloat[Basic]{\includegraphics*[scale=0.13]{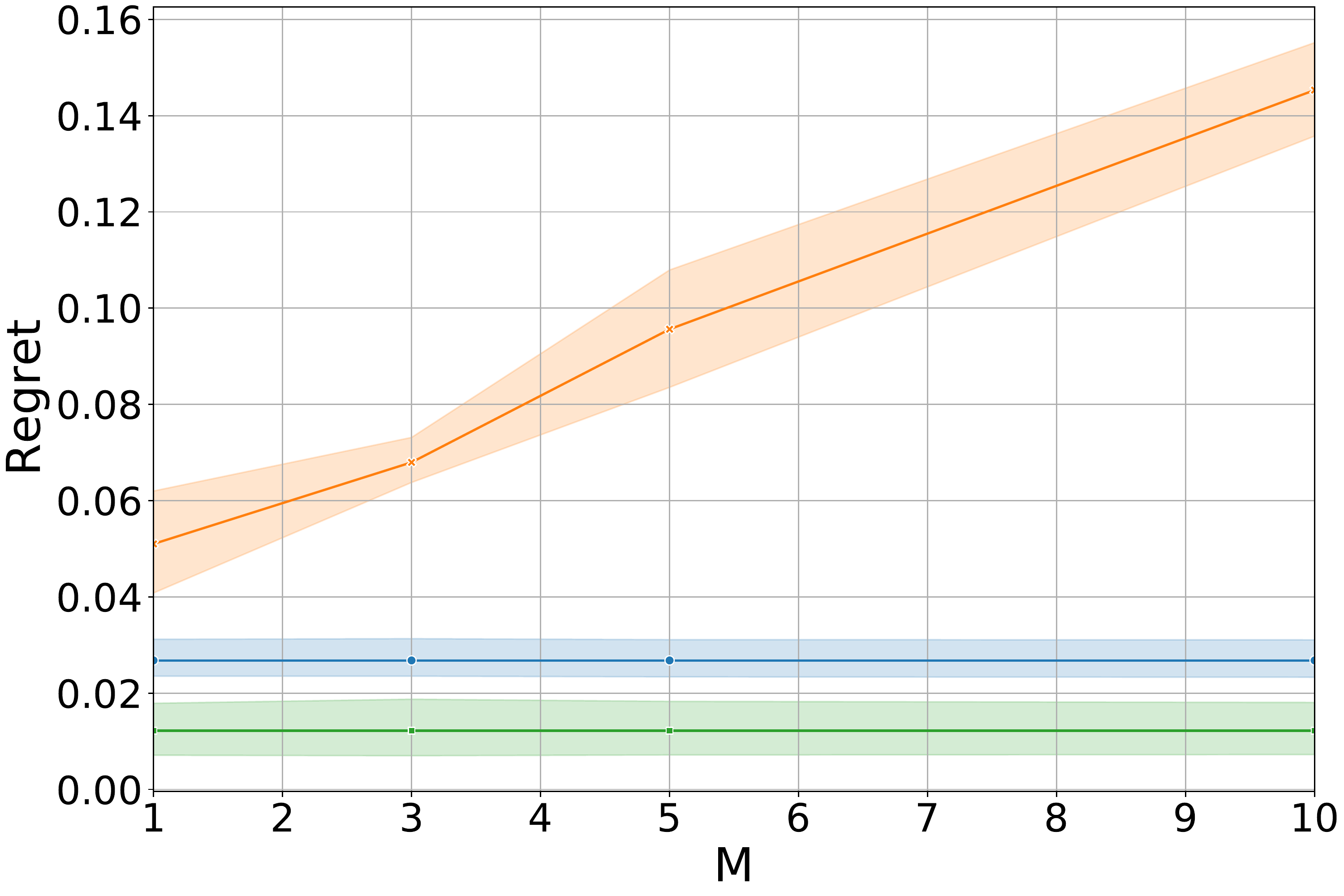}}
		\subfloat[Complex]{\includegraphics*[scale=0.13]{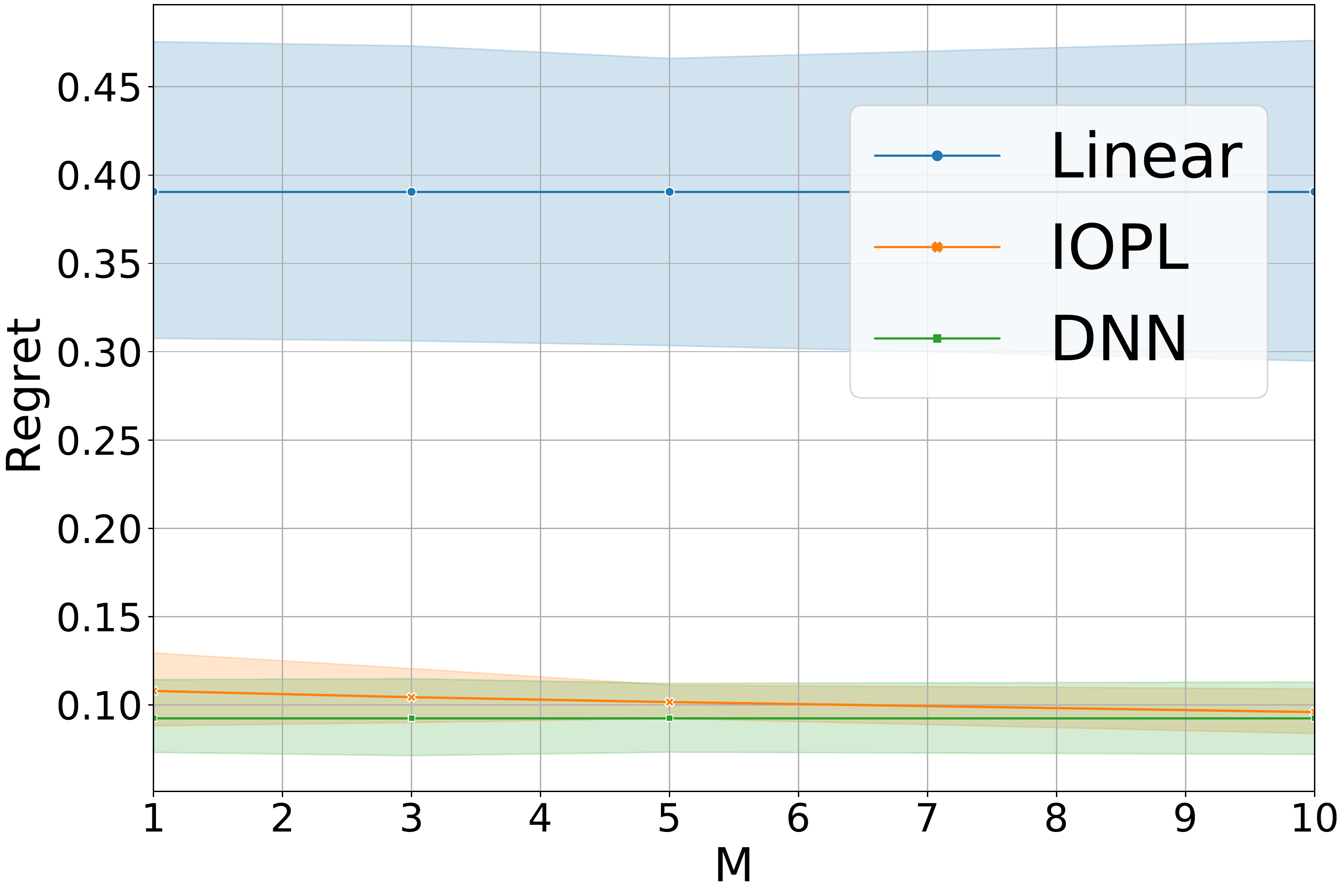}}
		\subfloat[Very complex]{\includegraphics*[scale=0.13]{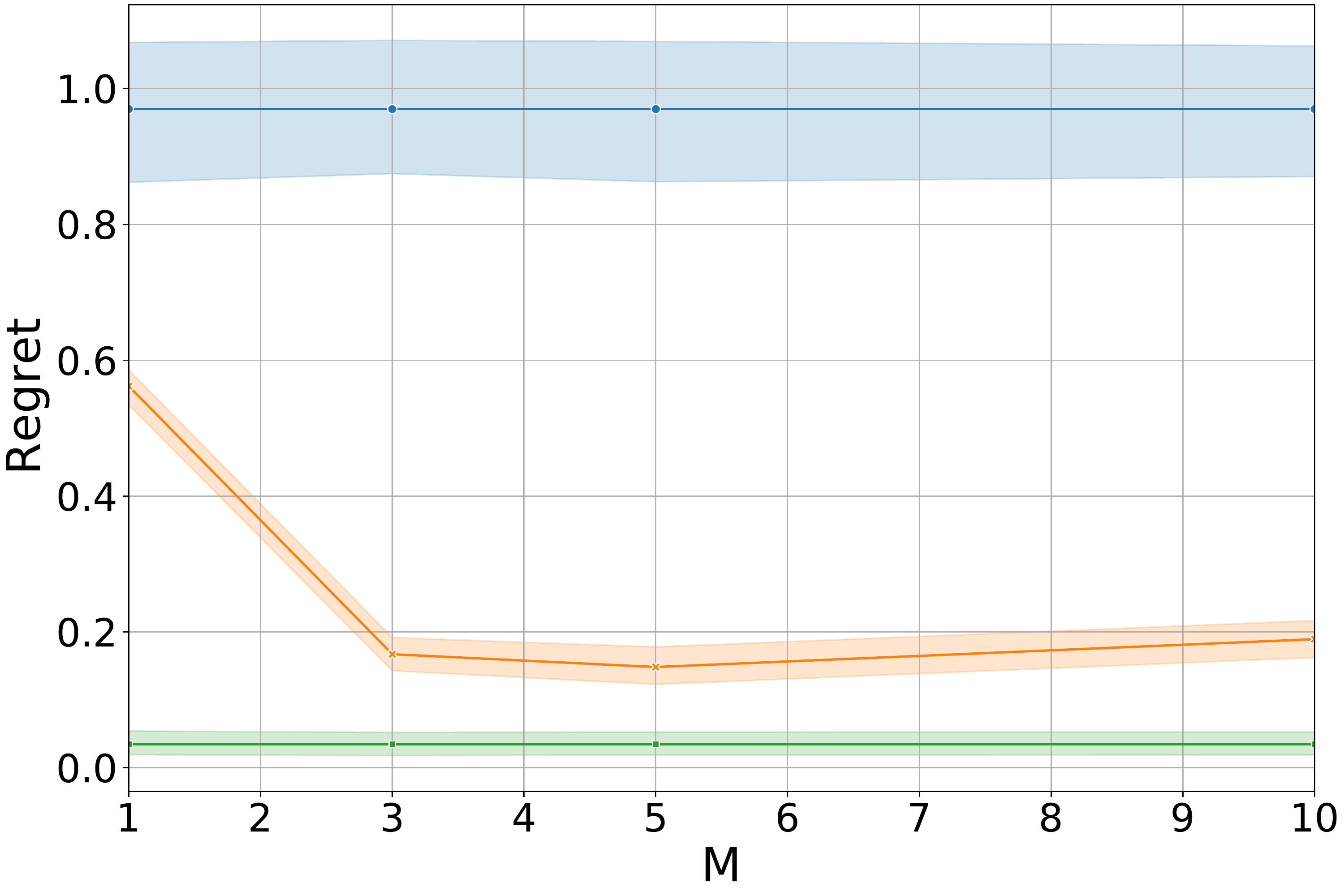}}	
	\end{center}
	{\scriptsize{\emph{Notes}. We set the method to IPS. We sample 10 different datasets using the described data generating processes. All baselines and IOPL are trained on all 10 datasets. For IOPL, we vary the parameter $M$ in $\{1,3,5,10\}$. The shaded areas correspond to the 95\% confidence intervals around the mean.}}
	\label{fig:results_IPS}
\end{figure}

\FloatBarrier

\newpage
\section{Additional Experiments}
\label{app:additional_experiments}

\subsection{Regret Analysis in a Simple Setting}
In this section we show how IOPL and the baselines perform in a simpler setting. That is, we study a setting in which the optimal policy $\pi^\ast$ can be represented as a decision list and a simple decision tree. To do so, we generate again 4-dimensional patient covariates $X_0,X_1,X_2,X_3\sim U[-1,1]$, a treatment assignment which is independent of $X$ with $\Prb{T=1}=\nicefrac{1}{2}$, and an outcome $Y\mid X, T\sim \mathcal{N}(m(X, T), \sigma^2)$ with $\sigma=0.1$. This follows exactly the setting in the regret analysis of \Cref{sec:experiments}. However, we now set the mean to $m(X,T)=\max(X_2+X_3,0)+\nicefrac{1}{2} \ T \ \text{sign}(X_0X_1)$. That is, we replace the term $\prod_{i=0}^3 X_i$ by $X_0X_1$. In this way, it is possible to represent the optimal policy
\begin{align}
\pi^\ast(X)=\begin{cases}
1 & \text{if } \text{sign}(X_0)\neq \text{sign}(X_1),\\
-1 & \text{else,}
\end{cases}
\end{align}
via a decision list and a simple decision tree.

\begin{figure}[H]
	\caption{Regret analysis}
	\begin{center}
		\includegraphics*[scale=0.15]{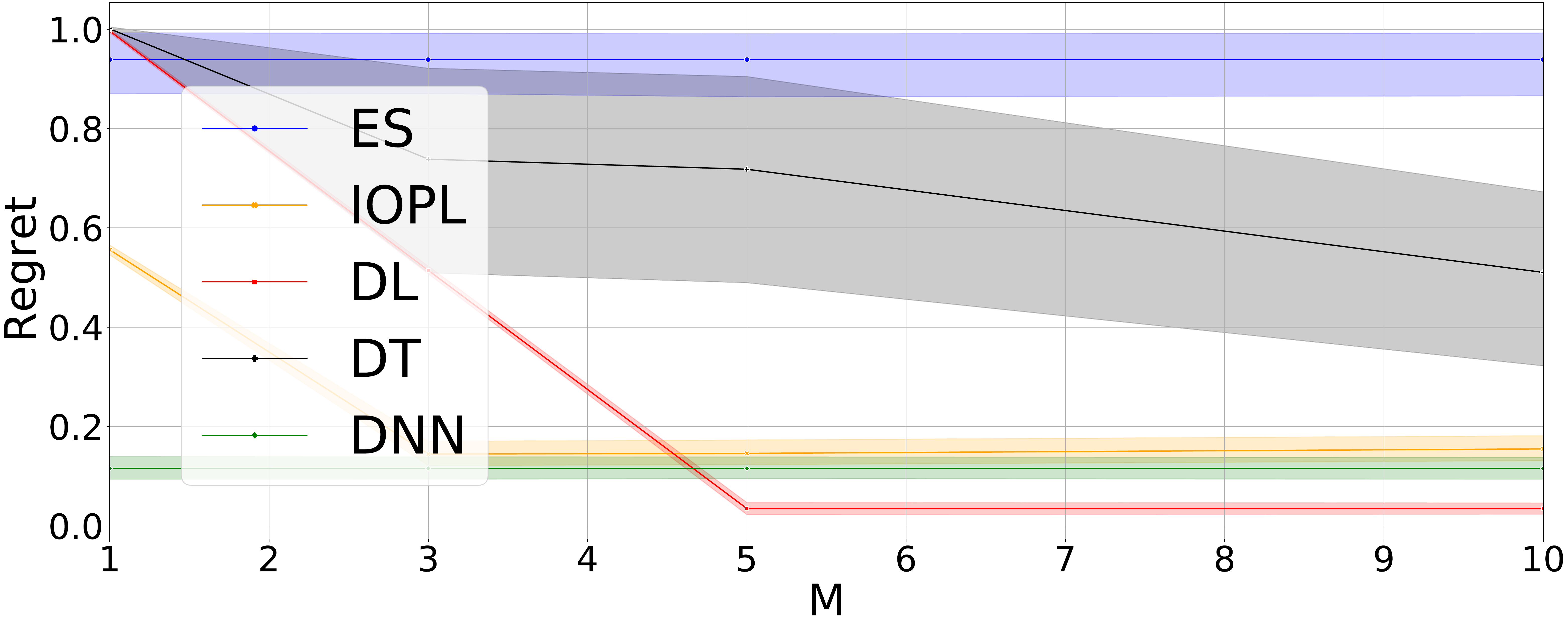}
	\end{center}
	\label{fig:simpler_setting}
	{\scriptsize{\emph{Notes}. We sample 10 different datasets using the described data-generating processes. For IOPL, DL, and DT, we vary the parameter $M$ in $\{1,3,5,10\}$. The shaded areas are the 95\% confidence intervals.}}
\end{figure}

The results of our experiments are summarized in Figure~\ref{fig:simpler_setting}. Evidently, now also the baselines are able to approximate the optimal policy and, hence, yield lower regrets. IOPL still outperforms the interpretable baselines with the exception of DL. This is due to the fact that \Cref{alg:DL_optimization} is implemented as a brute-force approach and hence yields a global optimum in the training task, while IOPL is stopped after $l=50$ iterations.

\subsection{Comparison to Optimal Prescriptive Trees (OPTs)}

In theory, decision trees are also universal approximators and, hence, are sufficiently rich (see Theorem 4.1 in \citet{Stein.2009} and use that simple functions can be approximated by decision trees). Thus, the lack of theoretical approximation guarantees stems from the fact that heuristic greedy algorithms are used to train decision trees, \eg, CART. That is, decision trees trained via such heuristics cannot leverage their full potential. As a remedy, \citet{Bertsimas.2019} recently proposed an algorithm to train so-called optimal prescriptive trees (OPTs). Therein, the authors are building the complete tree at once and, hence, refrain from common heuristic approaches to train decision trees. As a result, decision trees trained with the proposed algorithm become more expressive. A Python implementation of the algorithm is provided in the package \citet{InterpretableAI}. In the following, we demonstrate that IOPL is still superior to OPTs for two reasons: (i)~IOPL is less prone to overfitting. (ii)~To ensure that a decision tree has the same approximation quality as a policy from our policy class, one requires very deep trees, which renders them unintelligible.

\begin{figure}[h]
	\caption{Approximations of Optimal Prescriptive Trees and IOPL}
	\begin{center}
		\begin{tabular}{cccc}
			\subfloat[IOPL, $M=1$]{\includegraphics*[scale=0.17]{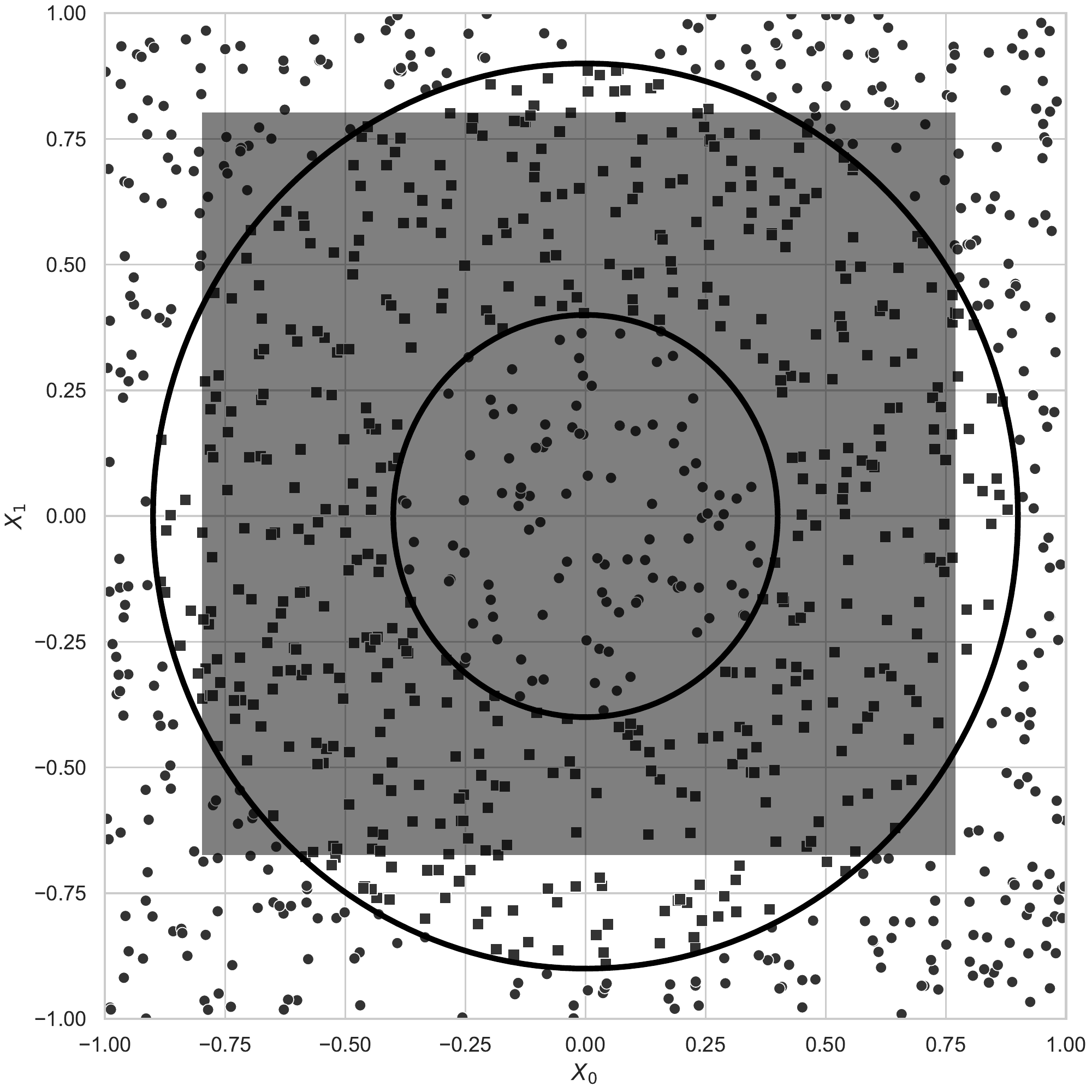}}&
			\subfloat[IOPL, $M=3$]{\includegraphics*[scale=0.17]{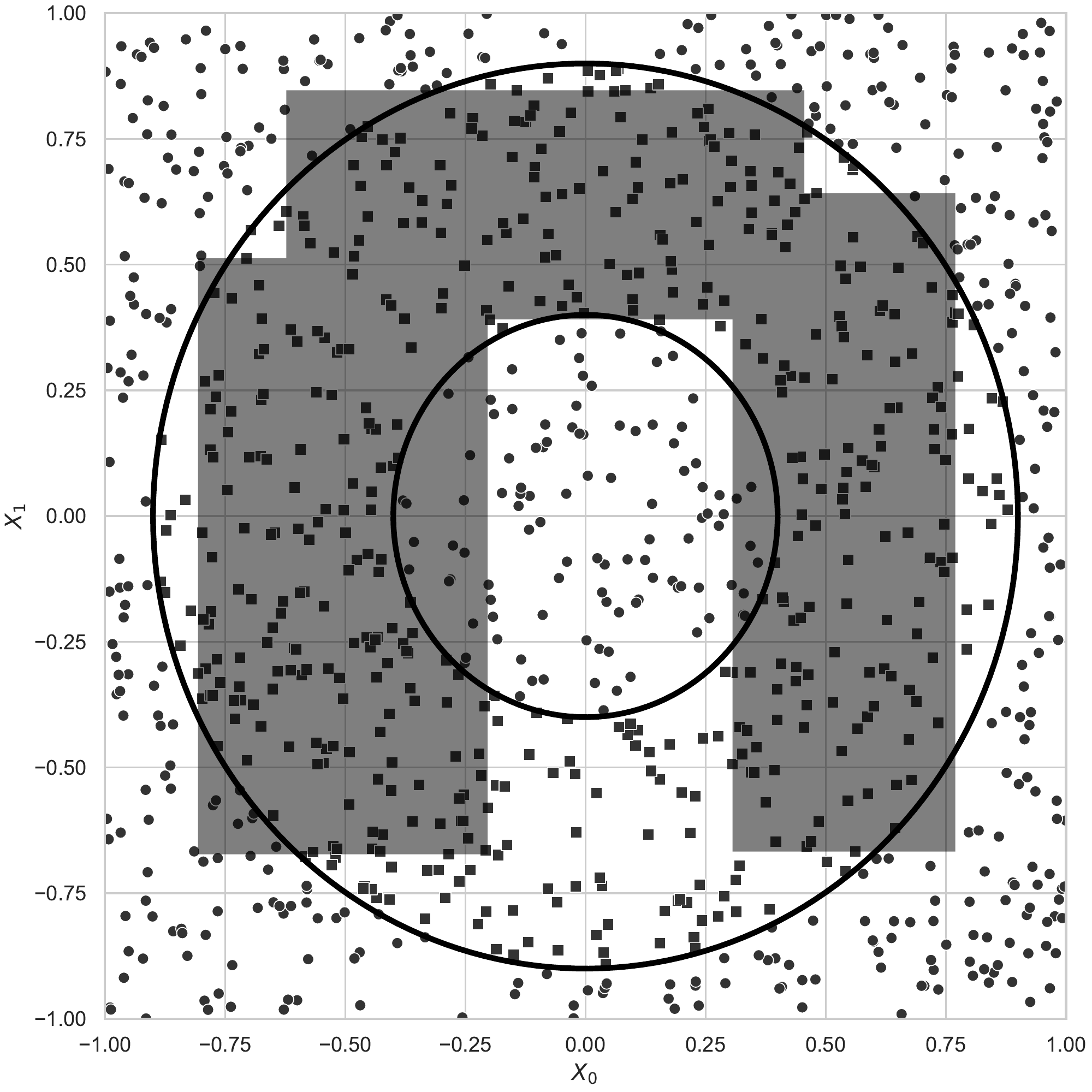}}&
			\subfloat[IOPL, $M=5$]{\includegraphics*[scale=0.17]{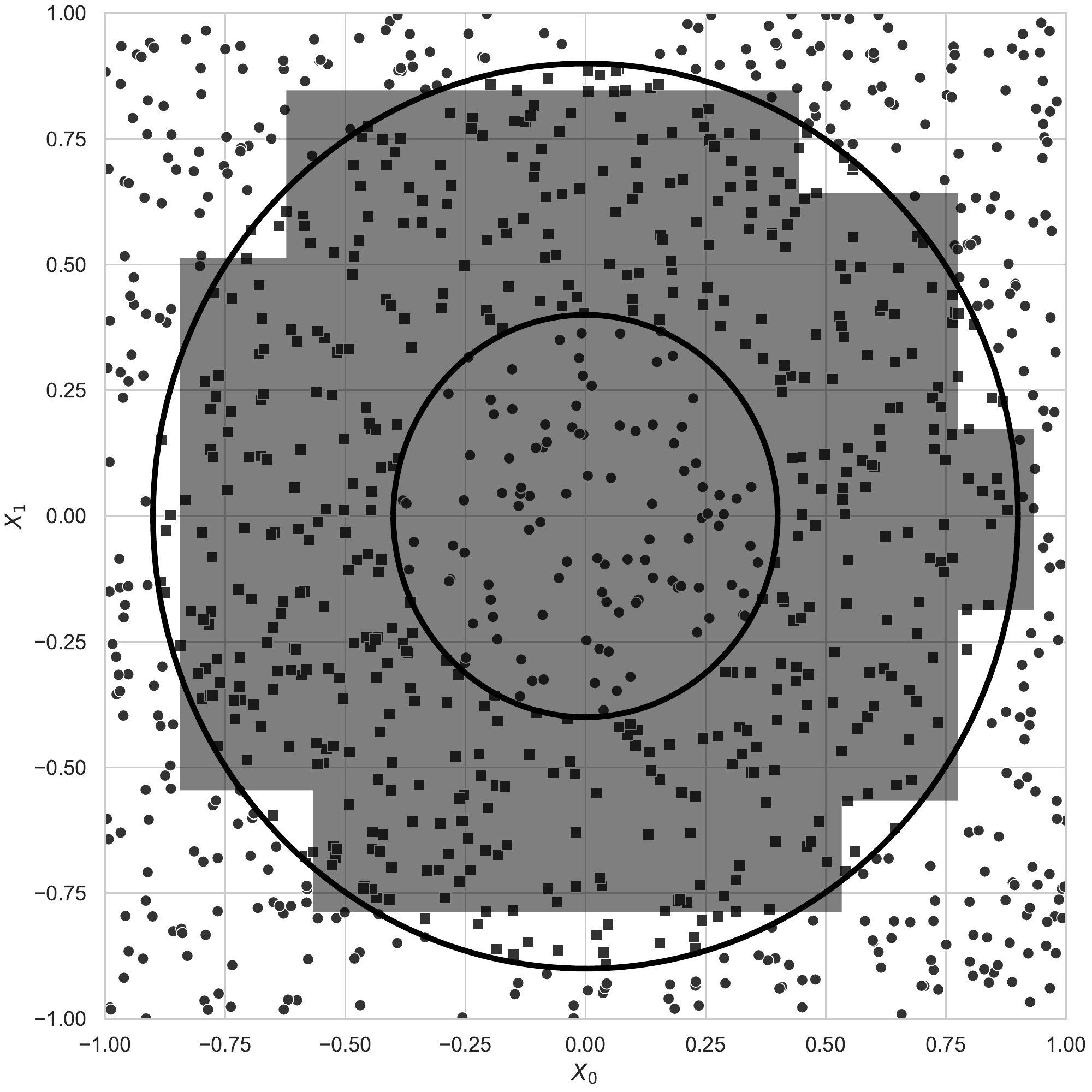}}&
			\subfloat[IOPL, $M=10$]{\includegraphics*[scale=0.17]{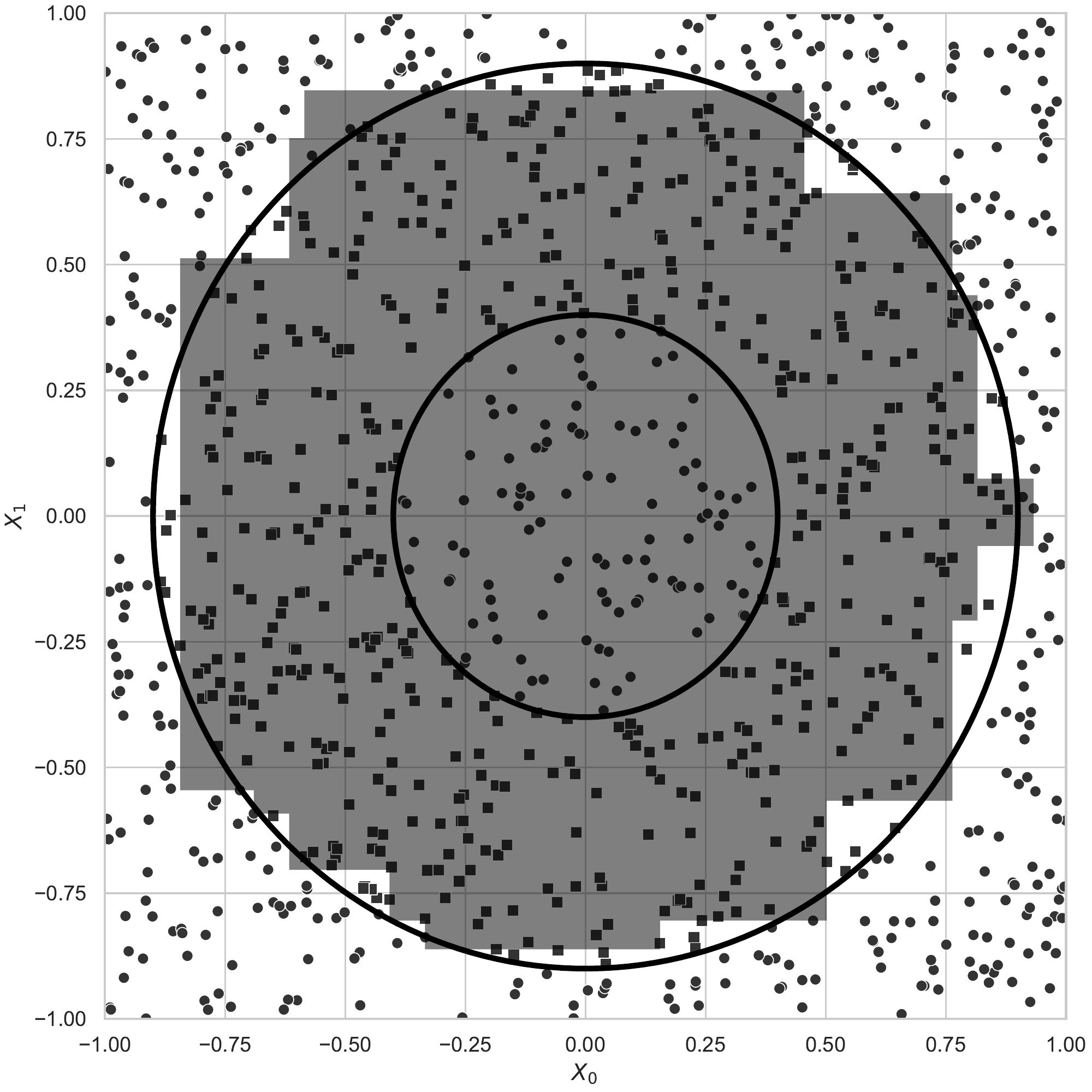}}\\
			
			\subfloat[OPT, $M=1$]{\includegraphics*[scale=0.17]{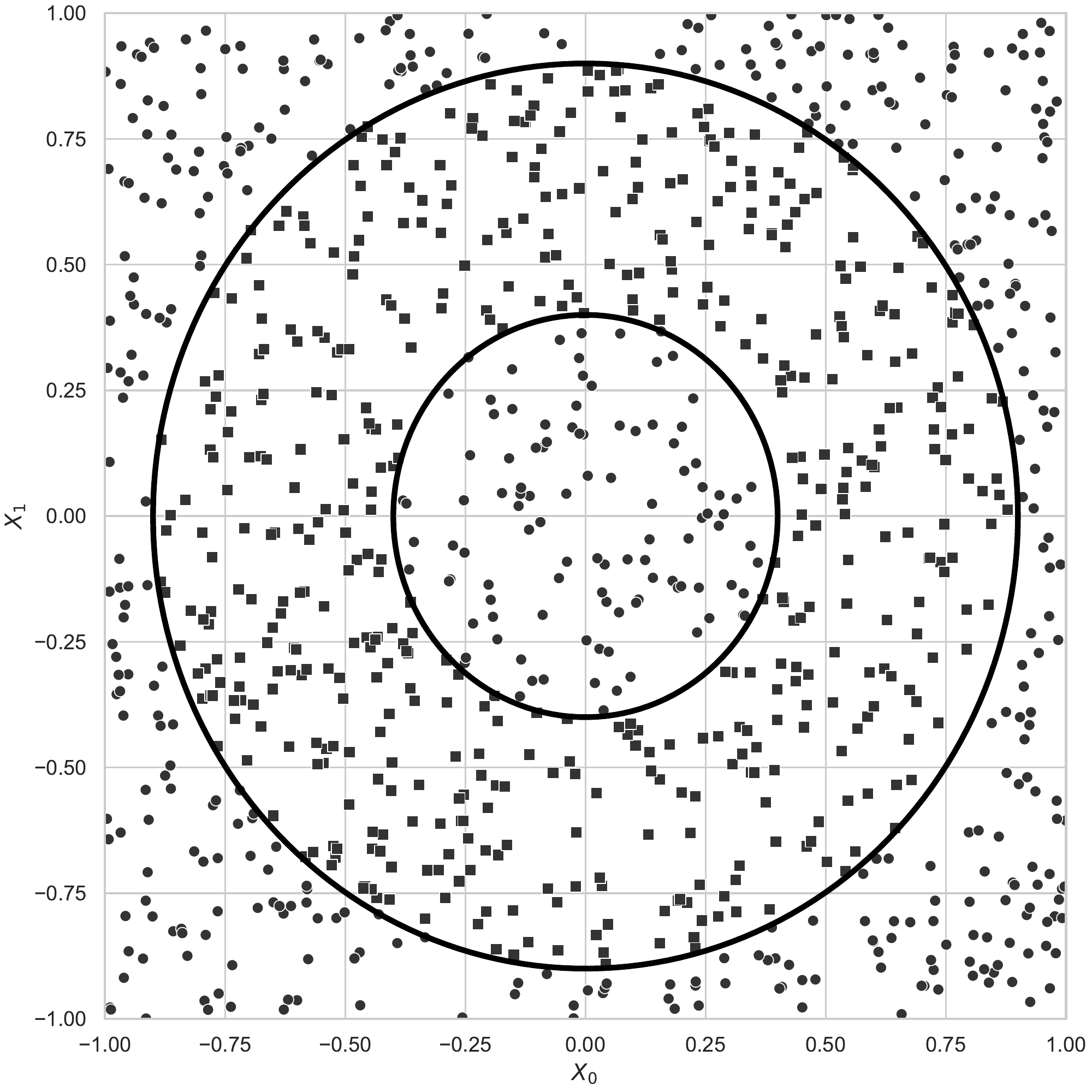}}&
			\subfloat[OPT, $M=3$]{\includegraphics*[scale=0.17]{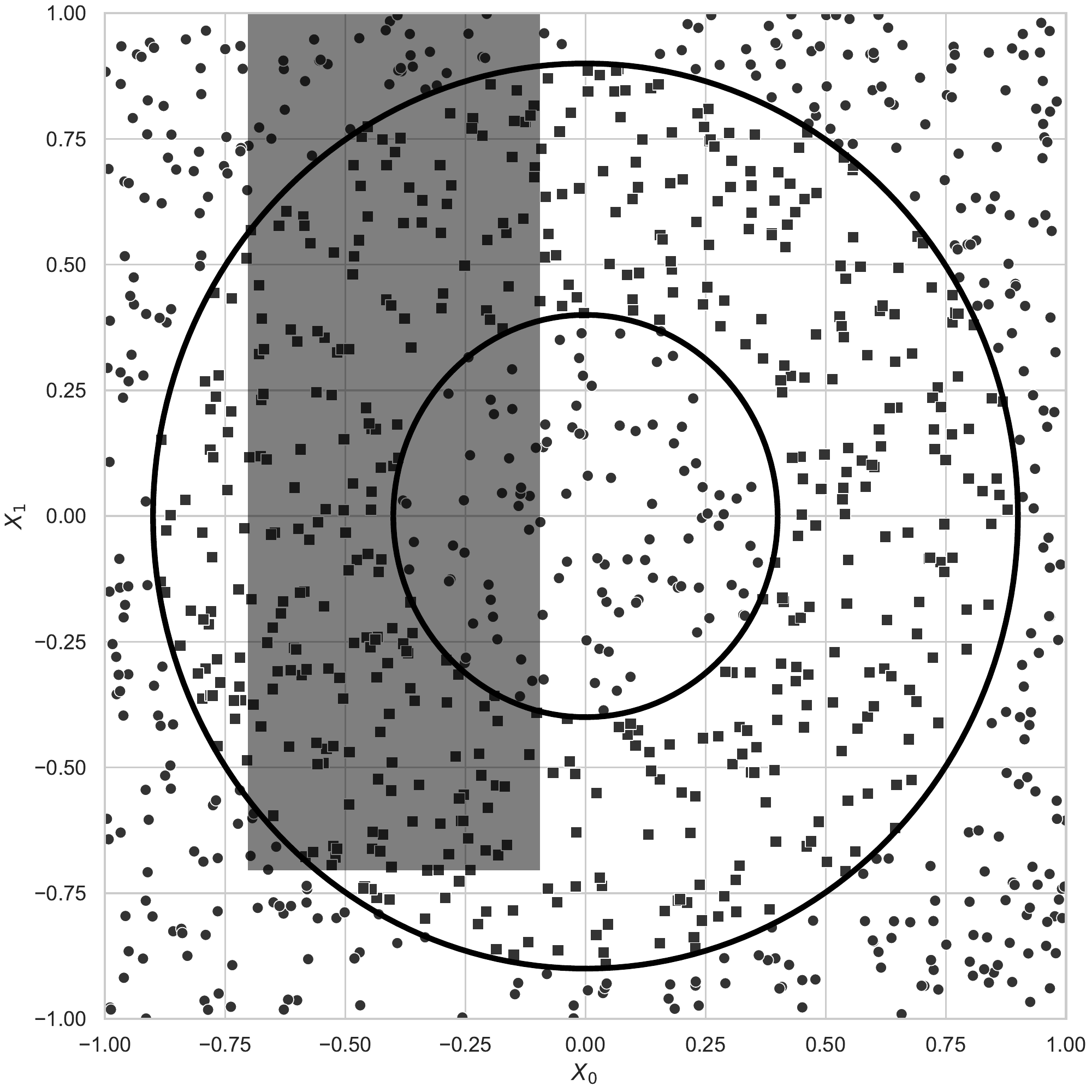}}&
			\subfloat[OPT, $M=5$]{\includegraphics*[scale=0.17]{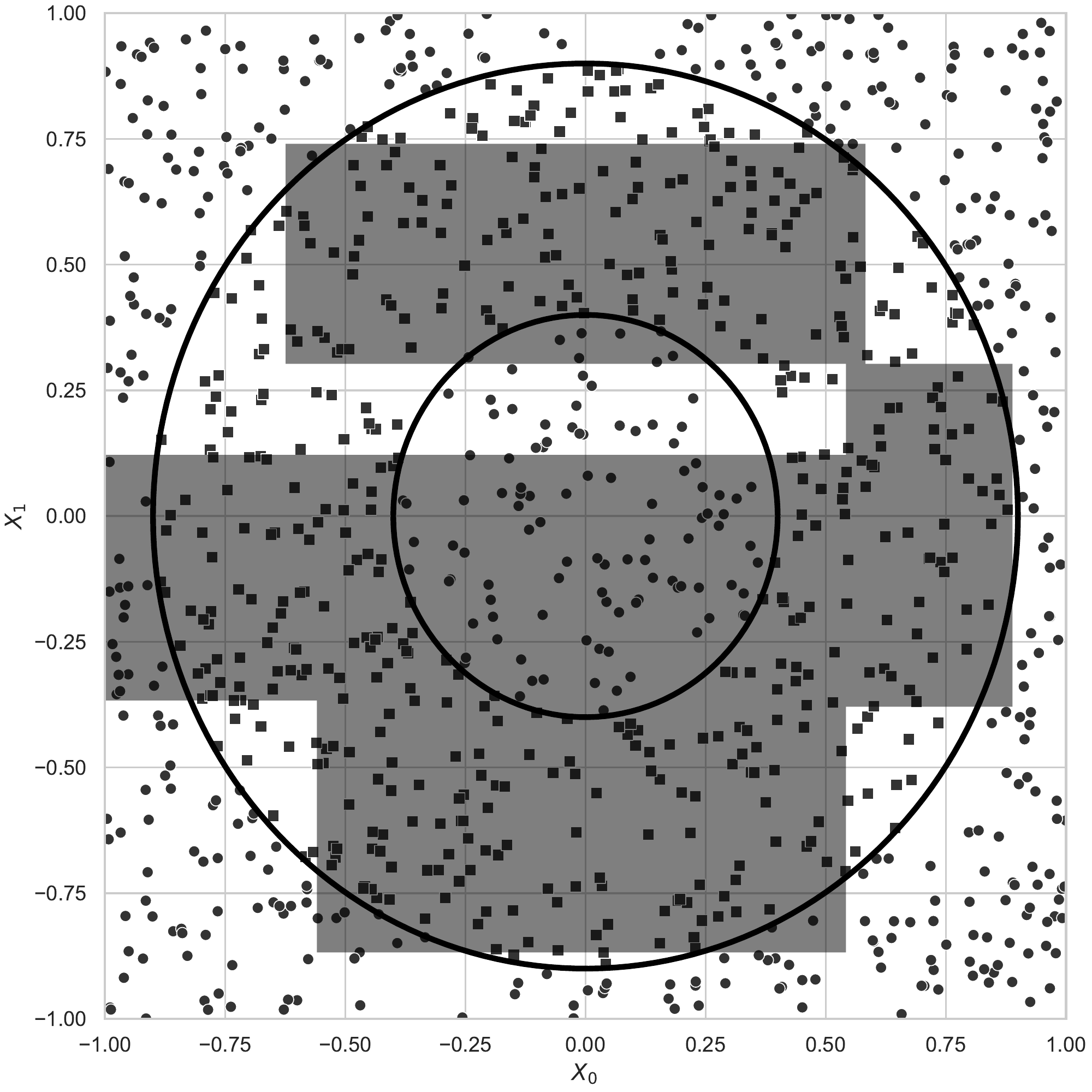}}&
			\subfloat[OPT, $M=10$]{\includegraphics*[scale=0.17]{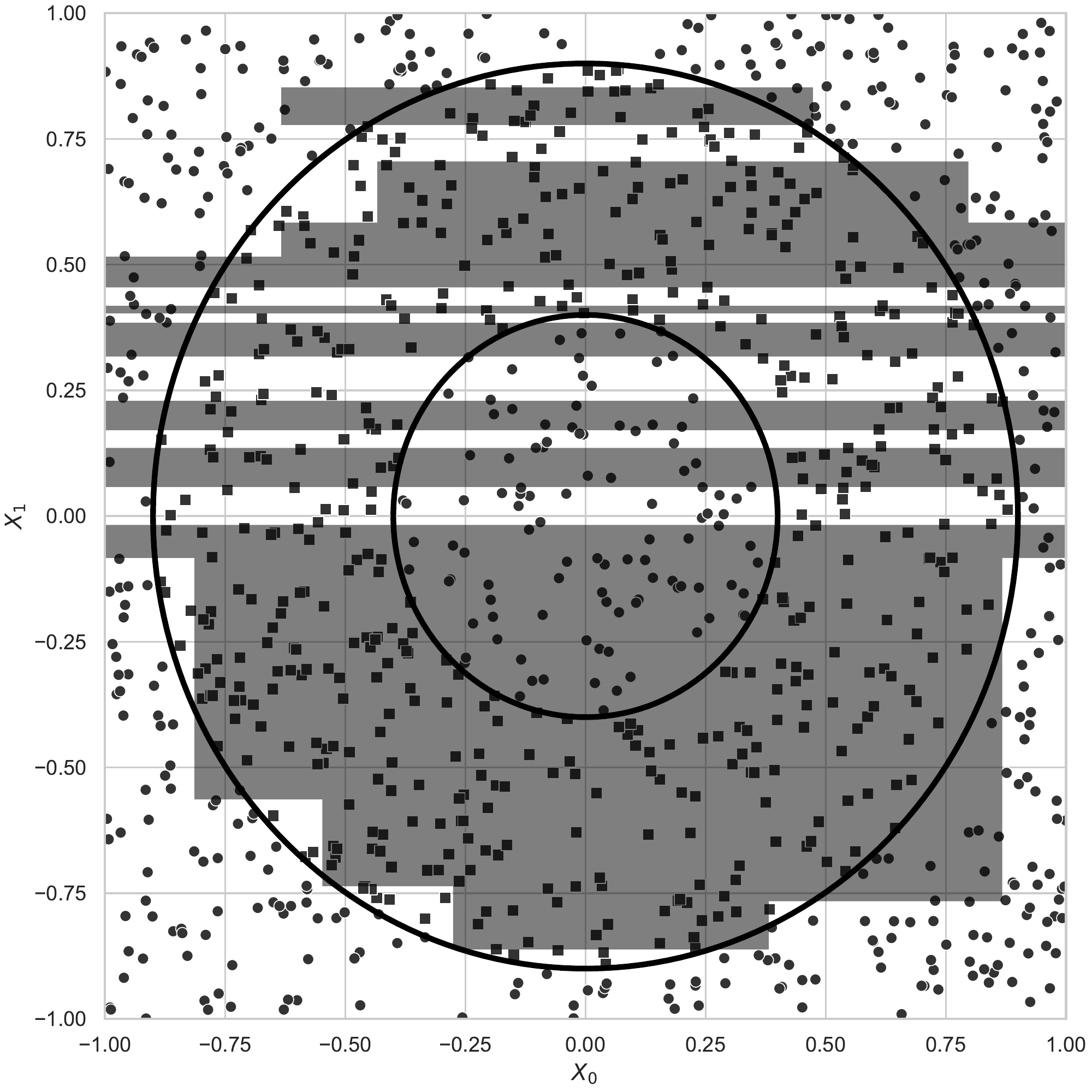}}\\
			
			\subfloat[OPT, $M=20$]{\includegraphics*[scale=0.17]{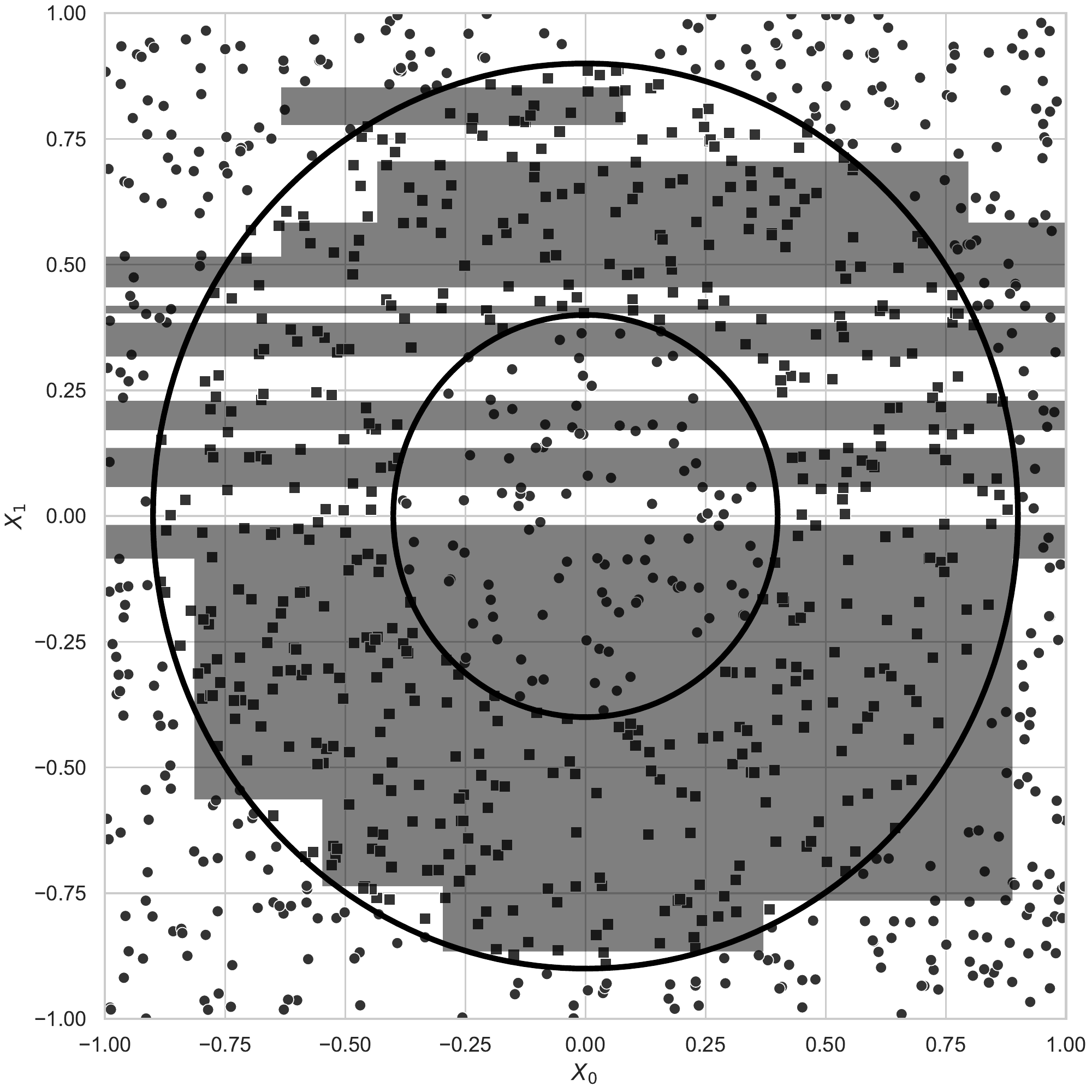}}&
			\subfloat[OPT, $M=30$]{\includegraphics*[scale=0.17]{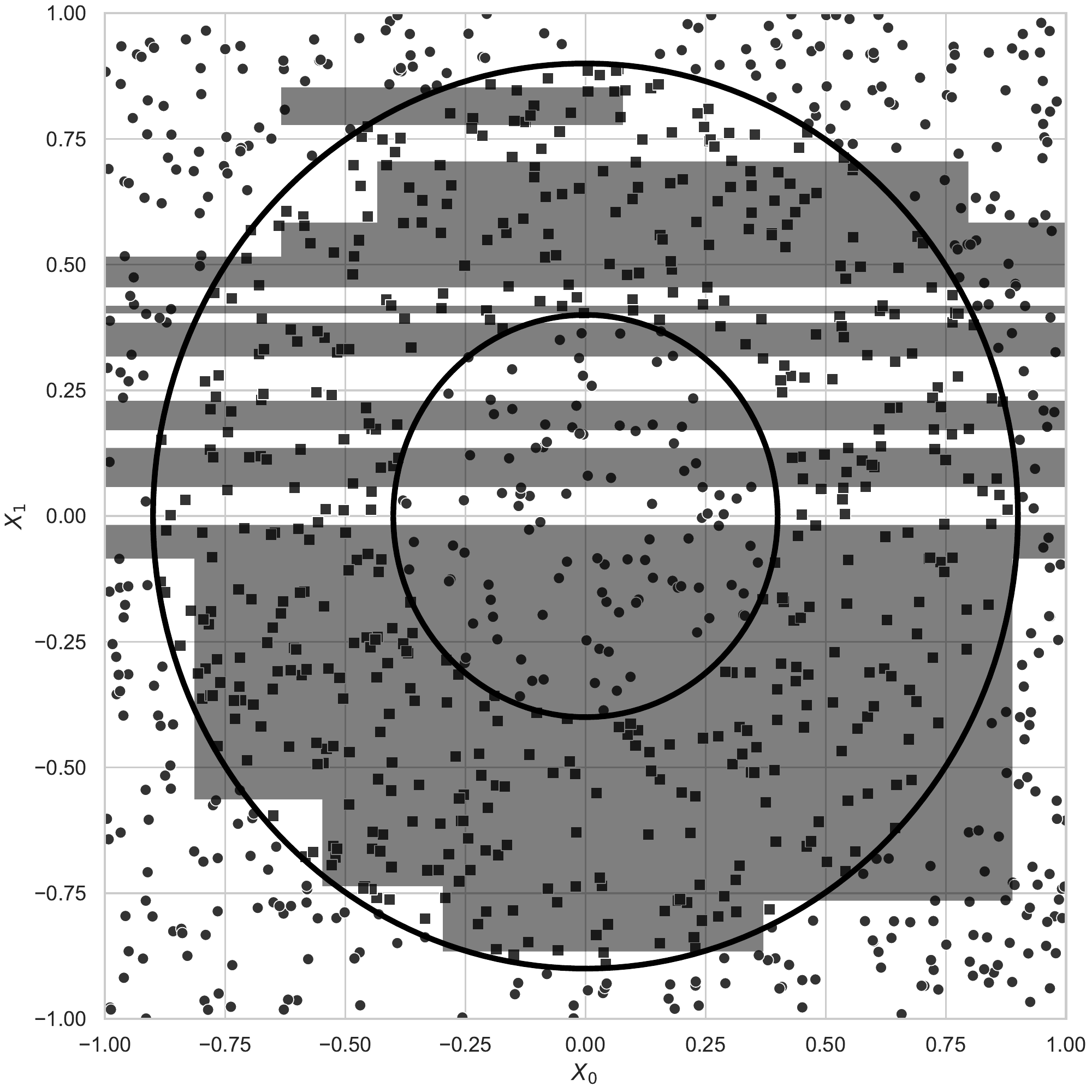}}&
			\subfloat[OPT, $M=50$]{\includegraphics*[scale=0.17]{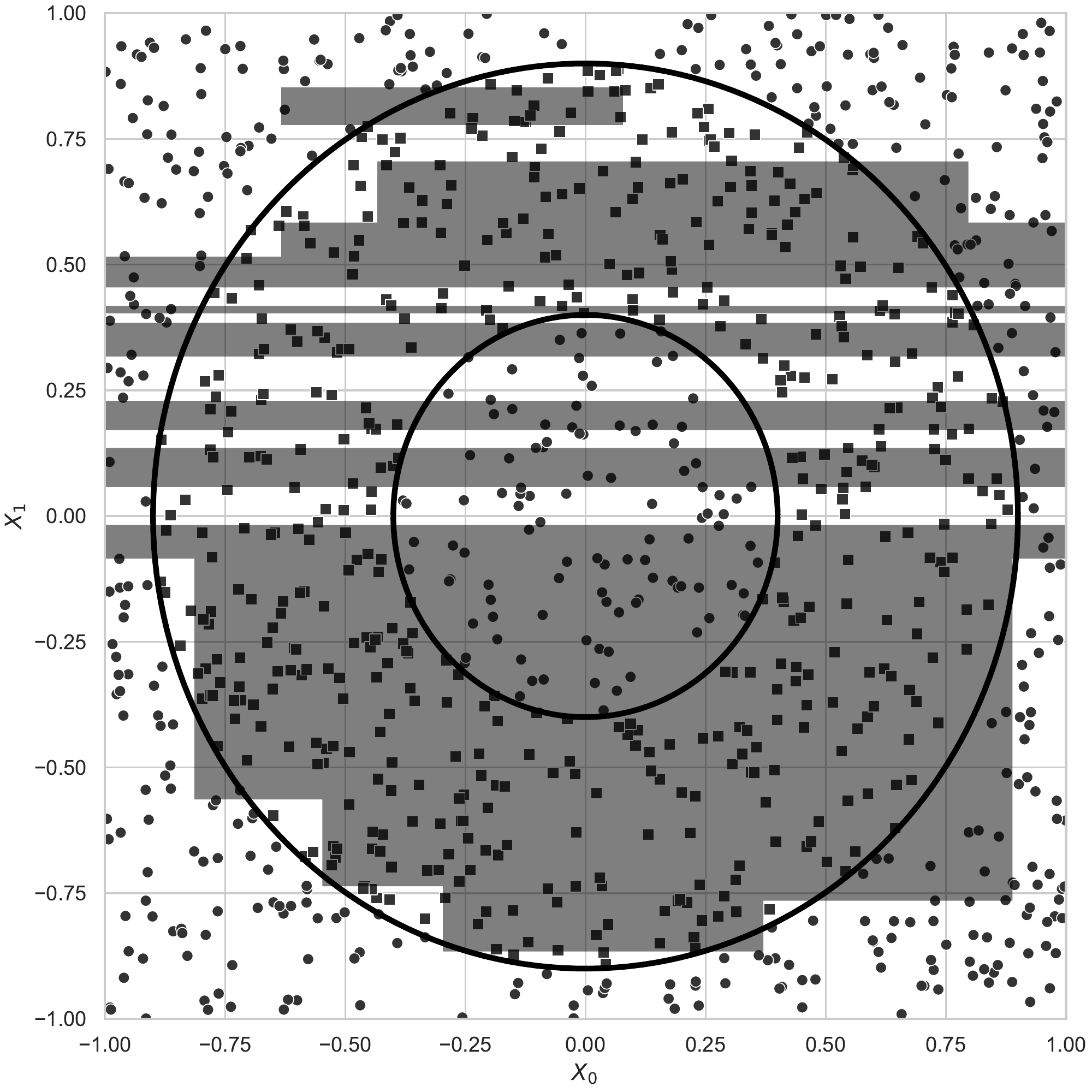}}&
			\subfloat[OPT, $M=100$]{\includegraphics*[scale=0.17]{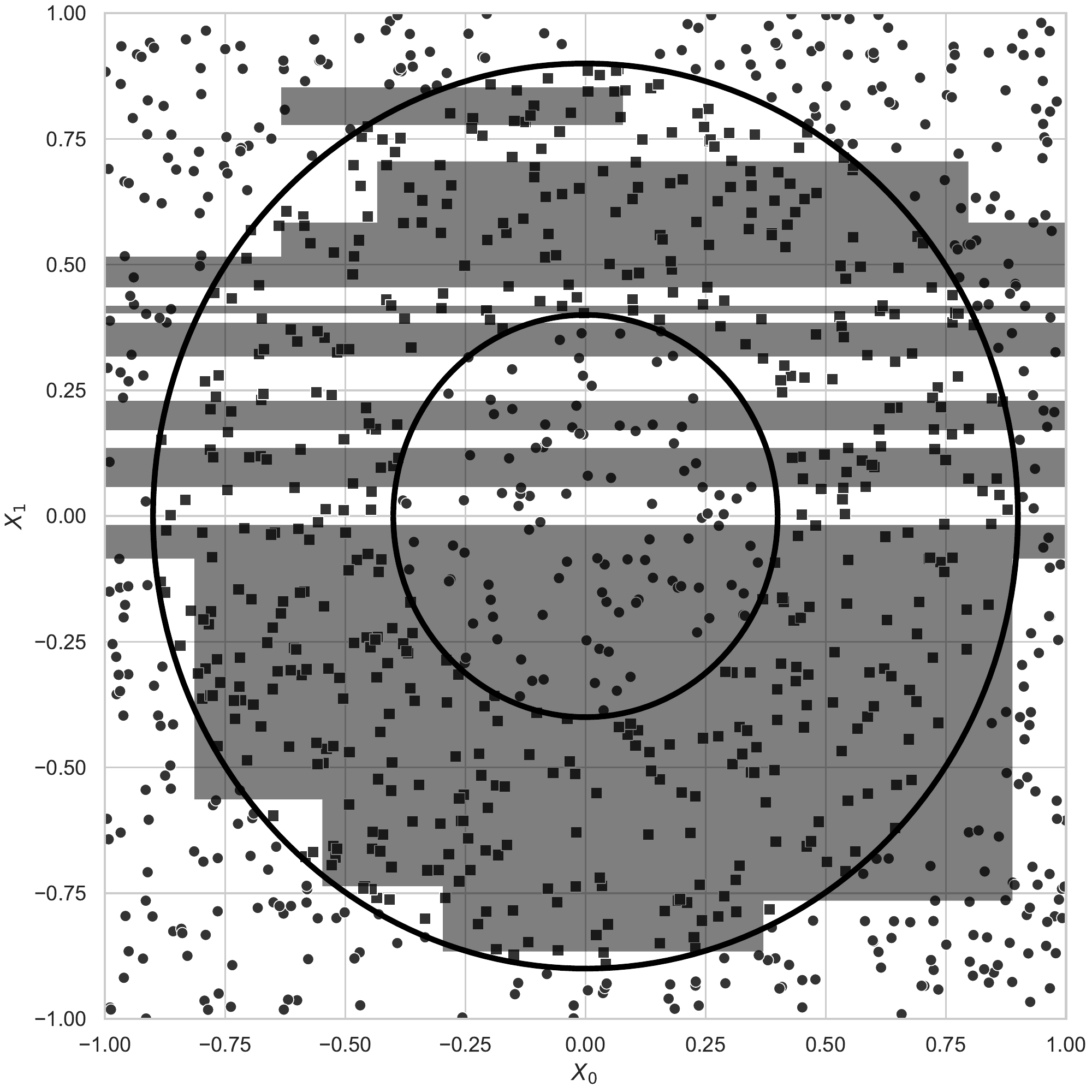}}\\
			
			\subfloat[OPT, $M=200$]{\includegraphics*[scale=0.17]{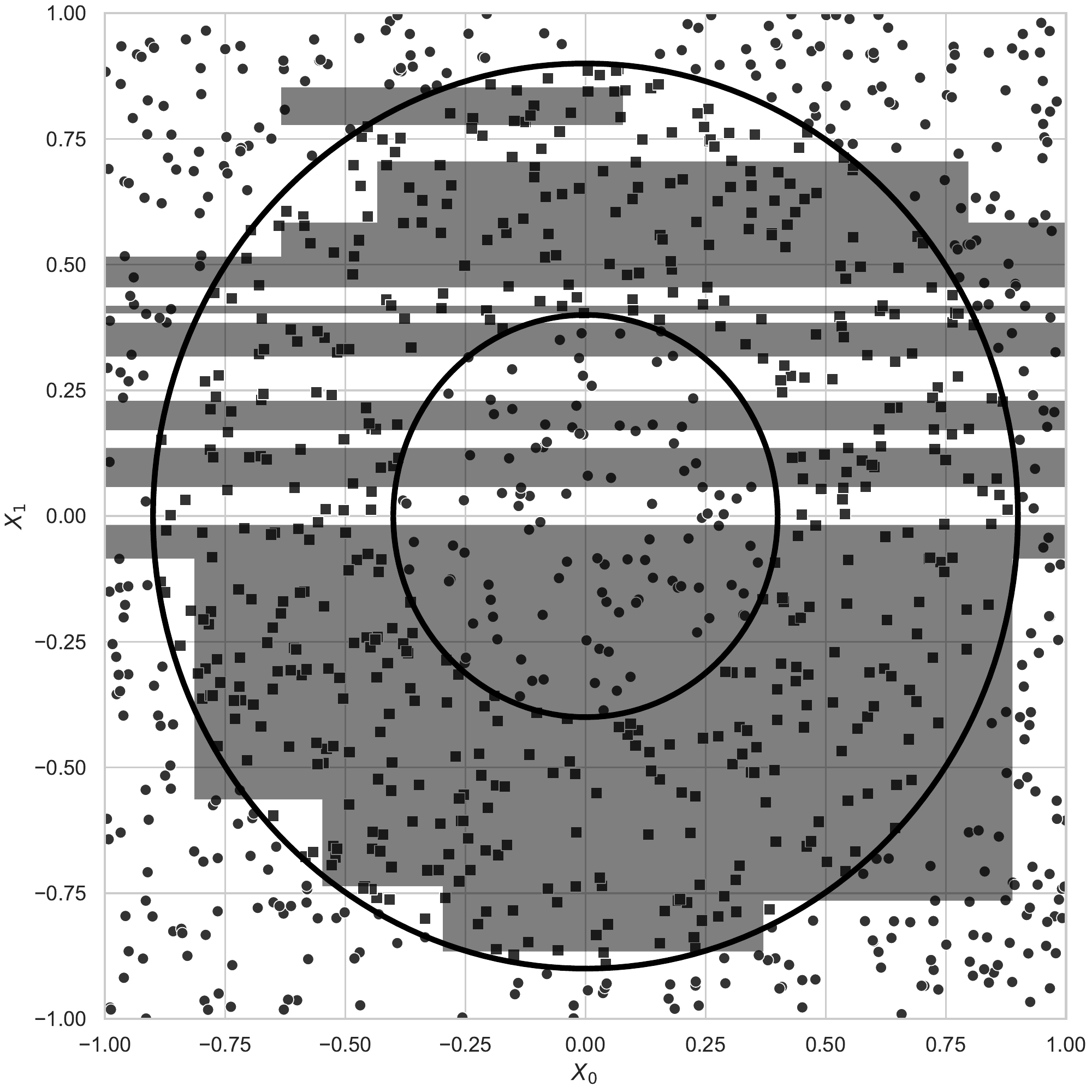}}&
			\subfloat[OPT, $M=300$]{\includegraphics*[scale=0.17]{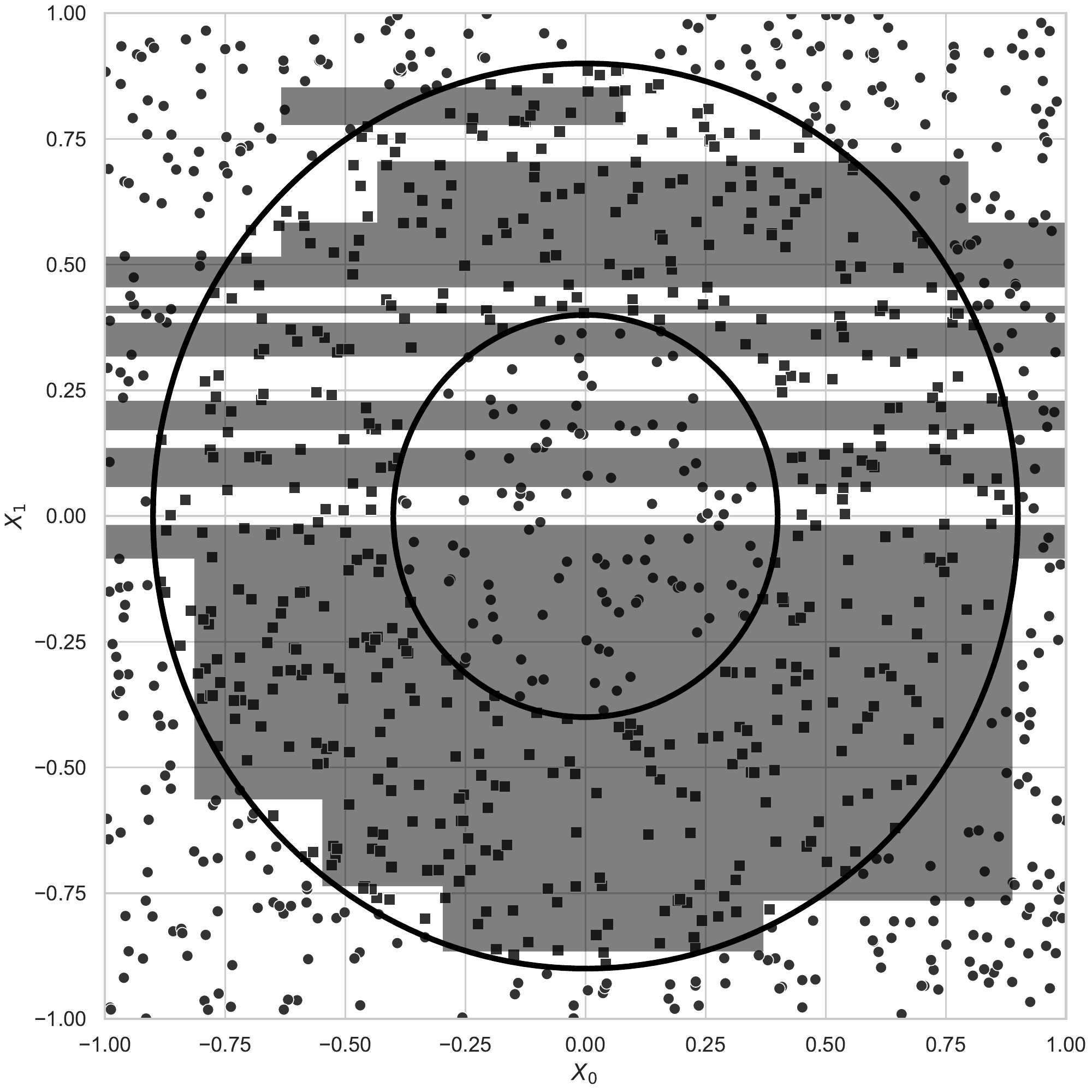}}&
			\subfloat[OPT, $M=500$]{\includegraphics*[scale=0.17]{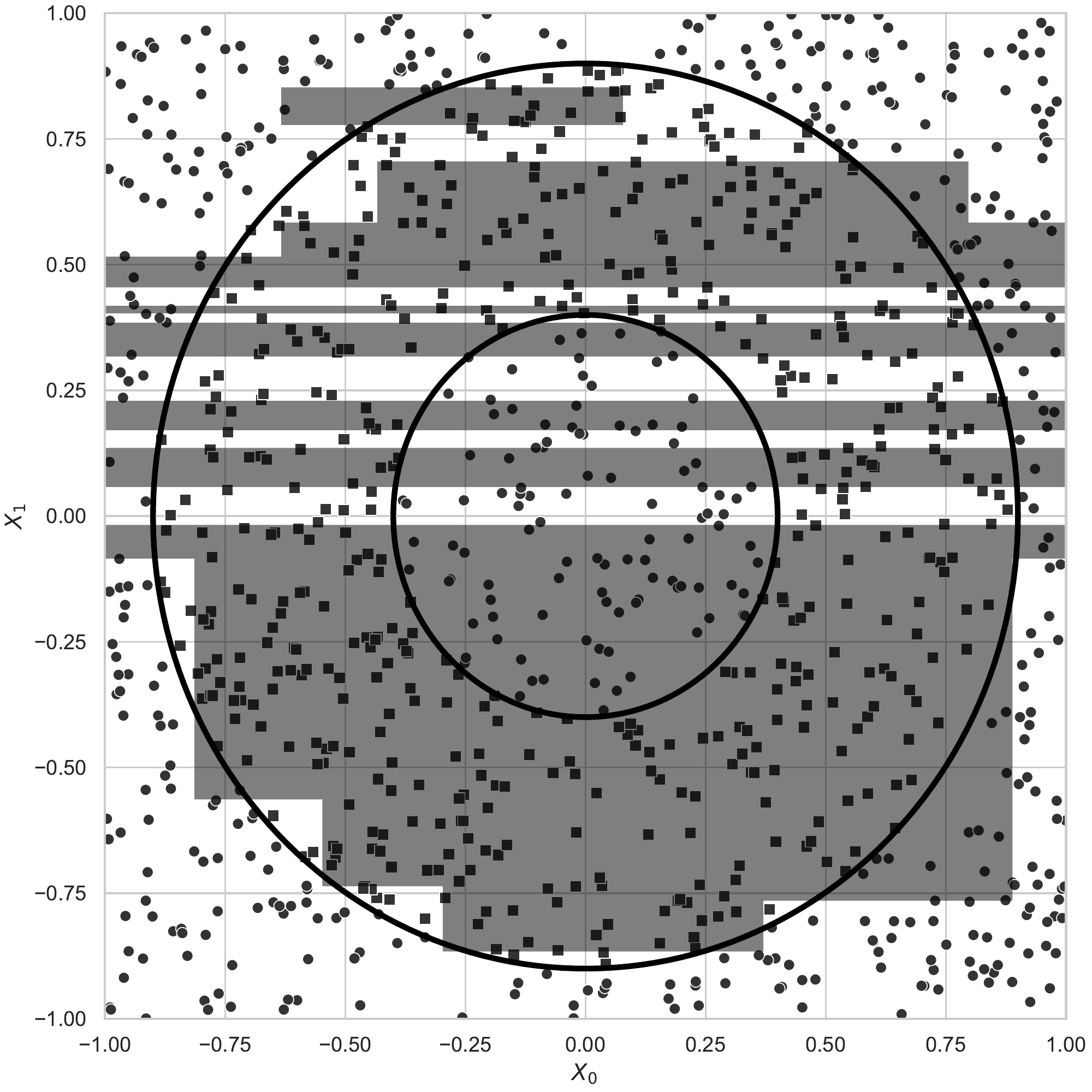}}&
			\subfloat[OPT, $M=1000$]{\includegraphics*[scale=0.17]{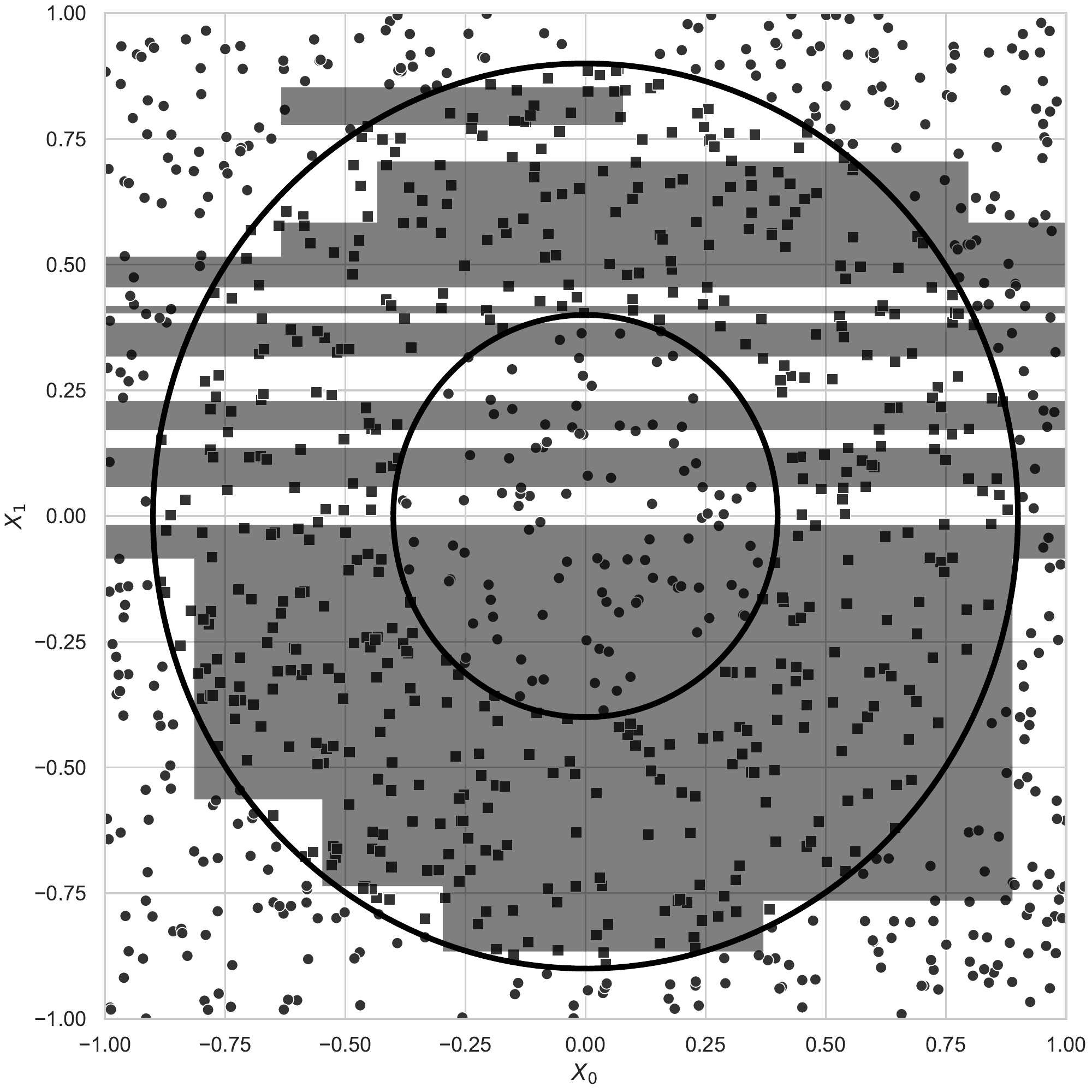}}\\			
		\end{tabular}
	\end{center}
	{\scriptsize{\emph{Notes}. We used the data generated in the second scenario from the illustration setting in \Cref{sec:simulation_study} ($\text{seed}=0$). The figure shows the decision regions returned from IOPL and OPT in gray together with the true decision boundaries (black lines).}}
	\label{fig:approximation_OPT}
\end{figure}

\begin{figure}[H]
	\caption{Regret analysis}
	\begin{center}
		\includegraphics*[scale=0.5]{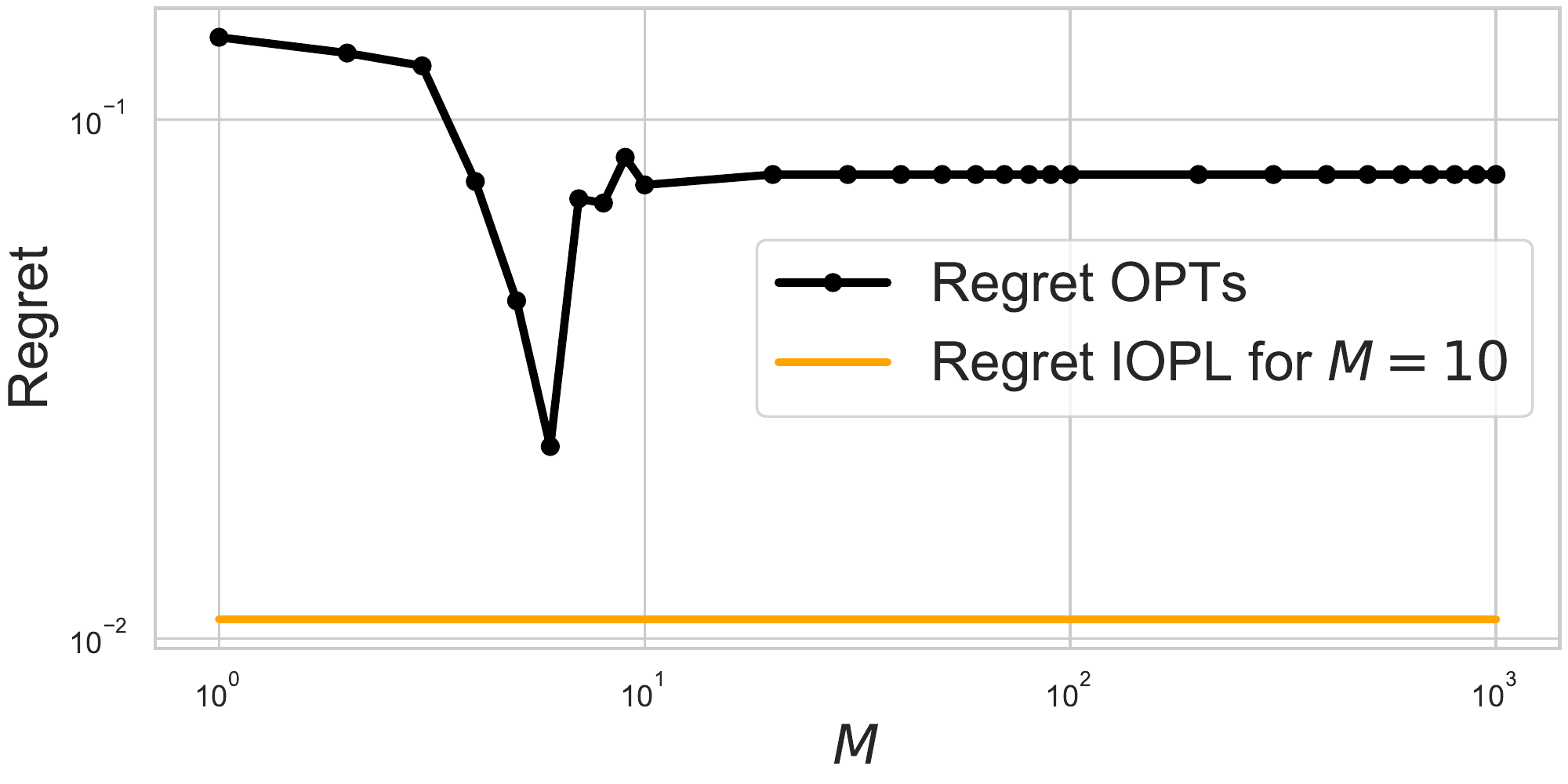}
	\end{center}
	\label{fig:regret_OPT}
	{\scriptsize{\emph{Notes}. Regrets of OPT for the above described experiments. $M$ is varied \emph{only} for OPT according to the hyperparameter grid given in \Cref{tab:hp_grid_OPTs}. The baseline regret from IOPL (with $M=10$) is given in orange.}}
\end{figure}

We demonstrate the first point empirically by fitting OPTs to the data generated in the second scenario in our 2-dimensional simulation study ($\text{seed}=0$, hyperparameters are given in \Cref{tab:hp_grid_OPTs}) and comparing it to IOPL. We note that OPT is not based on standard off-policy learning techniques, \eg, doubly robust methods. It rather estimates the counterfactual outcomes and the policy value at once via a loss function that is given by a convex combination of the two objectives. Hence, we cannot input the exact nuisance functions in OPT. As a remedy, we also refrain from using the exact nuisance functions in IOPL and rather estimate them from data using scikit-learn. We use a random forest for $\mu_t$ and logistic regression for $e_t$. The hyperparameters for the random forest are given in \Cref{tab:hp_grid_OPTs}. For the logistic regression, we used the standard settings in scikit-learn. The results are given in \Cref{fig:approximation_OPT,fig:regret_OPT}. Evidently, OPT is overfitting and cannot benefit from deeper trees, while IOPL yields superior approximations of the true decision region.

\begin{table}[h]
	\caption{Hyperparameter Grids}
	\label{tab:hp_grid_OPTs}
	\centering
	\begin{tabular}{l l l}
		\toprule
		\textbf{Model} & \textbf{Hyperparameters} & \textbf{Tuning range}\\
		\midrule
		Random Forest & minimum number of points in leaf & $\{1,3,5,10\}$\\
		& number of trees & $\{5,10,50,100,300\}$\\
		& maximal depth & $\{5,10,50,100,\text{None}\}$\\
		OPTs & prescription factor $\gamma$ & $\{0.0,0.1,0.2,0.3,0.4,0.5,0.6,0.7,0.8,0.9,1.0\}$\\
		& complexity parameter $\alpha$ & tuned via autotuning procedure in \citet{InterpretableAI}\\
		& minimum number of points in leaf & $\{5,10,15\}$\\
		& maximal depth $M$ & $\{1,2,3,4,5,6,7,8,9,10,\dots$\\
		& &$\phantom{\{1,}20,30,40,50,60,70,80,90,100,\dots$\\
		& &$\phantom{\{1,}200,300,400,500,600,700,800,900,1000\}$\\
		\bottomrule
	\end{tabular}
\end{table}

For the second point, we can think of a worst-case depth that a decision tree has to have, to ensure that a policy in $\Pi_{H}^M$ can be represented as a tree. One can easily check that writing a single hyperbox clause as a decision tree requires -- in general -- a depth of $2d$, where $d$ is the dimension of the covariate space. For every hyperbox that is added, an additional depth of $2d$ is required, as a point can always lie outside the box in just one dimension. That is, to write a policy in $\Pi_{H}^M$ as a decision tree, a depth of -- in general -- $2dM$ is required. For our real-world setting that would result in a worst-case depth of $2\cdot12\cdot5=120$ to represent a policy with $M=5$. That is, although the same information is represented, our policy class allows for a more compact representation, see, \eg, our policy in Figure~\labelcref{fig:example_policy}.

\FloatBarrier
\newpage
\section{Details on Regret Analysis}
\label{app:regret_analysis}

For the regret analysis, we sample 10,000 observations from the described data-generating process, cf. Equation (45) and (47) in \cite{Athey.2021} and use $n=250$ for training, cf. \cite{Zhang.2018}. We train IOPL and all baselines according to \Cref{app:baselines}. The hyperparameter grids for all baselines are given in \Cref{tab:hp_grids}.

\section{Details on ACTG 175 Study}
\label{app:actg_study}

The ACTG 175 study assigned four treatments randomly to 2,139 subjects with human immunodeficiency virus (HIV) type 1, whose CD4 counts were 200--500 cells/$\text{mm}^3$. The four treatments that were compared are the zidovudine (ZDV) monotherapy, the didanosine (ddI) monotherapy, the ZDV combined with ddI, and the ZDV combined with zalcitabine (ZAL).

There are 5 continuous covariates: age (years), weight (kg), CD4 count (cells/$\text{mm}^3$) at baseline, Karnofsky score (scale of 0-100), CD8 count (cells/$\text{mm}^3$) at baseline. They are scaled before further analysis. In addition, there are 7 binary variables: gender ($1 =$ male, $0 =$ female), homosexual activity ($1 =$ yes, $0 =$ no), race ($1 =$ nonwhite, $0 =$ white), history of intravenous drug use ($1 =$ yes, $0 =$ no), symptomatic status ($1 =$ symptomatic, $0 =$ asymptomatic), antiretroviral history ($1 =$ experienced, $0 =$ naive), and hemophilia ($1 =$ yes, $0 =$ no). We chose the increase in the CD4 cell count after 20 weeks to be the continuous response.

\section{Empirical Policy Values in Experiments with Real-World Clinical Data}
\label{app:empirical_policy_values_clinical_data}

As we use real clinical data, we do not have access to the true policy value, which impedes a regret analysis in this setting. For the sake of completeness, we report the empirical policy values in \Cref{tab:empirical_policy_values}. Due to the computational complexity of \Cref{alg:DL_optimization} for the computation of decision lists, we skipped DL in \Cref{tab:empirical_policy_values}.

\begin{table}[h]
	\caption{Empirical policy values for experiments with real-world clinical data}
	\begin{center}
		\begin{tabular}{lccc}
			\toprule
			\textbf{Algorithm} & \multicolumn{3}{c}{\textbf{Empirical Policy Value}}\\
			& $M=1$ & $M=3$ & $M=5$\\
			\midrule
			IOPL & -0.0353 & -0.0433 & -0.0540\\
			DT & -0.0326 & -0.0326 & -0.0349\\
			ES & \multicolumn{3}{c}{-0.0021}\\
			DNN & \multicolumn{3}{c}{-0.0340}\\
			\bottomrule
		\end{tabular}
		\label{tab:empirical_policy_values}\\
		{\scriptsize \emph{Notes}. Empirical policy values as defined in \Cref{eq:empirical_policy_value} for IOPL and baselines. Lower is better}
	\end{center}
\end{table}

\newpage
\section{Questionnaire for User Study}
\label{app:user_study}

\subsection{Questionnaire}

We sent the following questionnaire to our participants from clinical practice:

\emph{We are currently working on a novel artificial intelligence (AI) method for treatment recommendations in medicine. Can I kindly ask for your help with the following survey (5 min)?}

\emph{\textbf{Our research goal}}

\emph{Our goal is to develop a decision support system that recommends treatments to patients. The system will receive characteristics of patients as input (e.g., age, gender, previous diseases). It then returns a recommended treatment as output.}

\emph{What makes our work unique to earlier research is that our recommendations should be “interpretable” and at the same time “accurate”. We understand the need in medicine to make such treatment recommendations in a manner that is transparent and accountable, as well as, reliable. Specifically, practitioners should understand \underline{which} treatment is chosen \underline{when}.}

\emph{\textbf{Background}}

\emph{Many existing tools from artificial intelligence are ``black box''. A medical professional would \textbf{not} be able to understand which treatment is chosen when. Let’s make an example. Let’s assume Alice, 25 arrives at a medical professional. Let’s further assume she has a medical record with three previous conditions A, B, and C. Then, the artificial intelligence will recommend a treatment based on her gender, age, risk factors, and previous conditions. However, a medical professional would not know which risk factors led to the treatment decision. For example, it would be totally unclear whether age or gender was actually considered (and how). On the contrary, the artificial intelligence would only state treatment 1 or treatment 2 – and that’s it. Aka: only the treatment decision would be presented, without information and without explanation on how to arrive at it. It would thus also not be possible to compare the underlying decision logic with textbook knowledge. As a remedy, interpretable artificial intelligences have been proposed, \eg, eligibility scores, decision lists or decision trees. However, these models have no theoretical guarantees to be expressive enough to capture complicated dependencies between patient characteristics and treatments.}  

\emph{\textbf{Example setting}}

\emph{In the following, we would like to assess how interpretable our artificial intelligence is. More specifically, we would like to compare our artificial intelligence against two existing interpretable artificial intelligences in terms of interpretability: (i)~linear eligibility score and (ii)~decision trees. (We give you later examples of how these look like). However, the existing AI systems come without theoretical guarantees for “accuracy”, so one \underline{cannot} ensure that these are expressive enough to capture also complicated dependencies (\eg, rare comorbidities).}

\emph{For this, we decided upon the following example setting with actual clinical data. In particular, we draw upon the AIDS Clinical Trial Group (ACTG) study 175. The ACTG 175 study assigned two treatments randomly to 1,056 subjects with human immunodeficiency virus (HIV) type 1, whose CD4 counts were 200-500 cells/mm3. The two treatments are: (1) zidovudine (ZDV) monotherapy combined with zalcitabine (ZAL), denoted by \textbf{ZDV-ZAL}. (2) zidovudine monotherapy only, in the following denoted by \textbf{ZDV}.}

\emph{All AI systems will consider the following patient characteristics:}
\begin{itemize}
	\item \emph{Age (years)}
	\item \emph{Weight (kg)} 
	\item \emph{Gender (male, female)} 
	\item \emph{Homosexual activity (yes, no)} 
	\item \emph{Race (white, non-white)} 
	\item \emph{History of intravenous drug use (yes, no)} 
	\item \emph{Symptomatic Status (symptomatic, asymptomatic) }
	\item \emph{Antiretroviral history (experienced, na{\"i}ve)} 
	\item \emph{Hemophilia (yes, no)}
	\item \emph{Karnofsky Score (scale of 0--100)}
	\item \emph{CD4 count at baseline, i.e., before treatment (cells/mm3)} 
	\item \emph{CD8 count at baseline, i.e., before treatment (cells/mm3)} 
\end{itemize}

\emph{\textbf{SYSTEM 1: Our ‘new’ artificial intelligence}}

\emph{In our ‘new’ artificial intelligence, the treatment recommendation is presented in a simple flow chart. You simply read it from top to bottom. For each ``if'', you check whether the characteristics of a patient match the stated condition. For instance, in the first line, you first check whether you treat a patient between 13 and 62 years, and with a weight between 47.8 and 87.09 kg, etc. The recommended treatment -- ZDV-ZAL or ZDV -- is stated on the right-hand side.}

\emph{A medical practitioner would thus check which of the ``if'' conditions is fulfilled, and then pick the corresponding treatment. If one doesn’t fit, just check the next. If none fits, you follow the treatment suggested in the else clause.}

\emph{Let’s make an illustrative example. So, a patient at the age of 70 would receive ZDV because none of the green condition would apply. A patient at the age of 20, with a weight of 50, a CD4 count at 122, a Karnofksy score at 90, and a CD8 count at 200 would match the first clause and receive ZDV-ZAL.}

\FloatBarrier
\begin{figure}[h]
	\begin{center}
		\includegraphics*[scale=0.4]{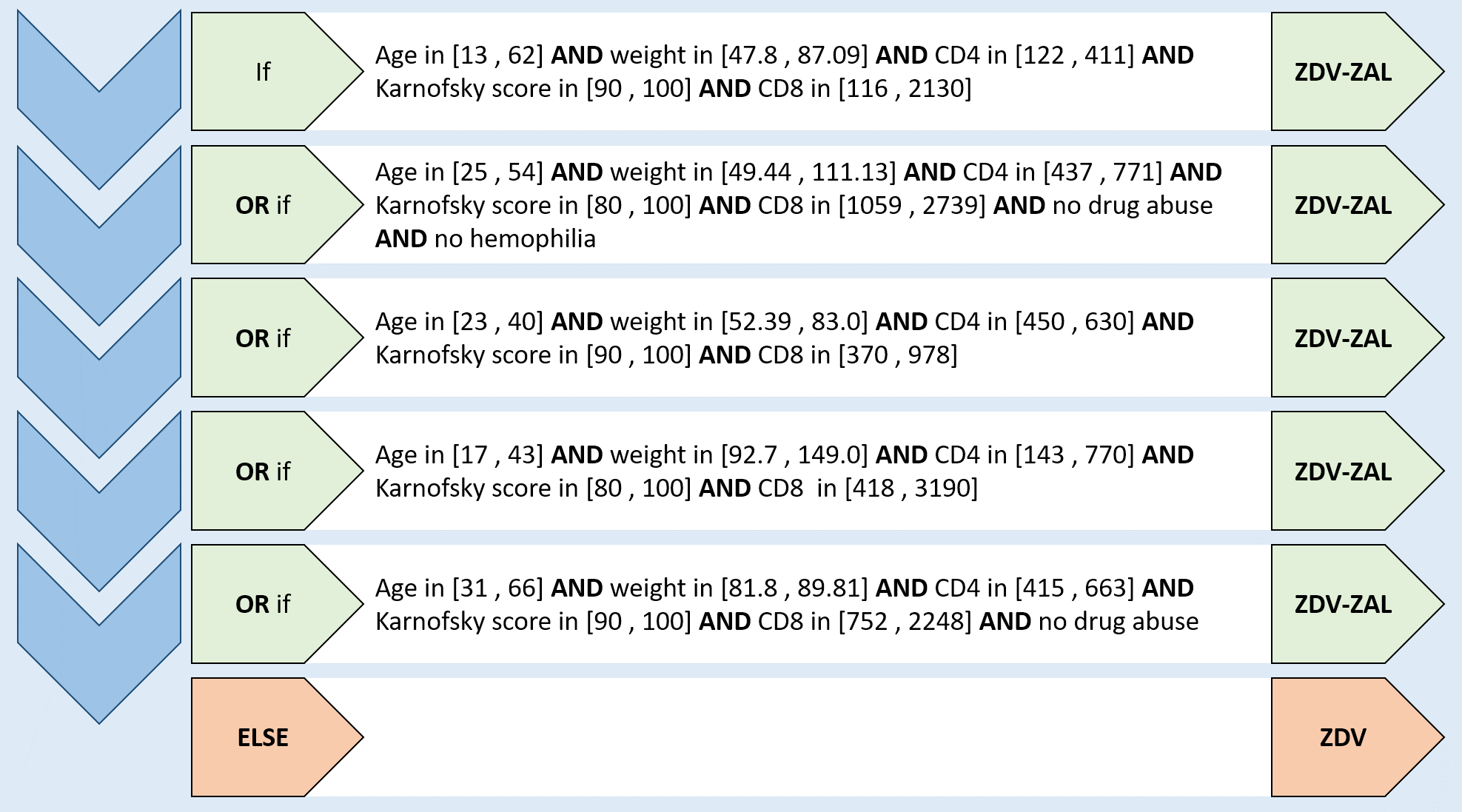}
	\end{center}
\end{figure}
\FloatBarrier

\newpage
\emph{\textbf{SYSTEM 2: Linear eligibility score}}

\emph{For this artificial intelligence, the treatment recommendation must be manually calculated via a linear mathematical rule. Here, each patient characteristic is first multiplied with a weight (e.g., you multiply patient age with 0.0262, you multiply the patient weight with 0.0075). Afterward, the values are added. The resulting number is called ``eligibility score''. If the score is greater or equal to zero, the treatment ZDV-ZAL is chosen. If the score is smaller than zero, treatment ZDV is chosen.}

\FloatBarrier
\begin{figure}[h]
	\begin{center}
		\includegraphics*[scale=0.4]{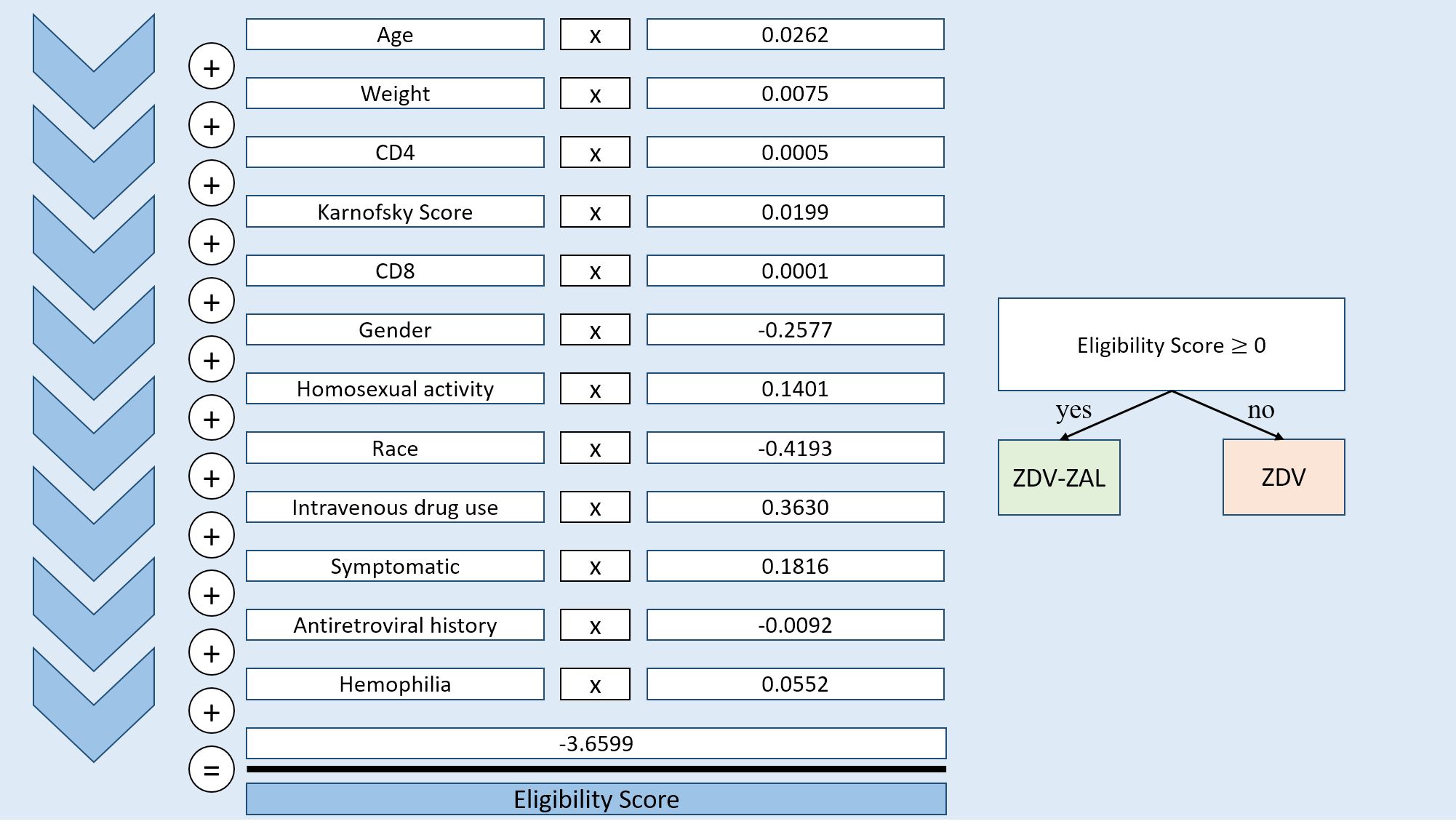}
	\end{center}
\end{figure}
\FloatBarrier

\emph{\textbf{SYSTEM 3: Decision tree}}

\emph{Here the treatment recommendation follows a simple tree structure. You simply check the conditions from top to bottom and follow the tree in order of the corresponding answers: for each white box, you check what the correct answer is. Then you move left if the answer is ‘yes’ and right if the answer is ‘no’. If you arrive at a leaf node, you read off the suggested treatment.}

\FloatBarrier
\begin{figure}[h]
	\begin{center}
		\includegraphics*[scale=0.4]{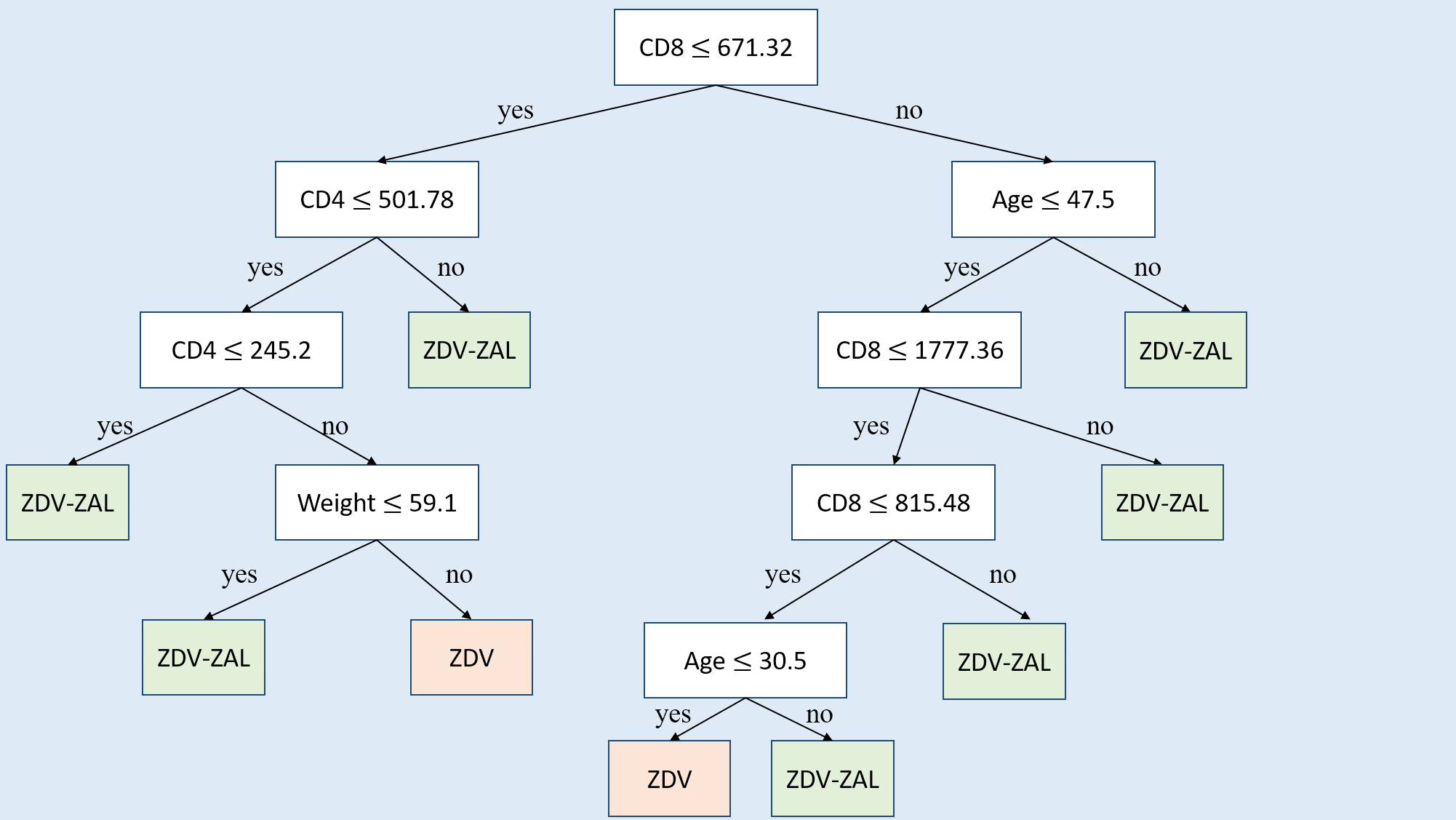}
	\end{center}
\end{figure}
\FloatBarrier

\newpage
\emph{\textbf{How to fill out the survey}}

\emph{Please find our four survey questions below. For simplicity, we would ask you to just return them in an email.}

\begin{enumerate}
	\item \emph{How interpretable do you find the logic in our ‘new’ artificial intelligence? Please rate it in a range from 0 to 10, where 0 is ``black box'' and 10 ``fully transparent''. A 5 would be neutral.}
	\item \emph{How interpretable do you find the logic in linear eligibility scores? Please rate it in a range from 0 to 10, where 0 is ``black box'' and 10 ``fully transparent''. A 5 would be neutral.}
	\item \emph{How interpretable do you find the logic in decision trees? Please rate it in a range from 0 to 10, where 0 is ``black box'' and 10 ``fully transparent''. A 5 would be neutral.}
	\item \emph{Would you consider using treatment recommendations when presented through our ‘new’ artificial intelligence? Yes or no?}
\end{enumerate}

\subsection{Evaluation}

\FloatBarrier

We received the following answers; see \Cref{tab:survey}.

\begin{table}[h]
	\caption{Evaluation of Survey}
	\begin{center}
		\begin{tabular}{clllcccc}
			\toprule
			\textbf{No.} & \textbf{Institution} & \textbf{Job Description} & \textbf{Country} & \textbf{Q 1} & \textbf{Q 2} & \textbf{Q 3} & \textbf{Q 4}\\
			\midrule
			1 & University Hospital & Medical Doctor & Germany & 7 & 3 & 8 & yes\\
			2 & Company & Pharmacist & Sweden & 7 & 4 & 9 & yes\\
			3 & Hospital & Clinical Pharmacist & Sweden & 9 & 5 & 7 & yes\\
			4 & Private Doctor's Office & Medical Doctor & Germany & 4 & 8 & 7 & yes\\
			5 & University Hospital & Medical Doctor & Germany & 7 & 5 & 8 & yes\\
			6 & University Hospital & Medical Doctor & Switzerland & 7 & 7 & 9 & yes\\
			7 & Hospital & Medical Doctor & Germany & 7 & 3 & 8 & yes\\
			8 & University Hospital & Medical Doctor & Switzerland & 6 & 4 & 9 & no\\
			9 & University Hospital & Medical Doctor & Switzerland & 7 & 9 & 8 & yes\\
			10 & University Hospital & Radiology specialist & Germany & 9 & 8 & 9 & yes\\
			\bottomrule
		\end{tabular}
		\label{tab:survey}
	\end{center}
\end{table}

\FloatBarrier

\subsection{Results}

\begin{figure}[H]
	\caption{Survey results}
	\begin{center}
		\includegraphics*[scale=0.3]{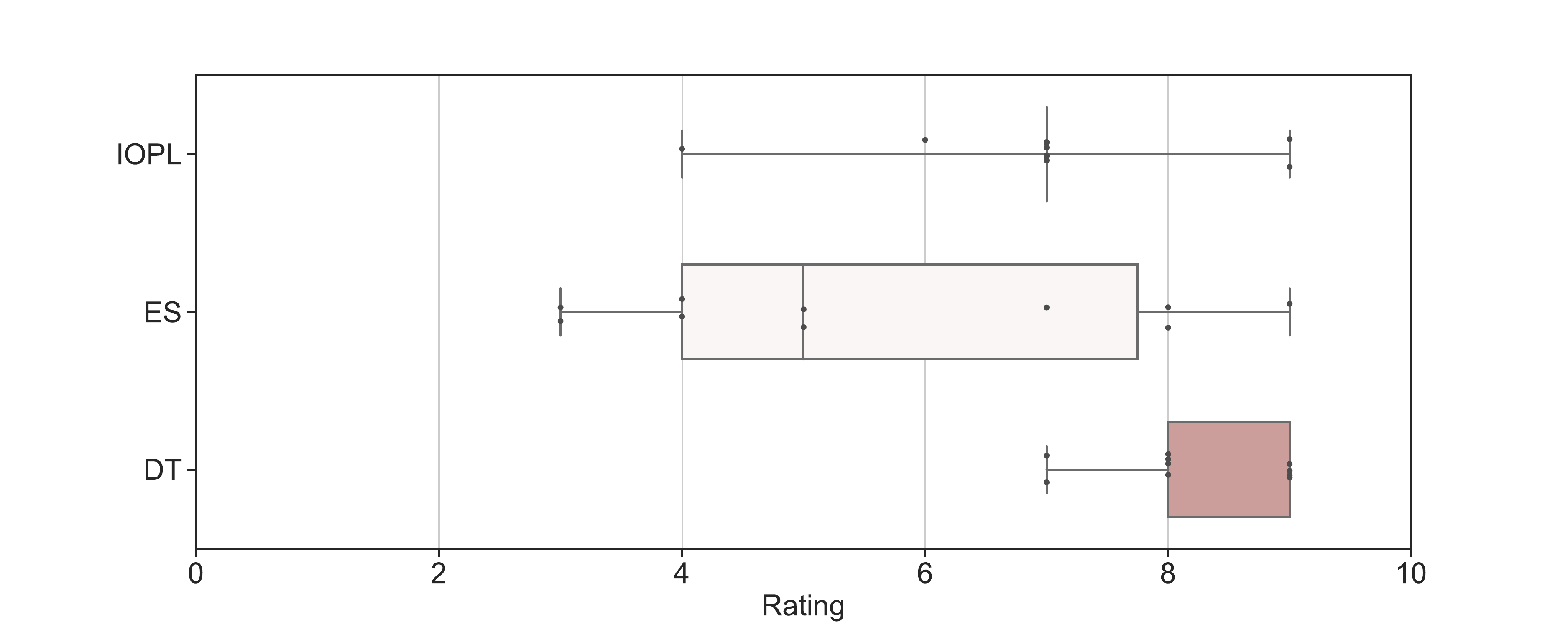}
	\end{center}
	\label{fig:survey_results}
\end{figure}

The ratings are visualized in Figure~\ref{fig:survey_results}. Our results demonstrate that, in terms of interpretability, IOPL is largely on par with our baselines with a slight preference for decision trees.

\section{Computational Considerations in Practice}
We note that our definition of policies in \Cref{eq:definition_policy} is symmetric in the labels $\{1,-1\}$. That is, one can easily change the signs of the training labels and learn a representation of the \emph{no-treatment} decision region. All of our theoretical derivations are of course still valid. In practice, it might thus be useful to compute both representations, as one of them might lead to a more compact policy. That is, in one representation, one might need smaller values of $M$ for the same predictive performance as in the other representation.

\section{Limitations}

The computational complexity of the pricing problem limits our approach to mid-sized datasets ($n\approx1,000$ observations) with a moderate dimensionality ($d<15$). Both of these assumptions are typically fulfilled in clinical practice, as for instance, in the ACTG 175 study used in this study ($n=1,056$, $d=12$).

The aim of this paper is to provide interpretable treatment decisions for clinical practice. We think that circumventing the black-box nature of state-of-the-art approaches in off-policy learning, while at the same time yielding policies with low regrets, is crucial to achieve this goal. At the same time, we want to emphasize that medical decisions should ultimately always be made by medical experts. That is, medical experts should always be informed about the risks associated with data-driven algorithms.

\end{document}